\documentclass{article}

\usepackage{microtype}
\usepackage{graphicx}
\usepackage{subcaption}
\usepackage{booktabs} 
\usepackage{wrapfig}  
\usepackage{ifthen}

\usepackage[ruled,vlined]{algorithm2e}

\DontPrintSemicolon
\SetKwComment{Comment}{\textcolor{violet}{\footnotesize /* }}{\textcolor{violet}{\footnotesize */}}

\usepackage{setspace}

\usepackage{multirow}
\usepackage{threeparttable}
\usepackage{placeins} 
\usepackage{multicol}

\usepackage{hyperref}

\usepackage[accepted]{icml2026}

\usepackage{amsmath}
\usepackage{amssymb}
\usepackage{mathtools}
\usepackage{amsthm}
\usepackage{amsfonts}

\usepackage[utf8]{inputenc}
\usepackage[T1]{fontenc}

\usepackage{url}
\usepackage{nicefrac}
\usepackage{xcolor}
\usepackage{makecell}

\usepackage{xspace}
\usepackage{xifthen}
\usepackage{ifthen}
\usepackage{xifthen}

\usepackage[capitalize,noabbrev]{cleveref}

\usepackage[textsize=tiny]{todonotes}
\usepackage{balance}
\usepackage{flushend}

\usepackage[table]{xcolor}

\icmltitlerunning{pTNAS: Progressive Neural Architecture Search for Tabular Data}

\newcommand{\cmt}[1]{\textcolor{violet}{\footnotesize /* #1 */}}

\newcommand{\eps}{\epsilon}

\newcommand{\framework}{pTNAS\xspace}
\newcommand{\atlasname}{approach\xspace}

\newcommand{\tfmem}[1][]{%
  \ifthenelse{\equal{#1}{}}{zero-cost proxy\xspace}{zero-cost proxies\xspace}%
}
\newcommand{\newtfmem}{pTProxy\xspace}

\newcommand{\fnn}{Fully-connected neural network\xspace}

\newcommand{\budget}{$T_{\max}$\xspace} 
\newcommand{\nasbench}{\textit{NAS-Bench-Tabular}\xspace}
\newcommand{\succhalf}{SH\xspace}

\newcommand{\highlight}[1]{\textbf{#1}}

\newcommand{\filter}{coarse-grained filtering\xspace}
\newcommand{\refine}{fine-grained refinement\xspace}

\newcommand{\scheme}{filter-and-refine optimization strategy\xspace}

\newcommand{\best}[1]{{\textbf{\underline{#1}}}}

\usepackage{tikz}

\begin{document}

\twocolumn[
  \icmltitle{pTNAS: Progressive Neural Architecture Search for Tabular Data}

  \begin{icmlauthorlist}
    \icmlauthor{Naili Xing}{nus}
    \icmlauthor{Shaofeng Cai}{nus} 
    \icmlauthor{Lingze Zeng}{nus}
    \icmlauthor{Jiaqi Zhu}{nus}
    \icmlauthor{Peng LU}{zju}
    \icmlauthor{Jian Pei}{dk}
    \icmlauthor{Beng Chin Ooi}{zju}
  \end{icmlauthorlist}

  \icmlaffiliation{nus}{National University of Singapore, Singapore}
  \icmlaffiliation{zju}{Zhejiang University, China}
  \icmlaffiliation{dk}{Duke University, USA}

  \icmlcorrespondingauthor{Shaofeng Cai}{shaofeng@comp.nus.edu.sg}
  
  \icmlkeywords{Neural Architecture Search, Tabular Data Analysis}
  \vskip 0.3in
]

\printAffiliationsAndNotice{}

\begin{abstract}
Recent advances have shifted the paradigm of tabular learning toward tabular foundation models, yet their accuracy relies on a heavy inference cost that scales poorly with context size.
Deep neural networks remain a highly competitive and more efficient modeling paradigm when equipped with well-designed architectures; however, identifying such architectures in a data-adaptive and budget-aware manner remains challenging.
We propose pTNAS, the first progressive neural architecture search (NAS) approach tailored for tabular data, which enables fast identification of a viable architecture and continuously improves its search performance as more budget becomes available.
pTNAS adopts a filter-and-refine optimization strategy that combines efficient training-free and effective training-based architecture evaluation.
In the filtering phase, we introduce pTProxy, a novel zero-cost proxy specifically designed for tabular networks that jointly captures architectural trainability and expressivity, enabling fast filtering of large architecture search spaces.
In the refinement phase, pTNAS employs a fixed-budget scheduling algorithm to accurately identify the best-performing architecture from a small set of promising candidates.
We further propose a budget-aware coordinator to optimize budget allocation holistically.
Experiments show that pTNAS reduces the time to reach the globally best architecture by up to 82.75$\times$ compared with other NAS approaches, achieves the best average predictive rank, and improves end-to-end efficiency by up to 4.78$\times$ compared with TabPFN.
\end{abstract}

\section{Introduction}\label{sec:introduction}

Tabular data remains the most common format in real-world applications, such as financial modeling, medical diagnosis, and recommendation systems~\cite{ChengHYSW024,ChengJZLG25}. 
In these high-stakes domains, even marginal gains in predictive accuracy can translate into significant economic or social value~\cite{arm_net,zhu2023meter}. 
However, achieving peak performance on tabular data is challenging and typically requires sophisticated modeling~\cite{McElfreshKVCRGW23}.

Recent advances have rapidly reshaped tabular learning, shifting the dominant paradigm from robust tree ensembles (e.g., GBDTs)~\cite{KeMFWCMYL17,ProkhorenkovaGV18} to deep tabular models and, most recently, Tabular Foundation Models (TFMs)~\cite{BreugelS24,RobertsonH0AH25}.
TFMs leverage large-scale pretraining and in-context learning (ICL) to make predictions conditioned on a small labeled sample set provided as context, reducing or even eliminating per-dataset training~\cite{tabfpn,tabicl,zhu2025context}.
However, TFMs shift much of the computational cost to inference, exposing a critical efficiency–accuracy trade-off: while providing more labeled samples as context has the potential to improve adaptation to the target data distribution, inference latency increases with context length and can quickly become a bottleneck~\cite{BonetMGI24,MuellerC025}.
For example, TabPFN~\cite{tabfpn} infers by conditioning on the context set via a pretrained prior over tabular tasks. Yet, it is typically confined to small datasets (e.g., fewer than 1,000 training samples, 100 features, and 10 classes). 
It remains about 10$\times$ slower than comparable tree-based methods, limiting its practical use to only a few thousand training examples~\cite{BonetMGI24}.

The recent momentum around TFMs may suggest that tabular prediction is heading toward foundation-scale inference by default.
Yet, given their inference-time cost, it is worth asking whether foundation-scale inference is truly necessary for strong tabular performance.
Recent studies suggest a promising alternative: even relatively lightweight Deep Neural Networks (DNNs) (e.g., MLPs and ResNets) can be highly competitive if their architectures are properly configured~\cite{abs-2511-08667,gorishniy2021revisiting,BonetMGI24,MuellerC025}.
Despite their simplicity, their stacked nonlinear transformations can capture complex and high-order feature interactions across heterogeneous feature types~\cite{ChengHYSW024}, while remaining much more inference-efficient than TFMs.
This leads us to a natural question: \textit{Is it possible to efficiently discover and train a data-specific DNN architecture that matches the accuracy of long-context TFMs while maintaining the efficiency of GBDTs}? 
Ideally, such a process would be progressive and budget-aware: it would identify a viable architecture within a short time budget and continuously improve its performance as more budget becomes available.

Recently, Neural Architecture Search (NAS) has offered a principled approach to automate DNN architecture design and shows promise in other modalities such as vision and language~\cite{JiZY025,kangrevisiting,wang2020textnas}.
However, applying NAS to tabular prediction while delivering progressive improvements faces fundamental challenges:
First, existing tabular NAS methods largely rely on training-based architecture evaluations, requiring the training of a large number of candidate architectures over many iterations~\cite{yang2022tabnas}.
This makes the search process computationally expensive and undermines the progressive property in practice.
Second, training-free evaluation via \tfmem[s] has shown promise in vision tasks by efficiently estimating architecture performance from initialization statistics without full training~\cite{JiZY025,kangrevisiting}.
Yet, to the best of our knowledge, its effectiveness on tabular data remains unexplored, and no tabular-specific \tfmem has been designed.
This gap is non-trivial: unlike images or texts, which possess inherent spatial or sequential structures, tabular data is characterized by heterogeneous features and a lack of structural priors, making it far from trivial to capture complex, non-intuitive feature interactions through simple architectural statistics at initialization~\cite{BorisovLSHPK24,kangrevisiting}.
Further, in vision tasks, \tfmem[s] are often used in relatively static roles (e.g., to pre-train or warm-start search strategies)~\cite{naswot,zero-cost}, when directly transferred to tabular NAS, their inherent estimation noise makes such static usage insufficient to deliver reliable progressive improvements.
Additionally, the absence of a NAS benchmark dataset dedicated to tabular data further hinders systematic and reproducible research.

To this end, we present \textbf{\framework}, a \emph{progressive NAS \atlasname tailored for tabular data}.
\framework follows a \scheme: it first performs a \filter phase to efficiently explore a large set of candidate architectures and shortlist promising architectures using a tabular-specific \tfmem \newtfmem, and then enters a \refine phase to accurately identify the best-performing model among the shortlisted candidates based on more expensive training-based architecture evaluation.
A budget-aware coordinator holistically allocates budget across the two phases, enabling \framework to continuously improve search quality as more budget becomes available.
In summary, this paper makes the following contributions:

\begin{itemize}
\item We introduce \nasbench, the first benchmark dataset for tabular NAS, featuring over 160K unique architectures evaluated across multiple real-world datasets to facilitate reproducible research.

\item Based on \nasbench, we analyze and benchmark existing state-of-the-art \tfmem[s] from vision tasks on tabular datasets, and introduce the first tabular-specific \tfmem, \newtfmem, which captures both expressivity and trainability of architectures.

\item We propose \framework, a progressive NAS \atlasname for tabular data that introduces a novel \scheme combining the benefits of training-free and training-based architecture evaluation.

\item Extensive evaluations demonstrate that \framework achieves up to an $82.75\times$ speedup in search efficiency and outperforms recently proposed deep tabular models and TFMs across 8 benchmark datasets. 

\end{itemize}

\framework is the core algorithm behind the model selection operator in NeurDB~\cite{ooi2024neurdb,neurdb2025cidr,DBLP:journals/pvldb/XingCCLOP24}.

The rest of the paper is organized as follows: Section~\ref{sec:notation} introduces notations, 
Section~\ref{sec:method} details the methodology, 
Section~\ref{sec:experiment} presents the experiments, and
Section~\ref{sec:conclusion} concludes the paper.
Related work is presented in Appendix~\ref{app:related_work}.

\section{Notation and Terminology} \label{sec:notation}

A typical NAS approach consists of three key components: a search space, a search strategy, and an architecture evaluation mechanism~\cite{nas1000,ren2021comprehensive}. 

\highlight{Search Space} $\mathcal{A}=\{a\}$ is a collection of architectures, where each $a$ has a unique topology.
Typically, $\mathcal{A}$ is characterized by multiple design dimensions, including depth (the number of layers), width (the number of channels or hidden units per layer), and macro-topology (the connectivity pattern between layers)~\cite{nb101,siems2020bench}.
For a DNN-style search space with a fixed macro-topology, each $a$ is parameterized by the widths of $L$ hidden layers, where each width is chosen from the set of layer sizes $\mathcal{H}$. 
The search space therefore contains $|\mathcal{H}|^{L}$ architectures.
For example, a search space $\mathcal{A}$ with $L=6$ and $\mathcal{H}={8,16,32,64,128}$ contains $5^{6}=15{,}625$ architectures.

\highlight{Search Strategy} is responsible for proposing a candidate architecture $a_{i+1}$ for evaluation from the search space,
denoted as $a_{i+1} = f_s(\mathcal{A}, \mathcal{S}_i)$, where $\mathcal{S}_i$ represents the state of the search strategy at the $i$-th iteration.
Its objective is to efficiently explore the search space by evaluating promising architectures~\cite{resourcenas, once4all, yang2022tabnas}.
Popular search strategies include random sampling~\cite{bergstra2012random}, reinforcement learning~\cite{rlnas}, evolutionary algorithms~\cite{nb101,reg_evo}, and Bayesian optimization with HyperBand~\cite{bohb}.

\highlight{Architecture Evaluation} refers to assessing the performance of a given architecture.
Training-based methods evaluate architectures after full training~\cite{rlnas}, while training-free evaluation estimates performance by computing architecture statistics at initialization from a small data batch~\cite{synflow,JiZY025,kangrevisiting}.
Given $a$ with parameters $\boldsymbol{\theta}$ and a batch $X_B$ of $B$ samples, a \tfmem computes a score $s_a = \rho(a, \boldsymbol{\theta}, X_B)$, where $\rho(\cdot)$ is the assessment function.

\section{Methodology}
\label{sec:method}

\framework is structured into two phases: the \textit{\filter phase} and the \textit{\refine phase}, based on training-free and training-based architecture evaluation, respectively, and optimized holistically via a budget-aware coordinator to support progressive NAS.
In the \filter phase, \framework efficiently explores the search space, directed by a search strategy using our new \tfmem.
Next, in the \refine phase, \framework evaluates the most promising architectures accurately via training-based evaluation.
A coordinator is also introduced to guide the two phases, ensuring \framework delivers a high-performing architecture within the time budget \budget.

To ensure fair and consistent benchmarking across different NAS approaches, we first establish \nasbench (Sec.~\ref{sec:space}).
Next, we analyze \tfmem[s] in terms of trainability and expressivity, and develop a new training-free proxy for tabular architectures (Sec.~\ref{sec:ptproxy}).
We then use this proxy for coarse-grained architecture filtering in the \filter phase (Sec.~\ref{sec:filter_phase}) and employ a scheduling algorithm to optimize the \refine phase (Sec.~\ref{sec:refine_phase}).
Finally, we introduce a budget-aware coordinator to facilitate progressive NAS (Sec.~\ref{sec:coord}).

\subsection{NAS-Bench-Tabular Design}
\label{sec:space}

We build \nasbench as a tabular NAS benchmark dataset by
(i) defining a discrete search space that contains diverse possible architectures, and 
(ii) precomputing the fully-trained performance of each architecture on multiple tabular datasets, enabling NAS approaches to directly query per-dataset architecture performance without training.

\begin{figure}[t]
  \begin{center}
    \includegraphics[width=1\columnwidth]{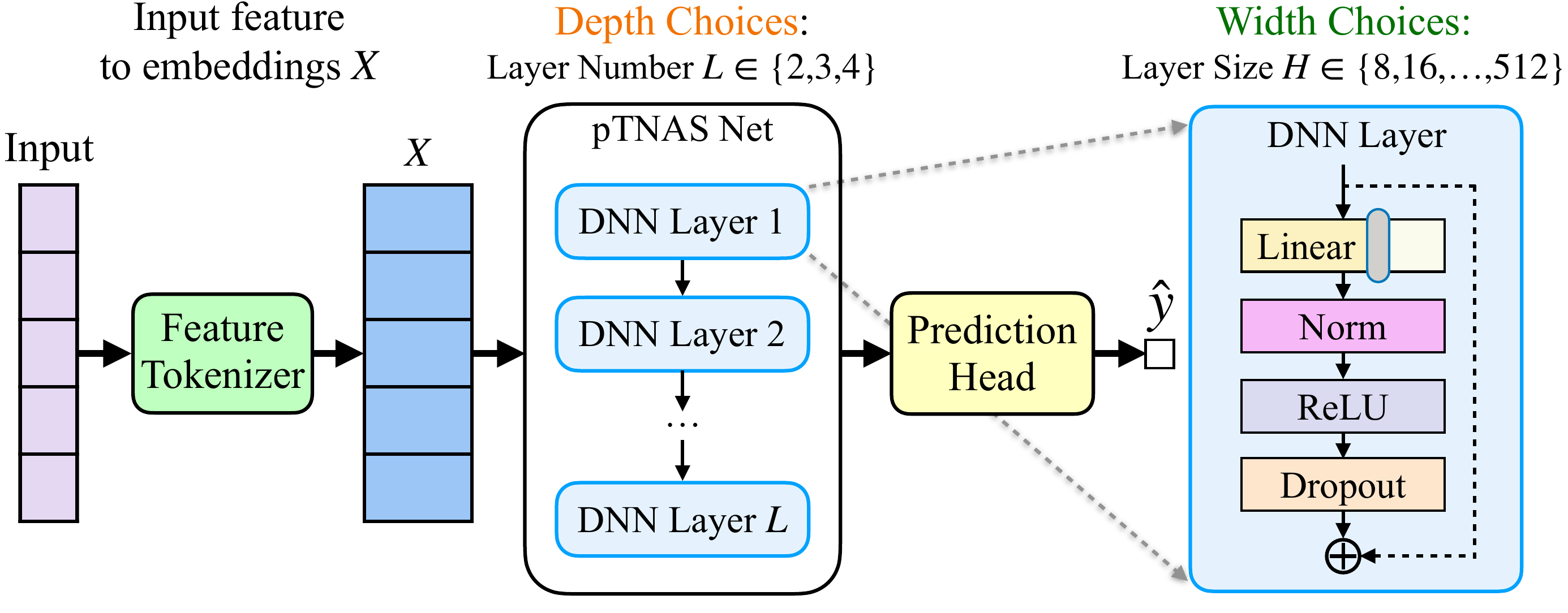}
  \end{center}
  \caption{Search Space in \framework.}
  \label{fig:space}
\end{figure}

Existing studies show that lightweight DNNs with modest depth and properly tuned hidden sizes could already achieve top performance on tabular data~\cite{HolzmullerGS24,yang2022tabnas,kadra2021well}, and recent work~\cite{abs-2511-08667,MuellerC025} further shows that even TFMs can be distilled into compact MLPs that preserve most of the accuracy while achieving much lower latency. 
Therefore, we employ a DNN backbone to construct the tabular search space, parameterized by $L$ hidden layers and a candidate set of layer sizes $\mathcal{H}$, as shown in Figure~\ref{fig:space}.
Each DNN layer consists of a linear transformation, batch normalization, ReLU activation, and dropout as described in Equation~\ref{eq:space}.
The search goal is then to determine the layer number and the size of each layer.
\begin{align}
\text{pTNAS}(X) &= \text{Pred} \left( h_L( \cdots (h_1 (X) \right))) \notag \\
h_\ell(x) &= \text{Dropout} \left( \text{ReLU} \left( \text{Norm} \left( \text{Linear}(x) \right) \right) \right) \notag \\
\text{Pred}(x) &= \text{Linear} \left( \text{ReLU} \left( \text{Norm}(x) \right) \right)
\label{eq:space}
\end{align}

We adopt three widely benchmarked tabular datasets: Frappe, Diabetes, and Criteo~\cite{yang2022tabnas}.
These datasets are well-known and commonly used tabular datasets~\cite{arm_net,regluo2} covering diverse domains (e.g., app recommendation, healthcare) with substantially different scales (about 100K to 46M samples) and feature space (369 to 5,382 features).
More details, analysis, and discussions of \nasbench are provided in Appendix~\ref{app:nasbench}.
We also include a more heterogeneous space composed of MLP, attention, transformer, and residual blocks, each with its own width. Please refer to Appendix~\ref{app:new_search_spaces} for details.

\begin{figure*}[t]
  \begin{center}
    \includegraphics[width=\textwidth]{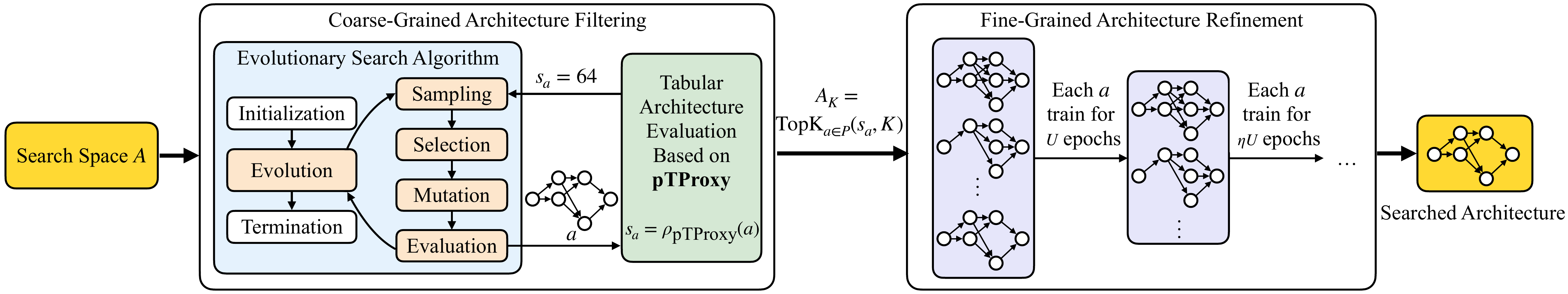}
  \end{center}
  \caption{The Overview of \framework.}
  \label{fig:system}
  \vspace{-4 mm}
\end{figure*}

\subsection{\newtfmem: A Tabular-Specific Zero-Cost Proxy}
\label{sec:ptproxy}

Theoretically, \tfmem[s] characterize two key properties of the architecture that are related to its performance:
\textit{trainability}~\cite{trainabilitypreserving,shin2020trainability,tenas} and \textit{expressivity}~\cite{hornik1989multilayer,hierarchyexpressivity,raghu2017expressive}.

\noindent
\highlight{Trainability} measures an architecture's optimization efficiency via gradient descent, where its state at initialization often determines final performance. 
Since individual DNN parameters vary in their contribution to task learning, characterizing trainability requires aggregating these parameter-wise importance scores.
Specifically, \textit{synaptic saliency}~\cite{synflow,snip} defines the importance of each parameter as $\Phi(\boldsymbol{\theta}) = f(\frac{\partial \mathcal{L}}{\partial \boldsymbol{\theta}})\bigodot g(\boldsymbol{\theta})$,
where $\mathcal{L}$ is the loss function and $\bigodot$ denotes the Hadamard product.
Consequently, the trainability of an architecture can be characterized by the global aggregation of these scores~\cite{synflow,naswot}: 
\begin{equation}
s_{\text{trainability}} \propto \sum_{\boldsymbol{\theta}_i} \Phi(\boldsymbol{\theta}_i) 
=\sum_{\boldsymbol{\theta}_i} f(\frac{\partial \mathcal{L}}{\partial \boldsymbol{\theta_i}})\bigodot g(\boldsymbol{\theta_i})
\end{equation}

\noindent
\textbf{Expressivity} characterizes the hypothesis space that an architecture can realize, specifically, the complexity of the input-to-output mappings it is capable of approximating~\cite{expressmeettrain}. 
For tabular DNNs structured as sequences of stacked nonlinear transformations, expressivity is fundamentally governed by architectural depth and width~\cite{complexirtyofnn}. 
Increased depth facilitates the hierarchical composition of successive nonlinearities, while greater width expands the latent dimensionality at each transformation step. 
Consequently, architectures with a larger number of layers $L$ and expanded hidden dimensions within the candidate set $\mathcal{H}$ 
can represent more complex nonlinear functions~\cite{expressmeettrain,complexirtyofnn}. 
Formally, we define the expressivity as:
\begin{equation} 
s_{\text{expressivity}} = \Psi(\{h_l\}_{l=1}^L), \quad h_l \in \mathcal{H}
\end{equation}
where $L$ denotes the network depth, $\mathcal{H}$ represents the set of hidden layer sizes, and $\Psi$ models the inter-layer interactions, capturing how the sequence of transformations collectively defines the DNN's representation.

\noindent
\highlight{\newtfmem Design.}
To characterize architecture performance in terms of both trainability and expressivity, we propose \newtfmem for efficient tabular DNNs evaluation.
\newtfmem is based on \textit{neuron saliency}, a more effective characterization of architecture performance.
For the $n$-th neuron in the DNN, we quantify its saliency in the architecture, denoted as $\nu_n$.
Specifically, this is computed as the product of the absolute value of the derivative of $\mathcal{L}$ with respect to the activated output of the neuron $z_n$, and the value of $z_n$ itself, i.e., $\nu_n = |\frac{\partial \mathcal{L}}{\partial z_n}| \bigodot z_n$, where $z_n = \sigma(\mathbf{w}\mathbf{x} + b)$, and $\mathbf{w}$ represents the incoming weights of the neuron, $\mathbf{x}$ is the neuron inputs, $b$ is the bias, and $\sigma$ is the activation function.
For the ReLU activation function, $\nu_n = |\frac{\partial \mathcal{L}}{\partial z_n}| \bigodot z_n$ if $z_n > 0$, otherwise $\nu_n = 0$.

The raw $\nu_n$ are not naturally comparable across different depths and widths. 
For depth, a direct summation of $\nu_n$ across layers is inherently biased toward deeper representations. 
Prior work shows that small perturbations introduced at an earlier layer can be amplified by subsequent transformations, and this amplification grows with the remaining depth of the network~\cite{raghu2017expressive}. 
This rapid amplification causes $\nu_n$ in later layers to appear larger, thereby dominating the aggregated score and masking contributions from earlier neurons. 
To mitigate such bias, we first estimate a layer-wise amplification factor via finite differences:
\begin{equation}
d_l
\;=\;
\frac{\left\lVert z^{(l)}(x+\epsilon\,\delta)-z^{(l)}(x)\right\rVert_2}{\epsilon}
\label{eq:layer_sensitivity}
\end{equation}
where $z^{(l)}(x)$ is the activation vector of layer $l$, $\delta$ is sampled from a standard Gaussian distribution with the same shape as $x$, and the perturbed input is $x+\epsilon\delta$. 
$d_l$ denotes layer sensitivity, which typically scales with depth~\cite{raghu2017expressive}. 
Then, we recalibrate the neuron saliency $\nu_n$ at layer $l$ by the inverse factor $1/d_l$. 
This normalization prevents late-layer saliency from dominating the aggregated score, thereby preserving early-layer contributions.
For width, a layer’s influence reflects not only the saliency of individual neurons but also the dimensionality of the feature space it spans.
Following~\cite{expressmeettrain}, we scale the recalibrated $\nu_n$ by the layer width $\mathcal{K}_l$. 
Intuitively, given equal per-neuron saliency, a wider layer provides a larger representational subspace and thus a greater cumulative impact. 
Multiplying by $\mathcal{K}_l$ ensures this dimensional bandwidth is captured in the final aggregated score.

Altogether, the recalibration weight for $\nu_n$ at layer $l$ is $\frac{\mathcal{K}_l}{d_l}$.
We finally perform a weighted aggregation of neuron saliency to derive a proxy score that depicts the performance of an architecture $a$ on a batch of data $X_B$, as follows:
\begin{align}
\label{eq:tfmem}
s_a
&= \rho_{\text{\newtfmem}}(a; X_B) && \notag \\
&= \sum_{i=1}^{B}\sum_{n=1}^{N} \frac{\mathcal{K}_{l(n)}}{d_{l(n)}} \,\nu_{in} && \notag \\
&= \sum_{i=1}^{B}\sum_{l=1}^{L} \frac{\mathcal{K}_l}{d_l}
\left(\sum_{n=1}^{N_l}\left|\frac{\partial \mathcal{L}}{\partial z_{in}}\right|\odot z_{in}\right) && 
\end{align}

\noindent
where $\nu_{in}$ and $z_{in}$ are the neuron saliency and activated output of the $n$-th neuron computed on the $i$-th sample, respectively,
$N$ denotes the total number of neurons in architecture, and $N_l$ is the number of neurons in the $l$-th layer.

\begin{algorithm}[t]
\caption{The Overall Workflow of \framework}
\label{alg:framework}
\small
\setlength{\baselineskip}{1.3\baselineskip} 
\SetKwInOut{Input}{Input}
\SetKwInOut{Output}{Output}
\SetKwBlock{PhaseI}{Coarse-Grained Architecture Filtering}{}
\SetKwBlock{PhaseII}{Fine-Grained Architecture Refinement}{}

\Input{Search space $\mathcal{A}$; budget $T_{\max}$; mini-batch $X_B$; \\ SH params $(U,\eta)$.}
\Output{Best-performing architecture $a^\star$.}

\BlankLine
\cmt{Budget allocation based on Eq.~\ref{eq:nk_tradeoff}}\; 

Choose $(M,K)$\;
\text{s.t.}\quad $T_1(M)+T_2(K,U)\le T_{\max}$\;

\BlankLine
\PhaseI{

  $\mathcal{P}\leftarrow$ InitPopulation$(\mathcal{A}, \rho_{\text{pTProxy}}(\cdot; X_B))$\; 

  \cmt{Evolutionary search guided by \newtfmem (Eq.~\ref{eq:tfmem})}\; 
  
  \While{$|\mathcal{P}| < M$}{
     $\mathcal{S}\leftarrow$ SampleSubset$(\mathcal{P})$\; 
     
     $a\leftarrow\arg\max\limits_{a\in \mathcal{S}} s_a$\; 
     
     $a' \leftarrow$ Mutate$(a)$\; 
     
     $s_{a'} \leftarrow \rho_{\text{pTProxy}}(a'; X_B)$\; 
     
     $\mathcal{P}\leftarrow$ UpdatePopulation$(\mathcal{P}, (a',s_{a'}))$\; 
  }
  $\mathcal{A}_K \leftarrow$ TopK$(\mathcal{P}, s_a, K)$\;
}

\BlankLine
\PhaseII{

  $U_{\text{cur}}\leftarrow U$; 
  
  $\mathcal{C}\leftarrow \mathcal{A}_K$\; 
  
  \cmt{Budget scheduling based on SH}\; 
  
  \While{$|\mathcal{C}|>1$}{
    \ForEach{$a\in \mathcal{C}$}{
        Train$(a, U_{\text{cur}})$\; 
        
        $p_a \leftarrow \mathcal{V}(a)$\;
    }
    $\mathcal{C}\leftarrow$ TopK$(\mathcal{C}, p_a, \left\lceil |\mathcal{C}|/\eta \right\rceil)$\;
    
    $U_{\text{cur}}\leftarrow U_{\text{cur}}\cdot \eta$\;
  }
  $a^\star \leftarrow$ the remaining architecture in $\mathcal{C}$\;
}

\Return $a^\star$\;
\end{algorithm}

\noindent
\highlight{Tabular-Specific Design.}
\newtfmem is specifically designed for tabular data.
Unlike vision tasks that rely heavily on spatial priors such as translation invariance, tabular learning requires modeling implicit and high-order interactions among heterogeneous features~\cite{song2019autoint,xie2021fives}.
Neurons serve as the basic feature-extraction units for modeling such interactions in tabular DNNs~\cite{tabulartransfer,arm_net}, combining signals from multiple input features and acting as local nonlinear detectors of useful feature patterns.
Therefore, \newtfmem operates at the neuron level rather than the parameter level used by many existing \tfmem[s].
Concretely, \newtfmem captures both trainability and expressivity through three mechanisms.
(1) Neuron saliency considers the activation value of neurons.
By computing the activated neuron output $z_n$, neuron saliency captures complex and non-intuitive relationships among input features in tabular data.
(2) Neuron saliency also accounts for neuron-level derivatives.
A larger gradient of the loss with respect to the activation $z_n$, i.e., $\frac{\partial \mathcal{L}}{\partial z_n}$, indicates higher importance of the features extracted by this neuron, and therefore greater significance for the prediction task.
The absolute value in $\nu_n$ prevents positive and negative neuron-saliency contributions from canceling each other, ensuring that important neuron contributions are retained in the aggregated score.
(3) Neuron saliency's recalibration weight, i.e., $\frac{\mathcal{K}_l}{d_l}$, is determined by the depth and width of its layer.
Neurons in earlier and wider layers receive higher saliency values, highlighting their larger influence on architecture performance.
More in-depth theoretical analysis of \newtfmem regarding its trainability and expressivity is provided in Appendix~\ref{app:theoretical_analysis}.

\subsection{Coarse-Grained Architecture Filtering}
\label{sec:filter_phase}

While \newtfmem enables fast architecture evaluation (within seconds), exhaustively scoring the entire search space $\mathcal{A}$ is computationally prohibitive.  
To address this, we adopt an evolutionary search algorithm (EA)~\cite{nb101,reg_evo} to explore the discrete search space efficiently.
EA maintains a population of architectures and iteratively samples a parent subset, selects the highest-scoring architecture based on \newtfmem ($s_a = \rho_{\text{\newtfmem}}(a; X_B)$), and applies a mutation to generate a new candidate, where each mutation randomly selects one layer and replaces its width with another candidate value from the predefined width set. 
For the BlockMixed space in Appendix~\ref{app:new_search_spaces}, each mutation randomly inserts or deletes a block, or changes one block attribute, including block type, width, normalization, activation, dropout, connectivity, or the auxiliary parameter of attention/transformer blocks.
Each new candidate is evaluated using \newtfmem and added to the evaluated candidate set.
This evolutionary process continues until the target number of architectures ($M$) has been explored, as shown in Algorithm~\ref{alg:framework}. 
Here, $M$ is determined by the coordinator (Section~\ref{sec:coord}). 
The procedure then yields a ranked set of evaluated candidates.
The top-$K$ architectures among these candidates are then passed to the \refine for training-based evaluation.

\subsection{Fine-Grained Architecture Refinement}
\label{sec:refine_phase}
To mitigate the estimation gap introduced by the training-free proxy \newtfmem, \framework performs a fine-grained training of the top-$K$ architectures $\mathcal{A}_K$.  
To reduce the computational cost of fully training all $K$ candidates, we adopt a budget-aware scheduling strategy based on Successive Halving (\succhalf)~\cite{SHALG}.  
\succhalf begins by assigning a small, equal budget to all candidates. 
In each round, the lower-performing fraction is discarded, while the remaining architectures receive increased training budgets.  
This process continues until a single best-performing architecture is identified.  
By progressively allocating budget, \succhalf enables early elimination of weak candidates and concentrates more budget on high-potential candidates.

\subsection{Progressive Neural Architecture Search}
\label{sec:coord}
As illustrated in Figure~\ref{fig:system}, \framework operates in a \scheme: a \filter phase for efficient exploration over the search space $\mathcal{A}$, and a \refine phase for effective exploitation of top-$K$ promising candidates.
A budget-aware coordinator is introduced to holistically allocate resources across the two phases, to maximize architecture performance under a total time budget \budget, as detailed in Algorithm~\ref{alg:framework}.

To formalize this optimization, we quantify the time of each phase.  
Let $t_1$ and $t_2$ denote the time required to evaluate an architecture using \newtfmem and to train for one epoch, respectively.  
Exploring $M$ architectures in the \filter phase incurs a total time of $T_1 = M t_1$.

\begin{table*}[t]
    \centering
    \caption{\textbf{SRCC summary of \tfmem[s] across benchmark datasets.}
    For each proxy, we report the mean and standard deviation of SRCC computed over these datasets.
    We also report the average rank, where methods are ranked \emph{per dataset} by descending $|\mathrm{SRCC}|$ (higher is better), and the ranks are then averaged across datasets.}
    
    \renewcommand{\arraystretch}{1}
    \setlength\tabcolsep{1mm}
    \begin{tabular}{c|ccccccccc|c}
    \toprule[1.5pt]
        & GradNorm & NASWOT & NTKCond & NTKTrace & NTKTrAppx & Fisher & GraSP & SNIP & SynFlow & \noindent\textbf{\newtfmem}\\
        \midrule[0.5pt]
        Mean SRCC & 0.40 & 0.65 & -0.66 & 0.47 & 0.15 & 0.38 & -0.24 & 0.70 & 0.74 & \noindent0.82 \\
        Std. SRCC & 0.05 & 0.05 & 0.11 & 0.08 & 0.16 & 0.14 & 0.04 & 0.09 & 0.05 & \noindent 0.08 \\
        Avg. Rank & 7.3 & 4.0 & 4.3 & 6.3 & 9.3 & 8.0 & 9.0 & 3.3 & 2.3 & 1.0 \\
    \bottomrule[1.5pt]
    \end{tabular}
    \label{table:tfmem_srcc}
\end{table*}

In the \refine phase, \succhalf progressively narrows the top-$K$ candidate set by retaining only the top $1/\eta$ fraction after each round.  
Initially, each architecture is trained for $U$ epochs using a total budget of $KUt_2$.  
At each round, the training budget per architecture is increased, and the number of candidates is reduced by a factor of $\eta$, ensuring that every round consumes the same total budget.  
Given approximately $\lceil\log_\eta K\rceil$ rounds, the total time usage is $T_2 \approx KUt_2 \lceil\log_\eta K\rceil$.

The overall optimization can therefore be formally summarized as the following constrained maximization problem:
\begin{align}
\label{eq:nk_tradeoff}
\max_{\Omega}\quad 
& p = \mathcal{V}\!\left( \mathcal{S}_{\text{refinement}} \circ \mathcal{S}_{\text{filtering}} (\mathcal{A}; \Omega) \right) \notag \\
\text{s.t.}\quad 
& T_1(M) + T_2(K, U) \le T_{\text{max}} \notag \\
\text{where}\quad 
& T_1 = M\, t_1,
T_2 \approx K\,U\,t_2\,\lceil \log_{\eta} K \rceil
\end{align}
where $\Omega=(M,K)$, $\mathcal{S}_{\text{filtering}}$ and $\mathcal{S}_{\text{refinement}}$ are selection operators that progressively map the search space $\mathcal{A}$ to the final architecture, while $\mathcal{V}$ is the function that measures the validation performance of the selected architecture.

In practice, $\eta$ controls how aggressively architectures are pruned in each round. 
We use the default setting $\eta=3$, following the original Hyperband paper~\cite{hyperband}. 
To balance the filtering and refinement phases, we further study the sensitivity of $M/K$ and $U$ to the final architecture performance, and set them empirically based on this analysis (Appendix~\ref{app:abs_coord}).
The coordinator therefore determines $M$ and $K$ for any predefined \budget.

\section{Experiments}
\label{sec:experiment}

\subsection{Effectiveness of \newtfmem}
\label{sec:tfmem_effectiveness}

\begin{figure*}[t]
\centering

\begin{subfigure}[t]{\textwidth}
  \centering
  \includegraphics[width=0.6\textwidth]{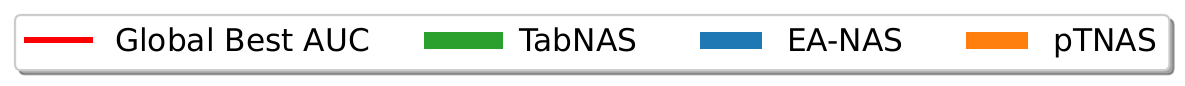}
\end{subfigure}

\begin{subfigure}[t]{0.3\textwidth}
  \centering
  \includegraphics[width=\linewidth]{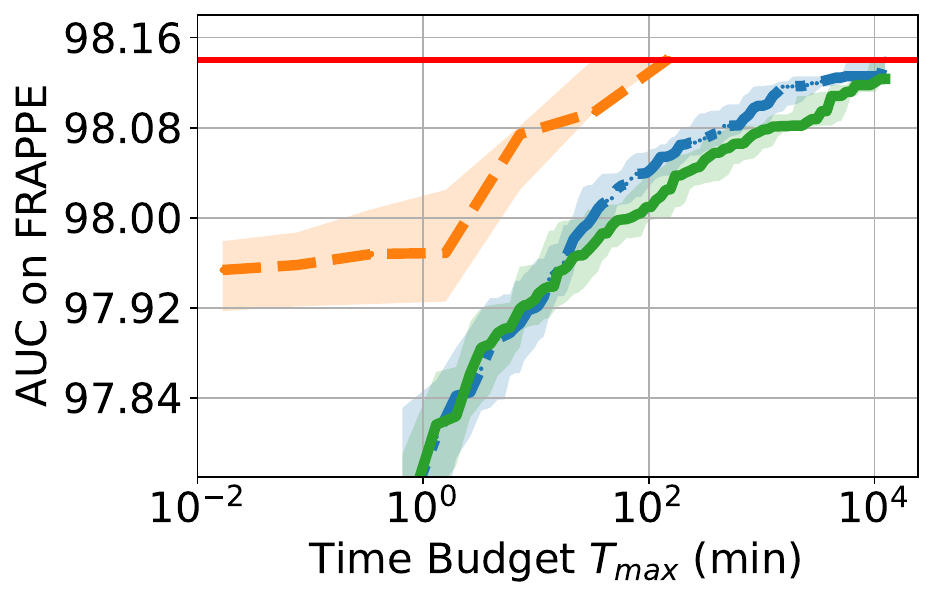}
  \caption{Frappe Dataset.}
\end{subfigure}\hfill
\begin{subfigure}[t]{0.3\textwidth}
  \centering
  \includegraphics[width=\linewidth]{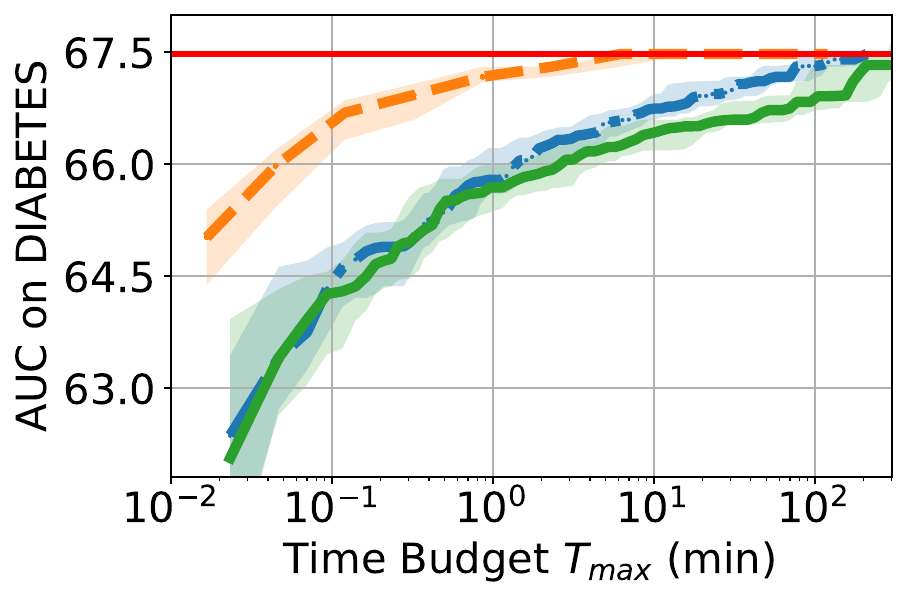}
  \caption{Diabetes Dataset.}
\end{subfigure}\hfill
\begin{subfigure}[t]{0.3\textwidth}
  \centering
  \includegraphics[width=\linewidth]{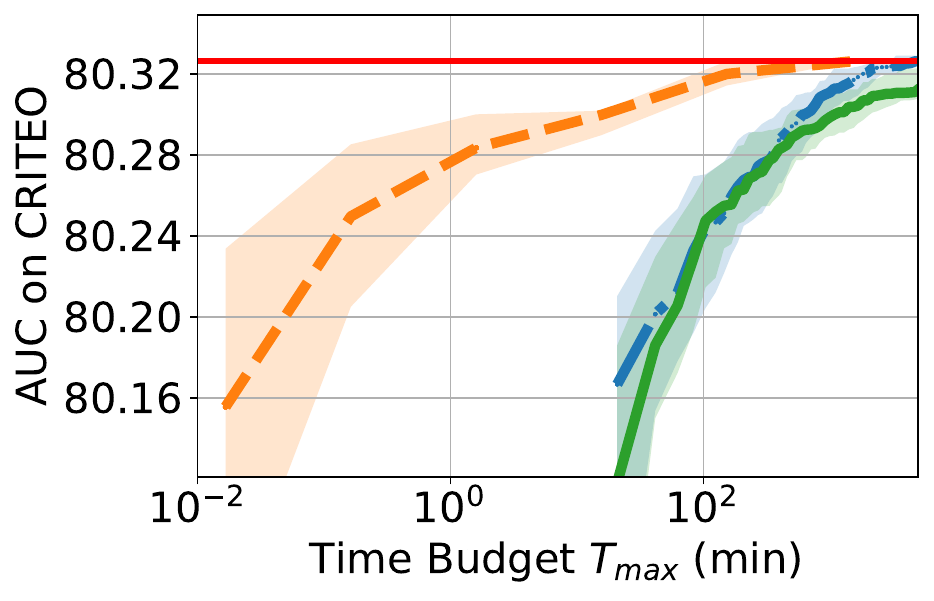}
  \caption{Criteo Dataset.}
\end{subfigure}

\caption{Progressive search performance of \framework compared with EA-NAS and TabNAS.}
\label{fig:search_performance}
\end{figure*}

\begin{table*}
  \centering
\caption{\textbf{Effectiveness evaluation of \framework against existing methods on eight datasets.}
For each dataset, methods are first ranked by their raw MAE/AUC values. We report the average rank within each task type, and overall across all available datasets for each method, where a lower rank indicates better performance. We also report an average normalized score to reflect how close a method is to the best result on each dataset: for regression, the score is $\text{best MAE}/\text{MAE}$; for classification, it is $(\text{AUC}-0.5)/(\text{best AUC}-0.5)$, where a higher value is better. The two statistics are complementary: average rank reflects relative ordering, while the normalized score reflects the magnitude of the gap to the best method.
}

  \label{tab:rel_bench_acc}
  \resizebox{0.92\textwidth}{!}{%
  \begin{threeparttable}
\begin{tabular}{@{}c|l|cc|cc|cc@{}}
  \toprule[1.5pt]
  \multicolumn{2}{c|}{Task/Metric} &
    \multicolumn{2}{c|}{Regression Task} &
    \multicolumn{2}{c|}{Classification Task} &
    \multicolumn{2}{c}{Overall} \\ \cmidrule(lr){1-2} \cmidrule(lr){3-4} \cmidrule(lr){5-6} \cmidrule(l){7-8}
  Type & Method &
  \begin{tabular}[c]{@{}c@{}}Avg.\\Rank ($\downarrow$)\end{tabular} &
  \begin{tabular}[c]{@{}c@{}}Avg. Norm.\\Score ($\uparrow$)\end{tabular} &
  \begin{tabular}[c]{@{}c@{}}Avg.\\Rank ($\downarrow$)\end{tabular} &
  \begin{tabular}[c]{@{}c@{}}Avg. Norm.\\Score ($\uparrow$)\end{tabular} &
  \begin{tabular}[c]{@{}c@{}}Avg.\\Rank ($\downarrow$)\end{tabular} &
  \begin{tabular}[c]{@{}c@{}}Avg. Norm.\\Score ($\uparrow$)\end{tabular} \\ \midrule
  
  \multicolumn{1}{c|}{\multirow{4}{*}{CM}} &
  \multicolumn{1}{l|}{LR~\cite{cox1958regression}} &
  12.25 & 0.7959 &
  11.25 & 0.8694 &
  11.75 & 0.8326 \\
  \multicolumn{1}{c|}{} &
  \multicolumn{1}{l|}{RF~\cite{Breiman01}} &
  10.63 & 0.8338 &
  12.75 & 0.8485 &
  11.69 & 0.8411 \\
  \multicolumn{1}{c|}{} &
  \multicolumn{1}{l|}{CatBoost~\cite{ProkhorenkovaGV18}} &
  5.50 & 0.9246 &
  7.00 & 0.9105 &
  6.25 & 0.9176 \\
  \multicolumn{1}{c|}{} &
  \multicolumn{1}{l|}{LightGBM~\cite{KeMFWCMYL17}} &
  5.63 & \textbf{0.9578} &
  7.00 & 0.9125 &
  6.31 & 0.9351 \\ \midrule
  
  \multicolumn{1}{c|}{\multirow{2}{*}{TFM}} &
  \multicolumn{1}{l|}{TabPFN~\cite{tabfpn}} &
  9.88 & 0.8576 &
  5.00 & \textbf{0.9486} &
  7.44 & 0.9031 \\
  \multicolumn{1}{c|}{} &
  \multicolumn{1}{l|}{TabICL\textsuperscript{$\ast$}~\cite{tabicl}} &
  - & - &
   \textbf{4.50} & 0.9439 &
  \textbf{4.50} & \textbf{0.9439} \\ \midrule
  
  \multicolumn{1}{c|}{\multirow{4}{*}{DTM}} &
  \multicolumn{1}{l|}{DNN~\cite{LeCunBH15}} &
  5.25 & 0.9343 &
  8.25 & 0.8909 &
  6.75 & 0.9126 \\
  \multicolumn{1}{c|}{} &
  \multicolumn{1}{l|}{DeepFM~\cite{GuoTYLH17}} &
  7.50 & 0.9214 &
  10.50 & 0.8581 &
  9.00 & 0.8898 \\
  \multicolumn{1}{c|}{} &
  \multicolumn{1}{l|}{FTTrans~\cite{gorishniy2021revisiting}} &
  \textbf{4.13} & 0.9471 &
  8.00 & 0.8936 &
  6.06 & 0.9204 \\
  \multicolumn{1}{c|}{} &
  \multicolumn{1}{l|}{ARM-Net~\cite{arm_net}} &
  5.50 & 0.9450 &
  6.75 & 0.9229 &
  6.13 & 0.9339 \\ \midrule
  
  \multicolumn{1}{c|}{\multirow{3}{*}{LLM}} &
  \multicolumn{1}{l|}{TP-BERTa~\cite{YanZXZC00C24}} &
  13.75 & 0.6296 &
  16.00 & 0.0922 &
  14.71 & 0.3993 \\
  \multicolumn{1}{c|}{} &
  \multicolumn{1}{l|}{Nomic~\cite{NussbaumMMD25}} &
  12.25 & 0.6535 &
  13.75 & 0.6941 &
  13.00 & 0.6738 \\
  \multicolumn{1}{c|}{} &
  \multicolumn{1}{l|}{BGE~\cite{XiaoLZMLN24}} &
  10.50 & 0.7865 &
  13.25 & 0.7819 &
  11.88 & 0.7842 \\ \midrule
  
  \multicolumn{1}{c|}{\multirow{3}{*}{NAS}} &
  \multicolumn{1}{l|}{TabNAS~\cite{yang2022tabnas} ($T_{\text{max}}=10s$)} &
  6.50 & 0.9370 &
  6.00 & 0.9262 &
  6.25 & 0.9316 \\
  \multicolumn{1}{c|}{} &
  \multicolumn{1}{l|}{EA-NAS ($T_{\text{max}}=10s$)} &
  9.00 & 0.8488 &
  5.00 & 0.9322 &
  7.00 & 0.8905 \\
  
  \multicolumn{1}{c|}{} &
  \multicolumn{1}{l|}{\cellcolor{blue!4}\textbf{pTNAS} ($T_{\text{max}}=10s$) } &
  \cellcolor{blue!4}\best{1.75} & \cellcolor{blue!4}\best{0.9891} &
  \cellcolor{blue!4}\best{1.00} & \cellcolor{blue!4}\best{1.0000} &
  \cellcolor{blue!4}\best{1.38} & \cellcolor{blue!4}\best{0.9946} \\
    
    \bottomrule[1.5pt]
    \end{tabular}%
    \begin{tablenotes}\footnotesize
    \item $\ast$ TabICL is inherently limited to classification tasks and does not apply to regression-based tasks.
    \end{tablenotes}
    \end{threeparttable}
    }
    \end{table*}

We first benchmark nine vision-based zero-cost proxies against \framework based on \nasbench, and quantify their ranking quality by the \textit{Spearman Rank Correlation Coefficient (SRCC)} between proxy scores and the ground-truth AUC.
A robust \tfmem exhibits consistently high correlations across diverse data distributions.

As shown in Table~\ref{table:tfmem_srcc}, while NASWOT, SNIP, and SynFlow consistently achieve SRCCs above $0.6$, \newtfmem outperforms all others with an average rank of $1.0$. 
Besides, \newtfmem maintains small SRCC variance across datasets, 
which confirms its robustness and transferability in characterizing tabular architecture performance.
The superior performance of \newtfmem stems from its ability to simultaneously characterize both trainability and expressivity, as theoretically established in Section~\ref{sec:ptproxy} and Appendix~\ref{app:theoretical_analysis}.

\subsection{Progressive NAS on Tabular Data}
\label{sec:progress_nas}

We then benchmark \framework against two other NAS methods proposed for tabular data, i.e., TabNAS~\cite{yang2022tabnas} and EA-NAS (a training-based baseline using evolutionary search~\cite{nb101}). 
By varying the time budget \budget from seconds to hours, we investigate:  
(i) Efficiency: How much time does each NAS approach require to identify architectures approaching the global optimum?  
(ii) Progressive property: Can the \framework discover comparably better-performing architectures within a given budget, and does performance improve steadily as more budget becomes available?

The results are shown in Figure~\ref{fig:search_performance}. 
In terms of efficiency, \framework significantly outperforms EA-NAS, achieving speedups of $82.75\times$, $1.75\times$, and $69.44\times$ on Frappe, Diabetes, and Criteo, while reaching strong AUCs of 0.9814, 0.6750, and 0.8033, respectively.  
This advantage stems from avoiding the heavy overhead of training-based evaluations, which hinders both EA-NAS and TabNAS. 
Instead, \framework holistically optimizes training-free and training-based evaluations, enabling rapid exploration over large architecture search spaces and effective pruning of suboptimal candidate architectures.

Regarding progressive property, \framework consistently discovers better-performing architectures than EA-NAS and TabNAS under various \budget and exhibits a stable upward trajectory as \budget increases.  
Both EA-NAS and TabNAS often require 5–10 minutes to evaluate a single architecture, while \framework provides a viable architecture even under a small \budget.  
This capability is enabled by the budget-aware coordinator (Section~\ref{sec:coord}), which dynamically partitions \budget between exploration and exploitation.  
By combining fast proxy-based filtering via \newtfmem with more precise refinement, \framework maintains strong performance across all \budget.

\subsection{Comparison with Existing Tabular Models}
\label{sec:full_compare}
Moreover, we benchmark \framework against existing NAS methods and four representative categories of tabular models:
Classical Models (CM), including LR~\cite{cox1958regression}, RF~\cite{Breiman01}, CatBoost~\cite{ProkhorenkovaGV18}, and LightGBM~\cite{KeMFWCMYL17}; 
Tabular Foundation Models (TFM), including TabPFN~\cite{tabfpn} and TabICL~\cite{tabicl}; 
Deep Tabular Models (DTM), including DNN~\cite{LeCunBH15}, DeepFM~\cite{GuoTYLH17}, FTTrans~\cite{gorishniy2021revisiting}, and ARM-Net~\cite{arm_net}; 
and Large Language Models (LLM), including TP-BERTa~\cite{YanZXZC00C24}, Nomic~\cite{NussbaumMMD25}, and BGE~\cite{XiaoLZMLN24}.
For all NAS methods, including \framework, we set the search budget to $10$s based on the observed tuning and training cost of LightGBM in our comparison ($10$s in total for classification tasks), thereby ensuring a fair comparison under a matched computational budget.
We evaluate their predictive performance on eight datasets covering both classification and regression tasks. 
Detailed descriptions of these datasets are deferred to Appendix~\ref{app:dataset}.

\subsubsection{Effectiveness}

Table~\ref{tab:rel_bench_acc} summarizes the results.
Among CMs, advanced tree-based methods such as CatBoost and LightGBM clearly outperform linear models (LR) and random forests (RF), indicating that non-linear feature interactions and robust handling of categorical variables are critical. 

TFMs, including TabPFN and TabICL, perform strongly on classification tasks. 
TabPFN achieves an average normalized score close to the best results, while TabICL ranks among the top despite being limited to classification. 
However, their performance in regression is less stable and suffers from task-type restrictions, e.g., TabICL is inherently incompatible with regression tasks.

DTMs such as DNN, DeepFM, FT-Transformer, and ARM-Net demonstrate competitive performance across both task types. 
However, these models rely on fixed backbone architectures, which may not generalize optimally across diverse tabular distributions.

LLM-based approaches such as TP-BERTa, Nomic, and BGE treat tabular rows as serialized text and struggle with high-cardinality fields and input-length constraints. 
Thus, they underperform across both classification and regression.

\begin{figure}[t]
\centering
\includegraphics[width=1\columnwidth]{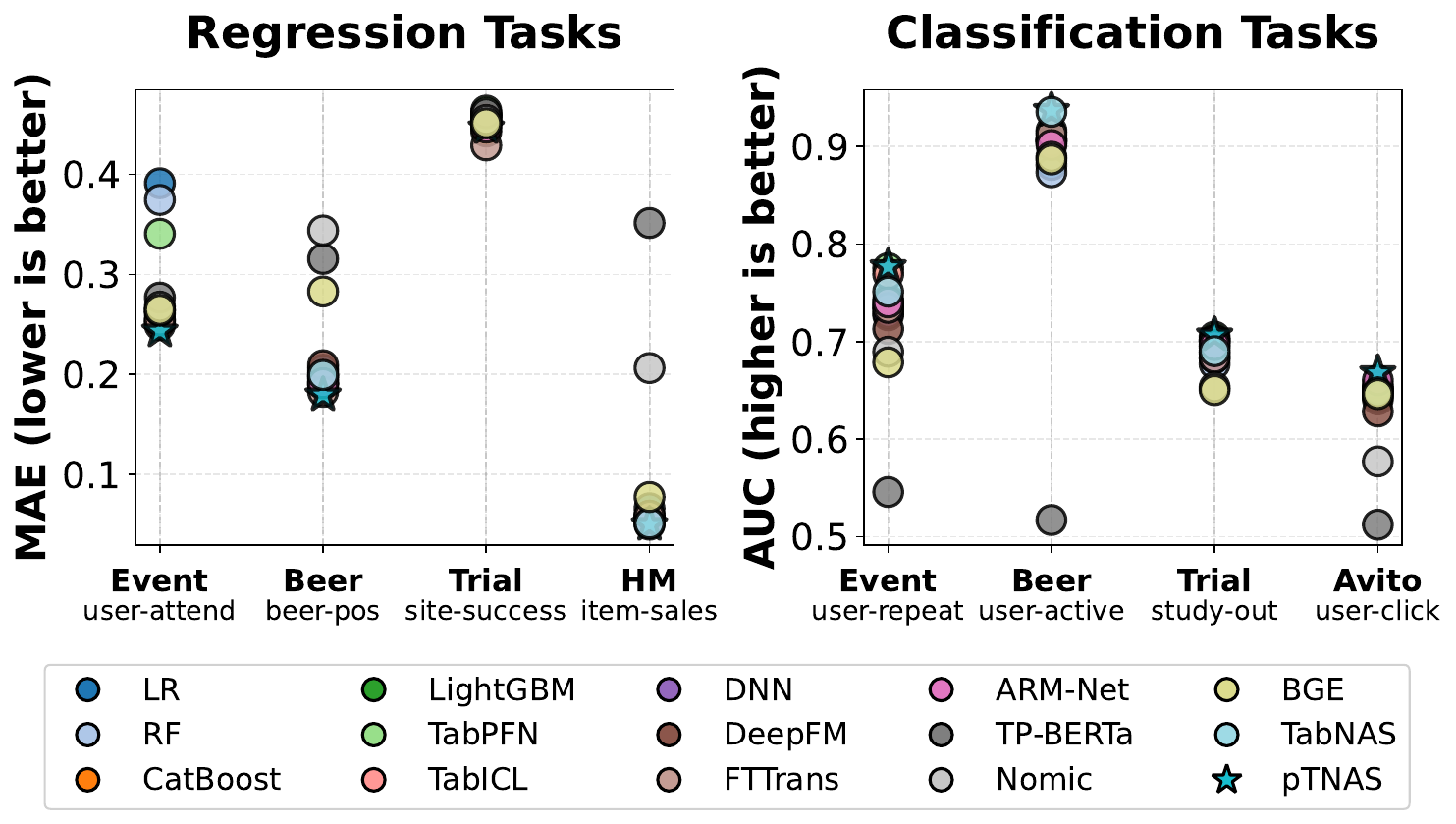}
\caption{\textbf{Per-dataset performance of all eight tabular datasets}. Each point denotes one tabular model on one dataset. In general, \framework achieves competitive performance across datasets.}
\label{fig:eff_point}
\end{figure}

TabNAS and EA-NAS perform poorly because the short budget ($10$s) allows them to explore only a limited number of architectures.
\framework achieves the best overall performance in Table~\ref{tab:rel_bench_acc}, ranking first on average across both classification and regression.
This suggests that searching for a data-specific DNN architecture can be an effective way to improve performance across heterogeneous tabular distributions.
We also visualize the per-dataset performance on all tabular datasets in Figure~\ref{fig:eff_point}. 
Clearly, \framework consistently ranks among the top-performing methods and remains very close to the best method even when it is not ranked first.

\subsubsection{Efficiency}

\begin{figure}[t]
\centering
\includegraphics[width=1\columnwidth]{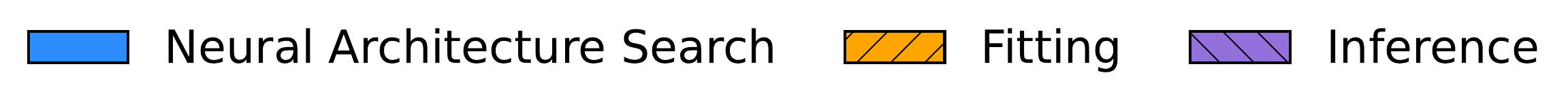}
\\
\subfloat[\textsf{Regression}]{
\label{overhead:latency}
\includegraphics[width=0.95\columnwidth]{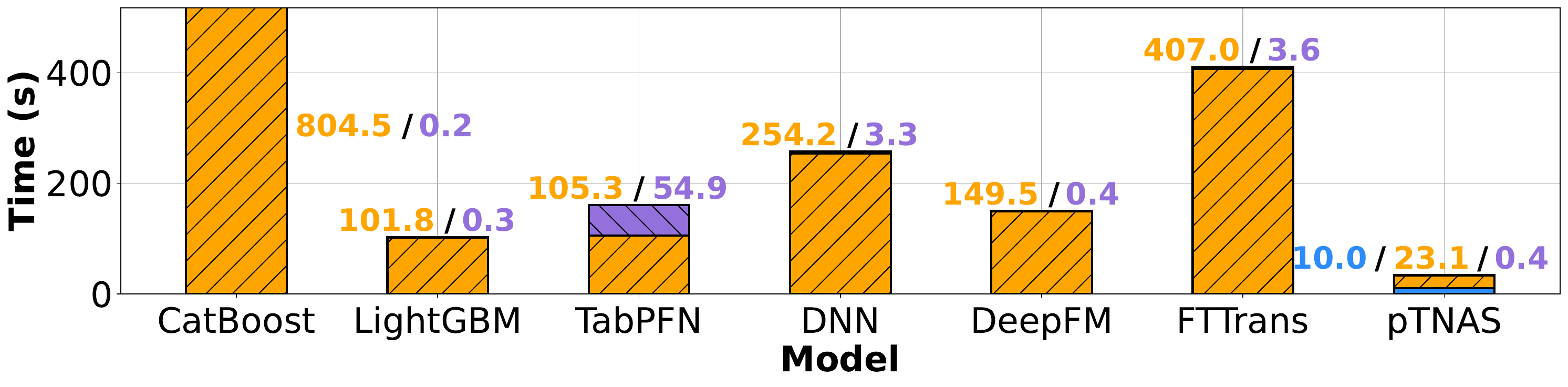}}
\\
\subfloat[\textsf{Classification}]{
\label{overhead:size}
\includegraphics[width=0.95\columnwidth]{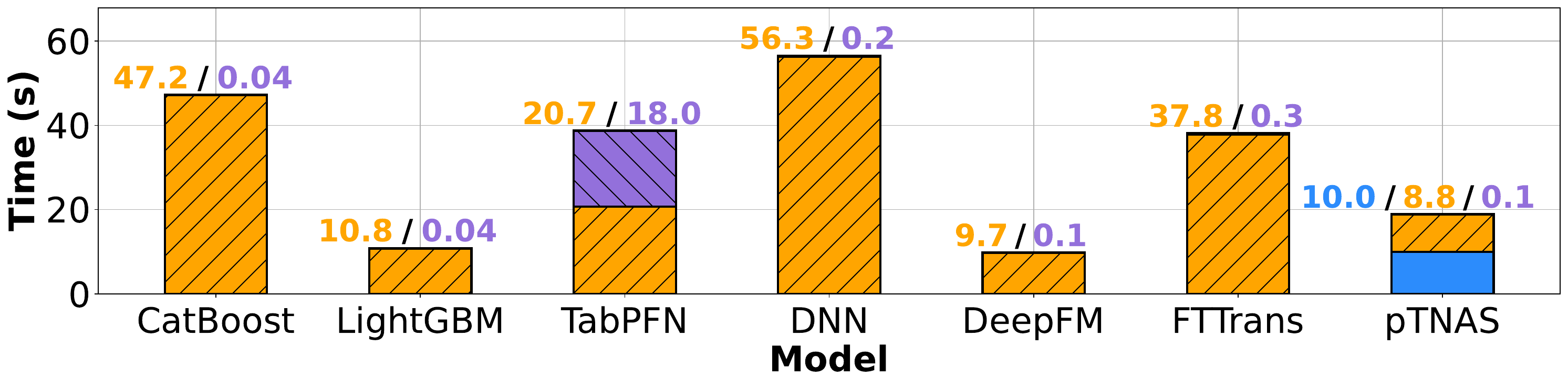}
}
\caption{\textbf{Efficiency evaluation of \framework against existing tabular models on eight datasets.}
We select representative tabular models from each model group based on their global ranking to ensure competitive baselines.
For both regression and classification tasks, we report the average runtime across all datasets in the corresponding task, decomposed into search/selection (NAS), fitting, and inference.
Fitting corresponds to context preprocessing for TFMs and full training for other models.
Overall, \framework (pTNAS) is \textbf{4.78$\times$} and \textbf{12.25$\times$} faster than TabPFN and FTTrans on regression, and \textbf{2.05$\times$} and \textbf{2.02$\times$} faster on classification, respectively (based on total time). Full results are reported in Table~\ref{tab:result_time_detail}.
}
\label{fig:rel_bench_time}
\end{figure}

The efficiency comparison of \framework is shown in Figure~\ref{fig:rel_bench_time}.
We evaluate end-to-end efficiency across three stages: neural architecture search, fitting (i.e., context preprocessing for TFM and full training for others), and inference.

The results reveal a clear trade-off between per-task adaptability and end-to-end latency.  
CMs, such as CatBoost and LightGBM, offer fast inference due to their tree-based design, but exhibit highly variable training efficiency.  
For instance, CatBoost encounters a major bottleneck on the Trial (site-success) dataset, even with early stopping.  
This inefficiency arises from recursive split-finding over complex relational structures and high-cardinality categorical features, which do not scale well during training.

TFMs, such as TabPFN, are theoretically zero-shot, but introduce substantial overhead in both the fit and inference stages.  
The fitting stage requires encoding the entire training set into a transformer context, leading to significant latency as the dataset size increases.  
More critically, TabPFN’s inference is consistently the slowest among all baselines due to expensive transformer forward passes that scale quadratically with sequence length.

In contrast, \framework implements an efficient NAS-based method that completes NAS within $10$s and produces dataset-specialized DNNs.
For example, on the Event \textit{user-attendance} dataset with $T_{\max}=10$s, the coordinator spends less than $1$s determining the time allocation based on Equation~\ref{eq:nk_tradeoff}, allocates $M=484$ for filtering ($T_1=1.6$s), and keeps $K=8$ architectures for refinement ($T_2=7.2$s).
While introducing a small search overhead, it maintains moderate fitting time and achieves inference speed comparable to optimized CM baselines.

Overall, \framework delivers a superior efficiency for tabular analytics.
All these results support our central claim: \textbf{effectively discovering dataset-specific DNN architectures is crucial for realizing competitive tabular performance without incurring foundation-scale inference costs.}

\subsection{Extended Analyses and Results}

We provide additional analyses to support the experimental results in the main text, with full details deferred to the appendix.
We begin by reporting supplementary benchmark construction and search-space details of \nasbench in Appendix~\ref{app:nasbench}.
We then present ablations for evaluating \newtfmem in Appendix~\ref{app:abs_tfmem}, including parameter positivity, initialization methods, batch size, and the recalibration weight used for aggregating neuron saliency.
In addition, we compare \newtfmem with other \tfmem[s] through correlation visualizations in Appendix~\ref{app:more_exp_visual}.

Next, we provide extended analysis of the \scheme and the budget-aware coordinator, including sensitivity to $(M, K, U)$ and coordinator variants, in Appendix~\ref{app:abs_two_phase} and Appendix~\ref{app:abs_coord}.
We further include additional baseline comparisons in Appendix~\ref{app:more_baseline} (e.g., different combinations of training-free and training-based NAS strategies, and one-shot NAS via weight sharing).
Finally, the full per-dataset effectiveness results and the corresponding efficiency measurements used to compute the aggregated statistics are reported in Appendix~\ref{app:relbench_full}.

\balance

\section{Conclusion}\label{sec:conclusion}

In this work, we introduce a progressive NAS \atlasname \framework tailored for tabular data. 
\framework equips machine learning practitioners with the capability of ascertaining high-performance architectures within any given time budget and further refining these architectures as larger time budgets are given. 
We first design a search space, denoted as \nasbench, to serve as a benchmarking platform for diverse NAS algorithms on tabular data. 
Then, we conduct an evaluation of several \tfmem[s] on tabular data and propose a tabular-specific \tfmem, \newtfmem, which characterizes both the trainability and expressivity of an architecture. 
Based on these foundations, we present \framework, which leverages the advantages of efficient training-free and effective training-based architecture evaluation through a novel \scheme with holistic optimization.
Empirical results demonstrate that \framework substantially accelerates the search for high-performing architectures compared with existing tabular NAS methods. 
Moreover, \framework outperforms recently proposed deep tabular models across multiple benchmark datasets. 
Despite these strengths, \framework currently focuses on search efficiency and predictive performance, while explicitly incorporating deployment constraints such as memory footprint remains an important direction for future work.

\section*{Acknowledgements}
We dedicate this work to the memory of Professor Beng Chin Ooi, whose wisdom, generosity, and vision have profoundly shaped our research journey. His guidance and legacy will continue to inspire us.

We thank Zhaojing Luo for his early contributions. 

This research is supported by the National Research Foundation, Singapore and Infocomm Media Development Authority under its Trust Tech Funding Initiative, and the National Natural Science Foundation of China (624B2027). Any opinions, findings and conclusions or recommendations expressed in this material are those of the author(s) and do not reflect the views of the National Research Foundation, Singapore and Infocomm Media Development Authority.

\section*{Impact Statement}
Tabular data is a core component of many real-world information systems and is commonly stored and managed in relational database management systems (RDBMSs). Improving learning methods for tabular data can therefore affect a broad range of applications, including business analytics, healthcare operations, and financial services, where structured records are prevalent.
This work contributes a method for automating tabular model design with an emphasis on computational efficiency. By reducing the need for extensive manual tuning and expensive trial-and-error search, the proposed approach can lower the barrier to obtaining competitive models under limited compute budgets. 



\bibliography{ref}

@inproceedings{MetaQNN,
  author   = {Bowen Baker and Otkrist Gupta and Nikhil Naik and Ramesh Raskar},
  title    = {Designing Neural Network Architectures using Reinforcement Learning},
  booktitle= {International Conference on Learning Representations},
  year     = {2017}
}

@inproceedings{ENAS,
  author   = {Hieu Pham and Melody Y. Guan and Barret Zoph and Quoc V. Le and Jeff Dean},
  title    = {Efficient Neural Architecture Search via Parameter Sharing},
  booktitle= {Proceedings of the International Conference on Machine Learning},
  volume   = {80},
  pages    = {4092--4101},
  year     = {2018}
}

@inproceedings{DARTS,
  author   = {Hanxiao Liu and Karen Simonyan and Yiming Yang},
  title    = {{DARTS:} Differentiable Architecture Search},
  booktitle= {International Conference on Learning Representations},
  year     = {2019}
}

@inproceedings{synflow,
  author   = {Hidenori Tanaka and Daniel Kunin and Daniel L. K. Yamins and Surya Ganguli},
  title    = {Pruning Neural Networks without Any Data by Iteratively Conserving Synaptic Flow},
  booktitle= {Advances in Neural Information Processing Systems},
  year     = {2020}
}

@inproceedings{GRASP,
  author   = {Chaoqi Wang and Guodong Zhang and Roger B. Grosse},
  title    = {Picking Winning Tickets Before Training by Preserving Gradient Flow},
  booktitle= {International Conference on Learning Representations},
  year     = {2020}
}

@inproceedings{zero-cost,
  author   = {Mohamed S. Abdelfattah and Abhinav Mehrotra and Lukasz Dudziak and Nicholas Donald Lane},
  title    = {Zero-Cost Proxies for Lightweight {NAS}},
  booktitle= {International Conference on Learning Representations},
  year     = {2021}
}

@inproceedings{naswot,
  author   = {Joe Mellor and Jack Turner and Amos J. Storkey and Elliot J. Crowley},
  title    = {Neural Architecture Search without Training},
  booktitle= {Proceedings of the International Conference on Machine Learning},
  volume   = {139},
  pages    = {7588--7598},
  year     = {2021}
}

@inproceedings{snip,
  author   = {Namhoon Lee and Thalaiyasingam Ajanthan and Philip H. S. Torr},
  title    = {Snip: single-Shot Network Pruning based on Connection sensitivity},
  booktitle= {International Conference on Learning Representations},
  year     = {2019}
}

@inproceedings{tenas,
  author   = {Wuyang Chen and Xinyu Gong and Zhangyang Wang},
  title    = {Neural Architecture Search on ImageNet in Four {GPU} Hours: {A} Theoretically Inspired Perspective},
  booktitle= {International Conference on Learning Representations},
  year     = {2021}
}

@inproceedings{hnas,
  author   = {Yao Shu and Zhongxiang Dai and Zhaoxuan Wu and Bryan Kian Hsiang Low},
  title    = {Unifying and Boosting Gradient-Based Training-Free Neural Architecture Search},
  booktitle= {Advances in Neural Information Processing Systems},
  year     = {2022}
}

@inproceedings{nasi,
  author   = {Yao Shu and Shaofeng Cai and Zhongxiang Dai and Beng Chin Ooi and Bryan Kian Hsiang Low},
  title    = {{NASI:} Label- and Data-agnostic Neural Architecture Search at Initialization},
  booktitle= {International Conference on Learning Representations},
  year     = {2022}
}

@inproceedings{reg_evo,
  author   = {Esteban Real and Alok Aggarwal and Yanping Huang and Quoc V. Le},
  title    = {Regularized Evolution for Image Classifier Architecture Search},
  booktitle= {Proceedings of the AAAI Conference on Artificial Intelligence},
  pages    = {4780--4789},
  year     = {2019}
}

@inproceedings{nb201,
  author   = {Xuanyi Dong and Yi Yang},
  title    = {NAS-Bench-201: Extending the Scope of Reproducible Neural Architecture Search},
  booktitle= {International Conference on Learning Representations},
  year     = {2020}
}

@inproceedings{nb101,
  author   = {Chris Ying and Aaron Klein and Eric Christiansen and Esteban Real and Kevin Murphy and Frank Hutter},
  title    = {NAS-Bench-101: Towards Reproducible Neural Architecture Search},
  booktitle= {Proceedings of the International Conference on Machine Learning},
  volume   = {97},
  pages    = {7105--7114},
  year     = {2019}
}

@inproceedings{SHALG,
  author   = {Kevin G. Jamieson and Ameet Talwalkar},
  title    = {Non-stochastic Best Arm Identification and Hyperparameter Optimization},
  booktitle= {Proceedings of the International Conference on Artificial Intelligence and Statistics},
  volume   = {51},
  pages    = {240--248},
  year     = {2016}
}

@inproceedings{predictornas2,
  author   = {Chenxi Liu and Barret Zoph and Maxim Neumann and Jonathon Shlens and Wei Hua and Li{-}Jia Li and Li Fei{-}Fei and Alan L. Yuille and Jonathan Huang and Kevin Murphy},
  title    = {Progressive Neural Architecture Search},
  booktitle= {Proceedings of the European Conference on Computer Vision},
  volume   = {11205},
  pages    = {19--35},
  year     = {2018}
}

@inproceedings{yang2022tabnas,
  author   = {Yang, Chengrun and Bender, Gabriel and Liu, Hanxiao and Kindermans, Pieter-Jan and Udell, Madeleine and Lu, Yifeng and Le, Quoc V and Huang, Da},
  title    = {TabNAS: Rejection Sampling for Neural Architecture Search on Tabular Datasets},
  booktitle= {Advances in Neural Information Processing Systems},
  year     = {2022}
}

@inproceedings{arm_net,
  author   = {Shaofeng Cai and Kaiping Zheng and Gang Chen and H. V. Jagadish and Beng Chin Ooi and Meihui Zhang},
  title    = {ARM-Net: Adaptive Relation Modeling Network for Structured Data},
  booktitle= {Proceedings of the ACM SIGMOD International Conference on Management of Data},
  year     = {2021}
}

@article{reg_luo,
  author   = {Zhaojing Luo and Shaofeng Cai and Gang Chen and Jinyang Gao and Wang{-}Chien Lee and Kee Yuan Ngiam and Meihui Zhang},
  title    = {Improving Data Analytics with Fast and Adaptive Regularization},
  journal  = {IEEE Transactions on Knowledge and Data Engineering},
  volume   = {33},
  year     = {2021}
}

@inproceedings{agebo,
  author   = {Romain {\'{E}}gel{\'{e}} and Prasanna Balaprakash and Isabelle Guyon and Venkatram Vishwanath and Fangfang Xia and Rick Stevens and Zhengying Liu},
  title    = {AgEBO-tabular: joint neural architecture and hyperparameter search with autotuned data-parallel training for tabular data},
  booktitle= {Proceedings of the International Conference for High Performance Computing, Networking, Storage and Analysis},
  year     = {2021}
}

@inproceedings{matrixaml,
  author   = {Nicol{\'{o}} Fusi and Rishit Sheth and Melih Elibol},
  title    = {Probabilistic Matrix Factorization for Automated Machine Learning},
  booktitle= {Advances in Neural Information Processing Systems},
  pages    = {3352--3361},
  year     = {2018}
}

@inproceedings{tpot,
  author   = {Randal S. Olson and Jason H. Moore},
  title    = {{TPOT:} {A} Tree-based Pipeline Optimization Tool for Automating Machine Learning},
  booktitle= {Proceedings of the Workshop on Automatic Machine Learning},
  volume   = {64},
  year     = {2016}
}

@inproceedings{oboe,
  author   = {Chengrun Yang and Yuji Akimoto and Dae Won Kim and Madeleine Udell},
  title    = {{OBOE:} Collaborative Filtering for AutoML Model Selection},
  booktitle= {Proceedings of the ACM SIGKDD Conference on Knowledge Discovery and Data Mining},
  year     = {2019}
}

@inproceedings{bohb,
  author   = {Stefan Falkner and Aaron Klein and Frank Hutter},
  title    = {{BOHB:} Robust and Efficient Hyperparameter Optimization at Scale},
  booktitle= {Proceedings of the International Conference on Machine Learning},
  volume   = {80},
  pages    = {1436--1445},
  year     = {2018}
}

@article{hyperband,
  author   = {Lisha Li and Kevin G. Jamieson and Giulia DeSalvo and Afshin Rostamizadeh and Ameet Talwalkar},
  title    = {Hyperband: {A} Novel Bandit-Based Approach to Hyperparameter Optimization},
  journal  = {Journal of Machine Learning Research},
  volume   = {18},
  pages    = {185:1--185:52},
  year     = {2017}
}

@inproceedings{rlnas,
  author   = {Barret Zoph and Quoc V. Le},
  title    = {Neural Architecture Search with Reinforcement Learning},
  booktitle= {International Conference on Learning Representations},
  year     = {2017}
}

@inproceedings{once4all,
  author   = {Han Cai and Chuang Gan and Tianzhe Wang and Zhekai Zhang and Song Han},
  title    = {Once-for-All: Train One Network and Specialize it for Efficient Deployment},
  booktitle= {International Conference on Learning Representations},
  year     = {2020}
}

@article{bergstra2012random,
  author   = {James Bergstra and Yoshua Bengio},
  title    = {Random Search for Hyper-Parameter Optimization},
  journal  = {Journal of Machine Learning Research},
  volume   = {13},
  pages    = {281--305},
  year     = {2012}
}

@article{shin2020trainability,
  author   = {Shin, Yeonjong and Karniadakis, George Em},
  title    = {Trainability of ReLU networks and data-dependent initialization},
  journal  = {Journal of Machine Learning for Modeling and Computing},
  volume   = {1},
  year     = {2020}
}

@inproceedings{jacot2018neural,
  author   = {Jacot, Arthur and Gabriel, Franck and Hongler, Cl{\'e}ment},
  title    = {Neural tangent kernel: Convergence and generalization in neural networks},
  booktitle= {Advances in Neural Information Processing Systems},
  year     = {2018}
}

@inproceedings{arora2019exact,
  author   = {Arora, Sanjeev and Du, Simon S and Hu, Wei and Li, Zhiyuan and Salakhutdinov, Russ R and Wang, Ruosong},
  title    = {On exact computation with an infinitely wide neural net},
  booktitle= {Advances in Neural Information Processing Systems},
  year     = {2019}
}

@inproceedings{allen2019convergence,
  author   = {Allen-Zhu, Zeyuan and Li, Yuanzhi and Song, Zhao},
  title    = {A convergence theory for deep learning via over-parameterization},
  booktitle= {Proceedings of the International Conference on Machine Learning},
  pages    = {242--252},
  year     = {2019}
}

@article{hornik1989multilayer,
  author   = {Hornik, Kurt and Stinchcombe, Maxwell and White, Halbert},
  title    = {Multilayer feedforward networks are universal approximators},
  journal  = {Neural Networks},
  volume   = {2},
  number   = {5},
  pages    = {359--366},
  year     = {1989}
}

@inproceedings{raghu2017expressive,
  author   = {Maithra Raghu and Ben Poole and Jon M. Kleinberg and Surya Ganguli and Jascha Sohl{-}Dickstein},
  title    = {On the Expressive Power of Deep Neural Networks},
  booktitle= {Proceedings of the International Conference on Machine Learning},
  volume   = {70},
  pages    = {2847--2854},
  year     = {2017}
}

@inproceedings{gorishniy2021revisiting,
  author   = {Gorishniy, Yury and Rubachev, Ivan and Khrulkov, Valentin and Babenko, Artem},
  title    = {Revisiting deep learning models for tabular data},
  booktitle= {Advances in Neural Information Processing Systems},
  pages    = {18932--18943},
  year     = {2021}
}

@inproceedings{kadra2021well,
  author   = {Kadra, Arlind and Lindauer, Marius and Hutter, Frank and Grabocka, Josif},
  title    = {Well-tuned simple nets excel on tabular datasets},
  booktitle= {Advances in Neural Information Processing Systems},
  year     = {2021}
}

@article{ren2021comprehensive,
  author   = {Ren, Pengzhen and Xiao, Yun and Chang, Xiaojun and Huang, Po-Yao and Li, Zhihui and Chen, Xiaojiang and Wang, Xin},
  title    = {A comprehensive survey of neural architecture search: Challenges and solutions},
  journal  = {ACM Computing Surveys},
  volume   = {54},
  number   = {4},
  pages    = {1--34},
  year     = {2021}
}

@inproceedings{Turner2020BlockSwap,
  author   = {Jack Turner and Elliot J. Crowley and Michael O'Boyle and Amos Storkey and Gavin Gray},
  title    = {BlockSwap: Fisher-guided Block Substitution for Network Compression on a Budget},
  booktitle= {International Conference on Learning Representations},
  year     = {2020}
}

@inproceedings{xie2021fives,
  author   = {Xie, Yuexiang and Wang, Zhen and Li, Yaliang and Ding, Bolin and G{\"u}rel, Nezihe Merve and Zhang, Ce and Huang, Minlie and Lin, Wei and Zhou, Jingren},
  title    = {Fives: Feature interaction via edge search for large-scale tabular data},
  booktitle= {Proceedings of the ACM SIGKDD Conference on Knowledge Discovery and Data Mining},
  pages    = {3795--3805},
  year     = {2021}
}

@inproceedings{klambauer2017self,
  author   = {Klambauer, G{\"u}nter and Unterthiner, Thomas and Mayr, Andreas and Hochreiter, Sepp},
  title    = {Self-normalizing neural networks},
  booktitle= {Advances in Neural Information Processing Systems},
  year     = {2017}
}

@inproceedings{song2019autoint,
  author   = {Song, Weiping and Shi, Chence and Xiao, Zhiping and Duan, Zhijian and Xu, Yewen and Zhang, Ming and Tang, Jian},
  title    = {Autoint: Automatic feature interaction learning via self-attentive neural networks},
  booktitle= {Proceedings of the ACM International Conference on Information and Knowledge Management},
  pages    = {1161--1170},
  year     = {2019}
}

@inproceedings{wang2020textnas,
  author   = {Wang, Yujing and Yang, Yaming and Chen, Yiren and Bai, Jing and Zhang, Ce and Su, Guinan and Kou, Xiaoyu and Tong, Yunhai and Yang, Mao and Zhou, Lidong},
  title    = {Textnas: A neural architecture search space tailored for text representation},
  booktitle= {Proceedings of the AAAI Conference on Artificial Intelligence},
  volume   = {34},
  pages    = {9242--9249},
  year     = {2020}
}

@article{theis2018faster,
  author   = {Theis, Lucas and Korshunova, Iryna and Tejani, Alykhan and Husz{\'a}r, Ferenc},
  title    = {Faster gaze prediction with dense networks and fisher pruning},
  journal  = {arXiv preprint},
  volume   = {arXiv:1801.05787},
  year     = {2018}
}

@inproceedings{siems2020bench,
  author   = {Julien Siems and Lucas Zimmer and Arber Zela and Jovita Lukasik and Margret Keuper and Frank Hutter},
  title    = {NAS-Bench-301 and the Case for Surrogate Benchmarks for Neural Architecture Search},
  booktitle= {NeurIPS 2020 Workshop on Meta-Learning},
  year     = {2020}
}

@inproceedings{kingma2014adam,
  author   = {Kingma, Diederik P and Ba, Jimmy},
  title    = {Adam: A method for stochastic optimization},
  booktitle= {International Conference on Learning Representations},
  year     = {2015}
}

@incollection{lecun2002efficient,
  author   = {LeCun, Yann and Bottou, L{\'e}on and Orr, Genevieve B and M{\"u}ller, Klaus-Robert},
  title    = {Efficient backprop},
  booktitle= {Neural Networks: Tricks of the Trade},
  pages    = {9--50},
  year     = {2002}
}

@inproceedings{glorot2010understanding,
  author   = {Glorot, Xavier and Bengio, Yoshua},
  title    = {Understanding the difficulty of training deep feedforward neural networks},
  booktitle= {Proceedings of the International Conference on Artificial Intelligence and Statistics},
  pages    = {249--256},
  year     = {2010}
}

@inproceedings{he2015delving,
  author   = {He, Kaiming and Zhang, Xiangyu and Ren, Shaoqing and Sun, Jian},
  title    = {Delving deep into rectifiers: Surpassing human-level performance on imagenet classification},
  booktitle= {Proceedings of the International Conference on Computer Vision},
  pages    = {1026--1034},
  year     = {2015}
}

@article{nas1000,
  author   = {Colin White and Mahmoud Safari and Rhea Sukthanker and Binxin Ru and Thomas Elsken and Arber Zela and Debadeepta Dey and Frank Hutter},
  title    = {Neural Architecture Search: Insights from 1000 Papers},
  journal  = {arXiv preprint},
  volume   = {arXiv:2301.08727},
  year     = {2023}
}

@inproceedings{metanas,
  author   = {Hayeon Lee and Sohyun An and Minseon Kim and Sung Ju Hwang},
  title    = {Meta-prediction Model for Distillation-Aware {NAS} on Unseen Datasets},
  booktitle= {International Conference on Learning Representations},
  year     = {2023}
}

@inproceedings{resourcenas,
  author   = {Ondrej Bohdal and Lukas Balles and Martin Wistuba and Beyza Ermis and C{\'{e}}dric Archambeau and Giovanni Zappella},
  title    = {{PASHA:} Efficient {HPO} and {NAS} with Progressive Resource Allocation},
  booktitle= {International Conference on Learning Representations},
  year     = {2023}
}

@inproceedings{zico,
  author   = {Guihong Li and Yuedong Yang and Kartikeya Bhardwaj and Radu Marculescu},
  title    = {ZiCo: Zero-shot {NAS} via inverse Coefficient of Variation on Gradients},
  booktitle= {International Conference on Learning Representations},
  year     = {2023}
}

@inproceedings{tarnsfernas,
  author   = {Gresa Shala and Thomas Elsken and Frank Hutter and Josif Grabocka},
  title    = {Transfer {NAS} with Meta-learned Bayesian Surrogates},
  booktitle= {International Conference on Learning Representations},
  year     = {2023}
}

@inproceedings{trainabilitypreserving,
  author   = {Huan Wang and Yun Fu},
  title    = {Trainability Preserving Neural Pruning},
  booktitle= {International Conference on Learning Representations},
  year     = {2023}
}

@inproceedings{hierarchyexpressivity,
  author   = {Qing Wang and Dillon Ze Chen and Asiri Wijesinghe and Shouheng Li and Muhammad Farhan},
  title    = {N-WL: A New Hierarchy of Expressivity for Graph Neural Networks},
  booktitle= {International Conference on Learning Representations},
  year     = {2023}
}

@article{regluo2,
  author   = {Zhaojing Luo and Shaofeng Cai and Yatong Wang and Beng Chin Ooi},
  title    = {Regularized Pairwise Relationship based Analytics for Structured Data},
  journal  = {Proceedings of the ACM SIGMOD International Conference on Management of Data},
  year     = {2023}
}

@inproceedings{tabulartransfer,
  author   = {Roman Levin and Valeriia Cherepanova and Avi Schwarzschild and Arpit Bansal and C. Bayan Bruss and Tom Goldstein and Andrew Gordon Wilson and Micah Goldblum},
  title    = {Transfer Learning with Deep Tabular Models},
  booktitle= {International Conference on Learning Representations},
  year     = {2023}
}

@inproceedings{complexirtyofnn,
  author   = {Boris Hanin and David Rolnick},
  title    = {Complexity of Linear Regions in Deep Networks},
  booktitle= {Proceedings of the International Conference on Machine Learning},
  volume   = {97},
  pages    = {2596--2604},
  year     = {2019}
}

@article{fernandez2014we,
  author   = {Fern{\'a}ndez-Delgado, Manuel and Cernadas, Eva and Barro, Sen{\'e}n and Amorim, Dinani},
  title    = {Do we need hundreds of classifiers to solve real world classification problems?},
  journal  = {Journal of Machine Learning Research},
  volume   = {15},
  number   = {1},
  pages    = {3133--3181},
  year     = {2014}
}

@inproceedings{expressmeettrain,
  author   = {Jiawei Zhang and Yushun Zhang and Mingyi Hong and Ruoyu Sun and Zhi{-}Quan Luo},
  title    = {When Expressivity Meets Trainability: Fewer than {\textdollar}n{\textdollar} Neurons Can Work},
  booktitle= {Advances in Neural Information Processing Systems},
  pages    = {9167--9180},
  year     = {2021}
}

@inproceedings{aucref,
  author   = {Fangye Wang and Yingxu Wang and Dongsheng Li and Hansu Gu and Tun Lu and Peng Zhang and Ning Gu},
  title    = {{CL4CTR:} {A} Contrastive Learning Framework for {CTR} Prediction},
  booktitle= {Proceedings of the ACM International Conference on Web Search and Data Mining},
  pages    = {805--813},
  year     = {2023}
}

@inproceedings{fairnas,
  author   = {Xiangxiang Chu and Bo Zhang and Ruijun Xu},
  title    = {FairNAS: Rethinking Evaluation Fairness of Weight Sharing Neural Architecture Search},
  booktitle= {Proceedings of the International Conference on Computer Vision},
  pages    = {12219--12228},
  year     = {2021}
}

@inproceedings{relbench,
  author   = {Matthias Fey and Weihua Hu and Kexin Huang and Jan Eric Lenssen and Rishabh Ranjan and Joshua Robinson and Rex Ying and Jiaxuan You and Jure Leskovec},
  title    = {Position: Relational Deep Learning - Graph Representation Learning on Relational Databases},
  booktitle= {Proceedings of the International Conference on Machine Learning},
  year     = {2024}
}

@inproceedings{beer,
  author   = {McAuley, Julian John and Leskovec, Jure},
  title    = {From amateurs to connoisseurs: modeling the evolution of user expertise through online reviews},
  booktitle= {Proceedings of the International Conference on World Wide Web},
  pages    = {897--908},
  year     = {2013}
}

@article{cox1958regression,
  author   = {Cox, David R},
  title    = {The regression analysis of binary sequences},
  journal  = {Journal of the Royal Statistical Society Series B: Statistical Methodology},
  volume   = {20},
  number   = {2},
  pages    = {215--232},
  year     = {1958}
}

@article{NussbaumMMD25,
  author   = {Zach Nussbaum and John Xavier Morris and Andriy Mulyar and Brandon Duderstadt},
  title    = {Nomic Embed: Training a Reproducible Long Context Text Embedder},
  journal  = {Transactions on Machine Learning Research},
  volume   = {2025},
  year     = {2025}
}

@inproceedings{XiaoLZMLN24,
  author   = {Shitao Xiao and Zheng Liu and Peitian Zhang and Niklas Muennighoff and Defu Lian and Jian{-}Yun Nie},
  title    = {C-Pack: Packed Resources For General Chinese Embeddings},
  booktitle= {Proceedings of the International ACM SIGIR Conference on Research and Development in Information Retrieval},
  pages    = {641--649},
  year     = {2024}
}

@article{Breiman01,
  author   = {Leo Breiman},
  title    = {Random Forests},
  journal  = {Machine Learning},
  volume   = {45},
  number   = {1},
  pages    = {5--32},
  year     = {2001}
}

@inproceedings{KeMFWCMYL17,
  author   = {Guolin Ke and Qi Meng and Thomas Finley and Taifeng Wang and Wei Chen and Weidong Ma and Qiwei Ye and Tie{-}Yan Liu},
  title    = {LightGBM: {A} Highly Efficient Gradient Boosting Decision Tree},
  booktitle= {Advances in Neural Information Processing Systems},
  pages    = {3146--3154},
  year     = {2017}
}

@inproceedings{ProkhorenkovaGV18,
  author   = {Liudmila Ostroumova Prokhorenkova and Gleb Gusev and Aleksandr Vorobev and Anna Veronika Dorogush and Andrey Gulin},
  title    = {CatBoost: unbiased boosting with categorical features},
  booktitle= {Advances in Neural Information Processing Systems},
  pages    = {6639--6649},
  year     = {2018}
}

@inproceedings{GuoTYLH17,
  author   = {Huifeng Guo and Ruiming Tang and Yunming Ye and Zhenguo Li and Xiuqiang He},
  title    = {DeepFM: {A} Factorization-Machine based Neural Network for {CTR} Prediction},
  booktitle= {Proceedings of the International Joint Conference on Artificial Intelligence},
  pages    = {1725--1731},
  year     = {2017}
}

@inproceedings{tabicl,
  author   = {Qu, Jingang and Holzm{\"u}ller, David and Varoquaux, Ga{\"e}l and Le Morvan, Marine},
  title    = {TabICL: A Tabular Foundation Model for In-Context Learning on Large Data},
  booktitle= {Proceedings of the International Conference on Machine Learning},
  year     = {2025}
}

@inproceedings{tabfpn,
  author   = {Noah Hollmann and Samuel M{\"{u}}ller and Katharina Eggensperger and Frank Hutter},
  title    = {TabPFN: {A} Transformer That Solves Small Tabular Classification Problems in a Second},
  booktitle= {International Conference on Learning Representations},
  year     = {2023}
}

@inproceedings{YanZXZC00C24,
  author   = {Jiahuan Yan and Bo Zheng and Hongxia Xu and Yiheng Zhu and Danny Z. Chen and Jimeng Sun and Jian Wu and Jintai Chen},
  title    = {Making Pre-trained Language Models Great on Tabular Prediction},
  booktitle= {International Conference on Learning Representations},
  year     = {2024}
}

@article{LeCunBH15,
  author   = {Yann LeCun and Yoshua Bengio and Geoffrey E. Hinton},
  title    = {Deep learning},
  journal  = {Nature},
  volume   = {521},
  number   = {7553},
  pages    = {436--444},
  year     = {2015}
}

@inproceedings{MuellerC025,
  author   = {Andreas C. Mueller and Carlo Curino and Raghu Ramakrishnan},
  title    = {MotherNet: Fast Training and Inference via Hyper-Network Transformers},
  booktitle= {International Conference on Learning Representations},
  year     = {2025}
}

@inproceedings{ChengJZLG25,
  author   = {Zi{-}Jian Cheng and Ziyi Jia and Zhi Zhou and Yufeng Li and Lan{-}Zhe Guo},
  title    = {TabFSBench: Tabular Benchmark for Feature Shifts in Open Environments},
  booktitle= {Proceedings of the International Conference on Machine Learning},
  year     = {2025}
}

@inproceedings{McElfreshKVCRGW23,
  author   = {Duncan C. McElfresh and Sujay Khandagale and Jonathan Valverde and Vishak Prasad C. and Ganesh Ramakrishnan and Micah Goldblum and Colin White},
  title    = {When Do Neural Nets Outperform Boosted Trees on Tabular Data?},
  booktitle= {Advances in Neural Information Processing Systems},
  year     = {2023}
}

@inproceedings{BonetMGI24,
  author   = {David Bonet and Daniel Mas Montserrat and Xavier Gir{\'{o}}{-}i{-}Nieto and Alexander G. Ioannidis},
  title    = {HyperFast: Instant Classification for Tabular Data},
  booktitle= {Proceedings of the AAAI Conference on Artificial Intelligence},
  pages    = {11114--11123},
  year     = {2024}
}

@inproceedings{ChengHYSW024,
  author   = {Yi Cheng and Renjun Hu and Haochao Ying and Xing Shi and Jian Wu and Wei Lin},
  title    = {Arithmetic Feature Interaction Is Necessary for Deep Tabular Learning},
  booktitle= {Proceedings of the AAAI Conference on Artificial Intelligence},
  pages    = {11516--11524},
  year     = {2024}
}

@inproceedings{JiZY025,
  author   = {Zipeng Ji and Guanghui Zhu and Chunfeng Yuan and Yihua Huang},
  title    = {{RZ-NAS:} Enhancing LLM-guided Neural Architecture Search via Reflective Zero-Cost Strategy},
  booktitle= {Proceedings of the International Conference on Machine Learning},
  year     = {2025}
}

@inproceedings{kangrevisiting,
  author   = {Kang, Haidong},
  title    = {Revisiting Neural Networks for Few-Shot Learning: A Zero-Cost NAS Perspective},
  booktitle= {Proceedings of the International Conference on Machine Learning},
  year     = {2025}
}

@inproceedings{dfs,
  author   = {James Max Kanter and Kalyan Veeramachaneni},
  title    = {Deep feature synthesis: Towards automating data science endeavors},
  booktitle= {Proceedings of the International Conference on Data Science and Advanced Analytics},
  pages    = {1--10},
  year     = {2015}
}

@article{BorisovLSHPK24,
  author   = {Vadim Borisov and Tobias Leemann and Kathrin Se{\ss}ler and Johannes Haug and Martin Pawelczyk and Gjergji Kasneci},
  title    = {Deep Neural Networks and Tabular Data: {A} Survey},
  journal  = {IEEE Transactions on Neural Networks and Learning Systems},
  volume   = {35},
  number   = {6},
  pages    = {7499--7519},
  year     = {2024}
}

@inproceedings{RobertsonH0AH25,
  author   = {Jake Robertson and Noah Hollmann and Samuel M{\"{u}}ller and Noor H. Awad and Frank Hutter},
  title    = {FairPFN: {A} Tabular Foundation Model for Causal Fairness},
  booktitle= {Proceedings of the International Conference on Machine Learning},
  year     = {2025}
}

@inproceedings{BreugelS24,
  author   = {Boris van Breugel and Mihaela van der Schaar},
  title    = {Position: Why Tabular Foundation Models Should Be a Research Priority},
  booktitle= {Proceedings of the International Conference on Machine Learning},
  year     = {2024}
}

@inproceedings{AkhauriA24,
  author   = {Yash Akhauri and Mohamed S. Abdelfattah},
  title    = {Encodings for Prediction-based Neural Architecture Search},
  booktitle= {Proceedings of the International Conference on Machine Learning},
  year     = {2024}
}

@inproceedings{KadlecovaLPVSNH24,
  author   = {Gabriela Kadlecov{\'{a}} and Jovita Lukasik and Martin Pil{\'{a}}t and Petra Vidnerov{\'{a}} and Mahmoud Safari and Roman Neruda and Frank Hutter},
  title    = {Surprisingly Strong Performance Prediction with Neural Graph Features},
  booktitle= {Proceedings of the International Conference on Machine Learning},
  year     = {2024}
}

@inproceedings{CasarinEL25,
  author   = {Sofia Casarin and Sergio Escalera and Oswald Lanz},
  title    = {{L-SWAG:} Layer-Sample Wise Activation with Gradients Information for Zero-Shot {NAS} on Vision Transformers},
  booktitle= {Proceedings of the IEEE Conference on Computer Vision and Pattern Recognition},
  pages    = {4441--4451},
  year     = {2025}
}

@inproceedings{OhLH25,
  author   = {Youngmin Oh and Hyunju Lee and Bumsub Ham},
  title    = {Efficient Few-Shot Neural Architecture Search by Counting the Number of Nonlinear Functions},
  booktitle= {Proceedings of the AAAI Conference on Artificial Intelligence},
  pages    = {19740--19748},
  year     = {2025}
}

@inproceedings{zhu2025context,
  author   = {Zhu, Jiaqi and Cai, Shaofeng and Shen, Yanyan and Chen, Gang and Deng, Fang and Ooi, Beng Chin},
  title    = {In-Context Adaptation to Concept Drift for Learned Database Operations},
  booktitle= {Proceedings of the International Conference on Machine Learning},
  pages    = {79699--79726},
  year     = {2025}
}

@article{zhu2023meter,
  author   = {Zhu, Jiaqi and Cai, Shaofeng and Deng, Fang and Ooi, Beng Chin and Zhang, Wenqiao},
  title    = {METER: A Dynamic Concept Adaptation Framework for Online Anomaly Detection},
  journal  = {Proceedings of the VLDB Endowment},
  volume   = {17},
  number   = {4},
  pages    = {794--807},
  year     = {2023}
}

@article{zhu2026catching,
  author   = {Zhu, Jiaqi and Cai, Shaofeng and Chen, Jie and Deng, Fang and Ooi, Beng Chin and Zhang, Wenqiao},
  title    = {Catching every Ripple: Enhanced Anomaly Awareness via Dynamic Concept Adaptation},
  journal  = {IEEE Transactions on Pattern Analysis and Machine Intelligence},
  year     = {2026}
}

@article{abs-2511-08667,
  author   = {L{\'{e}}o Grinsztajn and Klemens Fl{\"{o}}ge and Oscar Key and Felix Birkel and Philipp Jund and Brendan Roof and Benjamin J{\"{a}}ger and Dominik Safaric and Simone Alessi and Adrian Hayler and Mihir Manium and Rosen Yu and Felix Jablonski and Shi Bin Hoo and Anurag Garg and Jake Robertson and Magnus B{\"{u}}hler and Vladyslav Moroshan and Lennart Purucker and Clara Cornu and Lilly Charlotte Wehrhahn and Alessandro Bonetto and Bernhard Sch{\"{o}}lkopf and Sauraj Gambhir and Noah Hollmann and Frank Hutter},
  title    = {TabPFN-2.5: Advancing the State of the Art in Tabular Foundation Models},
  journal  = {arXiv preprint},
  volume   = {arXiv:2511.08667},
  year     = {2025}
}

@inproceedings{HolzmullerGS24,
  author   = {David Holzm{\"{u}}ller and L{\'{e}}o Grinsztajn and Ingo Steinwart},
  title    = {Better by default: Strong pre-tuned MLPs and boosted trees on tabular data},
  booktitle= {Advances in Neural Information Processing Systems},
  year     = {2024}
}

@article{ooi2024neurdb,
  author       = {Beng Chin Ooi and
                  Shaofeng Cai and
                  Gang Chen and
                  Yanyan Shen and
                  Kian{-}Lee Tan and
                  Yuncheng Wu and
                  Xiaokui Xiao and
                  Naili Xing and
                  Cong Yue and
                  Lingze Zeng and
                  Meihui Zhang and
                  Zhanhao Zhao},
  title        = {NeurDB: an AI-powered autonomous data system},
  journal      = {Sci. China Inf. Sci.},
  volume       = {67},
  number       = {10},
  year         = {2024}
}

@inproceedings{neurdb2025cidr,
  author       = {Zhanhao Zhao and
                  Shaofeng Cai and
                  Haotian Gao and
                  Hexiang Pan and
                  Siqi Xiang and
                  Naili Xing and
                  Gang Chen and
                  Beng Chin Ooi and
                  Yanyan Shen and
                  Yuncheng Wu and
                  Meihui Zhang},
  title        = {NeurDB: On the Design and Implementation of an AI-powered Autonomous
                  Database},
  booktitle    = {{CIDR}},
  publisher    = {www.cidrdb.org},
  year         = {2025}
}

@article{DBLP:journals/pvldb/XingCCLOP24,
  author       = {Naili Xing and
                  Shaofeng Cai and
                  Gang Chen and
                  Zhaojing Luo and
                  Beng Chin Ooi and
                  Jian Pei},
  title        = {Database Native Model Selection: Harnessing Deep Neural Networks in
                  Database Systems},
  journal      = {Proc. {VLDB} Endow.},
  volume       = {17},
  number       = {5},
  pages        = {1020--1033},
  year         = {2024}
}
\bibliographystyle{icml2026}


\newpage
\appendix
\onecolumn

\clearpage
\begin{appendix}

\section*{Appendix}

\begin{itemize}
  \item \textbf{\ref{app:notation}. Notation Table} \dotfill \pageref{app:notation}
  
  \item \textbf{\ref{app:related_work}. Literature Review} \dotfill \pageref{app:related_work}
  
  \item \textbf{\ref{app:exp_setup}. Experiment Setup} \dotfill \pageref{app:exp_setup}

  \begin{itemize}
     \item[$\cdot$] \textit{\ref{app:dataset}. Datasets and Preprocessing} \dotfill \pageref{app:dataset}
     
     \item[$\cdot$] \textit{\ref{app:train_detail}. Training Protocol and Hyperparameters} \dotfill \pageref{app:train_detail}

     \item[$\cdot$] \textit{\ref{app:code_data}. Implementation Details} \dotfill \pageref{app:code_data}

  \end{itemize}
  
  \item \textbf{\ref{app:nasbench}. NAS-Bench-Tabular and Search Space} \dotfill 
  \pageref{app:nasbench}
  
  \begin{itemize}
     \item[$\cdot$] \textit{\ref{app:nasbench_parameter}. Training Hyperparameters for Building \nasbench} \dotfill \pageref{app:nasbench_parameter}
     
     \item[$\cdot$] \textit{\ref{app:nasbench_space}. Search Space Design and Characterization} \dotfill \pageref{app:nasbench_space}

     \item[$\cdot$] \textit{\ref{app:nasbench_search_strategy}. Benchmarking Search Strategies on \nasbench} \dotfill \pageref{app:nasbench_search_strategy}

     \item[$\cdot$] \textit{\ref{app:new_search_spaces}. Generalization to Additional Search Spaces} \dotfill \pageref{app:new_search_spaces}

  \end{itemize}

  \item \textbf{\ref{app:theoretical_analysis}. Theoretical Analysis of \newtfmem} \dotfill \pageref{app:theoretical_analysis}

  \item \textbf{\ref{app:extended_exp}. Extended Experiments} \dotfill \pageref{app:extended_exp}
  
  \begin{itemize}
     \item[$\cdot$] \textit{\ref{app:abs_tfmem}. Ablation Studies of \newtfmem} \dotfill \pageref{app:abs_tfmem}
     
     \item[$\cdot$] \textit{\ref{app:abs_two_phase}. Component Analysis of Two-Phase Design} \dotfill \pageref{app:abs_two_phase}

     \item[$\cdot$] \textit{\ref{app:abs_coord}. Hyperparameter Analysis of the Coordinator} \dotfill \pageref{app:abs_coord}

     \item[$\cdot$] \textit{\ref{app:tfmem_compare_table}. Comparison and Analysis of \tfmem[s]} \dotfill \pageref{app:tfmem_compare_table}

     \item[$\cdot$] \textit{\ref{app:more_baseline}. Comparison of \framework with Additional Baselines} \dotfill \pageref{app:more_baseline}

      \item[$\cdot$] \textit{\ref{app:more_exp_visual}. Visualization of Correlation for \tfmem[s]} \dotfill \pageref{app:more_exp_visual}
          
  \end{itemize}
  
  \item \textbf{\ref{app:relbench_full}. Complete Results Across All Datasets} \dotfill \pageref{app:relbench_full}
   
\end{itemize}

\clearpage
\section{Notations}
\label{app:notation}

\begin{center}
\begin{minipage}{\textwidth}
    \centering
    \captionof{table}{Summary of Notation and Terminology.}
    \renewcommand{\arraystretch}{1}
    \setlength\tabcolsep{1.9mm}
    \begin{tabular}{ll}
    \toprule[1.5pt]
    
        \budget & Time budget \\
        
        $\mathcal{A} = \{a\}$  & Search space and architecture \\
        $L$ & Number of layers in the DNN \\
        $h_l$ & the $l$-th layer in the DNN \\
        $\mathcal{H}$ & A candidate set of layer sizes\\
        $|\mathcal{H}|^L$ & Number of candidate architectures \\
        $f_s $ & Search strategy\\
        $\mathcal{S}_i$ & State of the search strategy at the $i$-th iteration.\\
        $\boldsymbol{\theta}$ & Parameter of an architecture \\
        $B$, $X_B$ & Batch size. A batch of data samples\\

        \midrule[0.5pt] 

        $N$ & Number of neurons of the whole architecture \\
        $N_l$ & Number of neurons of the $l$-th layer of the architecture \\
        $\mathcal{L}$ & Loss function \\

        $s_a$ & \tfmem score of architecture\\
        $\rho(\cdot)$ & \tfmem assessment function \\
        $\bigodot$ &  Hadamard product \\
        $\Theta $ & NTK metrics\\
        $\Phi$ & Synaptic saliency \\
        $\nu_n$ & Neuron saliency of the $n$-th neuron computed on a batch of samples \\
        $\nu_{in}$ & Neuron saliency of the $n$-th neuron computed on the $i$-th sample \\
        $ z_n $ & Activated output of the $n$-th neuron computed on a batch of samples \\
        $ z_{in} $ & Activated output of the $n$-th neuron computed on the $i$-th sample \\
        $\mathbf{w} $ & Incoming weights of the neuron \\
        $\sigma $ & Activation function \\
        $\mathbf{x} $ & Neuron inputs  \\

        \midrule[0.5pt]
        $\mathcal{K}_l$ & Number of neurons in the $l$-th layer \\ 
        $z^{(l)}(x)$ & Activated representation (activation vector) at layer $l$ given input $x$ \\
        $\delta$ & Standard Gaussian perturbation with the same shape as $x$ \\
        $\epsilon$ & Perturbation magnitude (a small positive scalar) \\
        $d_l$ & Layer-wise amplification/sensitivity estimated by finite difference \\

        \midrule[0.5pt]
        $M $ &  Number of architectures explored in the \filter phase \\
        $K$ &  Number of architectures exploited in the \refine phase  \\
        $U$ & Computational unit in the \refine phase \\
        $t_1$ & Time to score an architecture\\
        $t_2$ & Time to train an architecture for a single epoch \\
        $T_1$ & Time allocated to the \filter phase \\
        $T_2$ & Time allocated to the \refine phase\\
        $\Omega$ & Joint search configuration tuple, defined as $\{M, K, U\}$ \\
        $H$ & Hessian vector \\

        \midrule[0.5pt]
        $\mathcal{N}$ & Number of architecture encodings \\
        $\eta$ & $\frac{K}{\eta}$ Architectures to keep per-round \\ 
        
    \bottomrule[1.5pt]
    \end{tabular}
\end{minipage}
\end{center}

\section{Literature Review}
\label{app:related_work}

\subsection{DNN on Tabular Data}
\label{app:tab_data_review}

Different approaches have attempted to apply DNN techniques to tabular data, ranging from DNN design~\cite{tabulartransfer,gorishniy2021revisiting,reg_luo,zhu2026catching}, Automated Machine Learning (AutoML) on tabular data~\cite{matrixaml,tpot,oboe}, to Tabular Foundation Models~\cite{tabicl,tabfpn,RobertsonH0AH25}.
TFMs leverage large-scale pretraining and in-context learning (ICL) to make predictions conditioned on a small labeled sample set provided as context, reducing or even eliminating per-dataset training.
However, TFMs push much of the computational cost to inference, exposing a critical efficiency--accuracy trade-off: while providing more labeled samples as context has the potential to improve adaptation to the target data distribution, inference latency increases with context length and can quickly become a bottleneck~\cite{BonetMGI24,MuellerC025}.

In contrast, dataset-specific DNNs can provide fast inference after fitting, but their performance is highly dependent on the choice of architecture and training configuration.
This motivates automated architecture discovery for tabular data, where the goal is to identify compact and high-performing DNNs for each dataset under practical search budgets.

\subsection{Neural Architecture Search (NAS)}
\label{app:nas_review}

Neural Architecture Search (NAS) is designed to automate the discovery of architectures optimized for specified datasets, thereby eliminating the need for manual design and experimentation. 
The process often entails the search for connection patterns within a predefined architectural backbone, such as \fnn~\cite{yang2022tabnas} or cell-based convolutional neural network~\cite{nb101,nb201}.

A critical component of NAS is the architecture evaluation. 
Initial research in NAS~\cite{rlnas,MetaQNN} primarily relied on the time-consuming and resource-intensive process of fully training each architecture to convergence. 
Several methods have been proposed to address this issue and to mitigate costs~\cite{metanas,tarnsfernas,zico,resourcenas}.
They generally fall into three categories: 
NAS with performance prediction, NAS based on weight sharing, and NAS with training-free architecture evaluation.

Performance prediction in NAS involves training a model to forecast the final performance of an architecture based on features derived from hyperparameters, architectural structures, and partially trained architectures.
This category of methods shows improvements in searching efficiency but it is limited in tuning the predictors and is hard to enhance generalizability~\cite{AkhauriA24,KadlecovaLPVSNH24}.

In contrast, NAS based on weight sharing seeks to identify a subgraph within a larger computation graph~\cite{ENAS,CasarinEL25}. This allows multiple sampled subgraphs sharing the same computation unit to utilize a common set of weights. 
However, the individual sampling and training procedure of each discrete subgraph leads to an increased number of architectures to be trained.
And inheriting weights from the larger computational graph does not necessarily ensure improved training efficiency~\cite{fairnas}.
To further reduce the number of architectures requiring training~\cite{DARTS} proposed shifting the search from a discrete space to a continuous one. 
This enables gradient optimization to expedite the search process. 

Last, zero-cost proxies estimate architecture performance by calculating certain statistics of the architecture at initialization without requiring full training~\cite{JiZY025,zico,kangrevisiting,OhLH25}. One of the main advantages of the training-free evaluation is its extremely high computational efficiency, requiring only a single forward and/or backward computation.

However, NAS on tabular data has been relatively less explored. 
Recently, AgEBO-Tabular~\cite{agebo} and TabNAS~\cite{yang2022tabnas} investigated the application of NAS on tabular data to achieve more efficient and higher-performing architectures. 
Specifically, AgEBO-Tabular integrates NAS with aging evolution in a search space that includes multiple branches and hyperparameter tuning using Bayesian optimization. 
In comparison, TabNAS aims to identify high-performing architectures from \fnn under specified resource constraints by utilizing reinforcement learning with reject sampling. 
It shows that simple multiple layers \fnn can already yield outstanding performance.
However, both approaches rely on training-based architecture evaluation and cannot achieve progressive NAS on tabular data.
Although progressive NAS has been studied in other domains, such as PNAS~\cite{predictornas2}, these methods are not directly applicable to tabular data:
(1) they are often designed for vision-specific, cell-based CNN search spaces rather than tabular DNN architectures;
(2) their search procedures remain training-based, which is inefficient under the tight budgets required by practical tabular analytics; and
(3) their predictors typically rely on architecture encodings rather than tabular-specific, data-aware signals, and therefore do not explicitly capture properties central to tabular learning.

These limitations motivate progressive tabular NAS that can use efficient architecture filtering while retaining training-based refinement only for promising candidates.

\subsection{Architecture Properties}
\label{app:arch_char}

Architecture performance is influenced by two factors: trainability and expressivity. Trainability measures the extent to which gradient descent can effectively optimize the architecture.
Expressivity denotes the complexity of the function that the architecture can model.
Many training-free evaluation approaches estimate the architecture's performance by characterizing both properties. 

More recently, TE-NAS~\cite{tenas} proposes to quantify the expressivity of a ReLU-based DNN by computing the number of linear regions that the architecture can divide for a batch of data.
Likewise, NASWOT~\cite{naswot} characterizes expressivity by measuring the distance between the vectors of activation patterns for any two samples within a batch.
A greater distance suggests a higher capability to distinguish different samples, indicating good expressivity.

The application of the Neural Tangent Kernel (NTK) as a measure of trainability has recently been explored, given a batch of data $X_B$, NTK~\cite{jacot2018neural,arora2019exact,allen2019convergence} characterizes the complexities of the training dynamics at initialization, which is defined as 
\begin{equation}
\Theta(X_B, X_B; \boldsymbol{\theta}) = \bigtriangledown_{\boldsymbol{\theta}}f(X_B;\boldsymbol{\theta}) \bigtriangledown_{\boldsymbol{\theta}}f(X_B;\boldsymbol{\theta})^{T}   
\end{equation}
NTK-related proxies are adopted in many training-free evaluation approaches, such as NTKTrace~\cite{hnas}, NTKTraceAppx~\cite{nasi}, and NTKCond~\cite{tenas}.

The trainability has also been studied in the context of network pruning~\cite{synflow,GRASP,snip}, which identifies and prunes less significant parameters. 
The notion of synaptic saliency~\cite{synflow} is proposed to quantify each parameter's importance, defined as 
\begin{equation}
    \Phi(\boldsymbol{\theta}) = f(\frac{\partial \mathcal{L}}{\partial \boldsymbol{\theta}})\bigodot g(\boldsymbol{\theta})
\end{equation}
Different \tfmem[s] basically differ in $f(\cdot)$ and $g(\cdot)$, for SNIP~\cite{snip}
\begin{equation}
    \Phi(\boldsymbol{\theta}) = | \frac{\partial \mathcal{L}}{\partial \boldsymbol{\theta}}| \bigodot |\boldsymbol{\theta} |
\end{equation}
for GraSP~\cite{GRASP}
\begin{equation}
\Phi(\boldsymbol{\theta}) = -( H \frac{\mathcal{L}}{\partial \boldsymbol{\theta}} ) \bigodot \boldsymbol{\theta}
\end{equation}
and for SynFlow~\cite{synflow}
\begin{equation}
    \Phi(\boldsymbol{\theta}) = \frac{\partial \mathcal{L}}{\partial \boldsymbol{\theta}} \bigodot \boldsymbol{\theta}
\end{equation}
where $H$ is the Hessian vector.
Similarly, Fisher~\cite{Turner2020BlockSwap} quantifies the performance by aggregating layer Fisher information~\cite{theis2018faster}.

We provide the full summarization of different \tfmem[s] in Table~\ref{tab:tfmem_ana_summary}, with their evaluation metrics, complexity, score computation definition, and the characterized properties of DNN.

Taken together, \framework advances existing work in three aspects.
First, unlike traditional tabular models with static architectures (Table~\ref{tab:rel_bench_acc}), \framework searches for dataset-specific DNN architectures.
Second, unlike existing tabular NAS methods such as TabNAS that rely on expensive training-based search, \framework introduces a training-free filtering paradigm for efficient NAS on tabular data and combines it with lightweight refinement.
Third, whereas existing zero-cost proxies are primarily designed for vision-oriented settings, \newtfmem is tailored to tabular data by characterizing properties central to tabular deep learning, including expressivity induced by high-order feature interactions.
To the best of our knowledge, this is the first work in this direction.

\section{Experiment Setup}
\label{app:exp_setup}

\subsection{Datasets and Preprocessing}
\label{app:dataset}

We evaluate our approach on 11 datasets, including 3 widely-used tabular datasets (Frappe, UCI Diabetes, and Criteo) and 8 realistic multi-table datasets sourced from the RelBench benchmark~\cite{relbench}. 
RelBench is designed for efficient and reproducible evaluation of end-to-end learning over relational databases, covering diverse real-world domains. 
Table~\ref{table:data_statistics} reports the key statistics of all datasets, and detailed descriptions are provided below.

For each multi-table dataset, we transform the relational database into a target-table learning setup via a DFS-style feature construction procedure~\cite{dfs}. 
Starting from the target table, we traverse the schema graph and generate relational features by aggregating values from neighboring tables along join paths using predefined operators (e.g., count, sum/mean, and max/min). 
Following the temporal split, all features for each example are computed using only historical records available before its prediction time, and the same transformation is applied to validation/test without accessing future information. 
This yields a unified feature representation for downstream models while preserving informative context from the original relational structure.

\begin{table}[t]
    \caption{A Comparison of Different Zero-Cost Proxies.}
    \label{tab:tfmem_ana_summary}
    \centering
    \begingroup
    \footnotesize
    \setlength{\tabcolsep}{1.5pt}
    \renewcommand{\arraystretch}{1.3}
    \resizebox{0.8\linewidth}{!}{
        \begin{tabular}{l|c|l|l|l}
            \toprule[1pt]
            Zero-Cost Proxy & Evaluation Metric & Complexity & Computation & Property \\
            \toprule[1pt]
            GradNorm & Frobenius norm & 1FC+1BC &
            $ s_a = \left\|\frac{\partial \mathcal{L}}{\partial \boldsymbol{\theta}}\right\|_F $
            & Express. \\
            
            NASWOT & Hamming distance & 1FC &
            $ s_a = \log|K_H| $
            & Express. \\
            
            NTKCond & Neural tangent kernel & 1FC+1BC &
            $ s_a = \frac{\lambda_{\max}(\Theta)}{\lambda_{\min}(\Theta)} $
            & Train. \\
            
            NTKTrace & Neural tangent kernel & 1FC+1BC &
            $ s_a = \|\Theta\|_{\mathrm{trace}} $
            & Train. \\
            
            NTKTraceAppx & Neural tangent kernel & 1FC+1BC &
            $ s_a = \|\Theta_{\mathrm{appx}}\|_{\mathrm{trace}} $
            & Train. \\
            
            Fisher & Hadamard product & 1FC+1BC &
            $ s_a = \sum_{l=1}^{L} \left(\frac{\partial \mathcal{L}}{\partial ac_l} ac_l\right)^2 $
            & Train. \\
            
            GraSP & Hessian vector product & 1FC+1BC &
            $ s_a = \sum -\left( H \frac{\partial \mathcal{L}}{\partial \boldsymbol{\theta}} \right) \bigodot \boldsymbol{\theta} $
            & Train. \\
            
            SNIP & Hadamard product & 1FC+1BC &
            $ s_a = \sum \left|\frac{\partial \mathcal{L}}{\partial \boldsymbol{\theta}} \bigodot \boldsymbol{\theta}\right| $
            & Train. \\
            
            SynFlow & Hadamard product & 1FC+1BC &
            $ s_a = \sum \frac{\partial \mathcal{L}}{\partial \boldsymbol{\theta}} \bigodot \boldsymbol{\theta} $
            & Train. \\
            
            WeightNorm & Frobenius norm & 1FC &
            $ s_a = \|\boldsymbol{\theta}\|_F $
            & Express. \\
            
            \toprule[1pt]
            \newtfmem & Hadamard product & 2FC+1BC &
            $ s_a = \sum_{n} \left|\frac{\partial \mathcal{L}}{\partial z_n}\right| \bigodot z_n $
            & Train. \& Express.  \\
            
            \toprule[1pt]
        \end{tabular}
    }
    \begin{tablenotes}
        \item
        $\mathcal{L}$: loss function. 
        $\boldsymbol{\theta}$: architecture parameters.
        $\Theta$: NTK matrix of the architecture.\\
        $\bigodot$: Hadamard product. 
        $s_a$: the score of an architecture $a$.\\
        $\lambda$: eigenvalue of the NTK matrix.
        $H$: Hessian vector.
        $L$: number of architecture layers.\\
        $\| \cdot \|_F$: Frobenius norm. 
        $ac_l$: activation saliency of layer $l$.\\
        $FC$: forward computation.
        $BC$: backward computation.
    \end{tablenotes}
    \endgroup
\end{table}

\highlight{Frappe}~\footnote{https://www.baltrunas.info/research-menu/frappe} is a dataset from the real-world application recommendation scenario, which incorporates context-aware app usage logs consisting of 96,203 tuples from 957 users across 4,082 apps used in various contexts. 
For each positive app usage log, Frappe generates two negative tuples, resulting in a total of 288,609 tuples. 
The learning objective is to predict app usage based on the context, encompassing 10 semantic attribute fields with 5,382 distinct numerical and categorical embedding vectors.

\highlight{UCI Diabetes}~\footnote{https://archive.ics.uci.edu/ml/datasets} (Diabetes) encompasses a decade (1999-2008) of clinical diabetes encounters from 130 US hospitals.
This dataset aims to analyze historical diabetes care to enhance patient safety and deliver personalized healthcare. 
With 101,766 encounters from diabetes-diagnosed patients, the primary learning objective is to predict inpatient readmissions.
This dataset consists of 43 attributes and 369 distinct numerical and categorical embedding vectors, including patient demographics and illness severity factors like gender, age, race, discharge disposition, and primary diagnosis.

\highlight{Criteo}~\footnote{https://labs.criteo.com/2014/02/kaggle-display-advertising-challenge-dataset/} is a CTR benchmark consisting of attribute values and click feedback for millions of display advertisements.
The learning objective is to predict if a user will click a specific ad in the context of a webpage. 
This dataset has 45,840,617 tuples across 39 attribute fields with 2,086,936 distinct numerical and categorical embedding vectors. 
These include 13 numerical attribute fields and 26 categorical attributes.

\highlight{Event}~\footnote{https://www.kaggle.com/c/event-recommendation-engine-challenge} is an event recommendation dataset derived from a mobile social networking app Hangtime, capturing users’ social plans, invitations, and interactions. It connects users, events, and friendships to model evolving engagement behavior over time. The dataset spans user activity records in 2012 and supports two prediction objectives:
    (1) user-attendance: estimating how many events a user will respond to in the following week (regression task), and 
    (2) user-repeat: predicting whether a user will attend another event within a week after prior participation (classification task).

\highlight{Avito}~\footnote{https://www.kaggle.com/c/avito-context-ad-clicks} is an online advertising marketplace dataset derived from the Avito advertisement platform, capturing user queries, ad metadata, locations, and interaction logs. It connects users, ads, and sessions to model evolving browsing and engagement behavior. 
The dataset covers user activity from late April to May 2015 and supports two prediction objectives:
    (1) user-clicks: predicting whether a user will click multiple ads in the next four days (classification task), and
    (2) ad-ctr: estimating an advertisement's click-through rate (regression task).

\highlight{Trial}~\footnote{https://aact.ctti-clinicaltrials.org/} is a relational dataset curated from the AACT initiative that captures study protocols, participating sites, medical conditions, interventions, and outcome reports. It links studies, sites, and interventions to model trial progress and safety, covering records from 2020 to 2021. The dataset supports two prediction objectives: 
    (1) study-outcome: predicting whether a trial achieves its primary outcome (classification task), and 
    (2) site-success: forecasting site-level success rates over the subsequent year (regression task).

\highlight{Beer}~\cite{beer} is a relational dataset of beer reviews linking users, beers, and drinking venues (e.g., bars/breweries). It captures user ratings and textual feedback across seasons and supports two prediction objectives:
    (1) user-active (classification task): predicting whether a user will post more than 10 reviews in the next season, and
    (2) beer-positive (regression task): estimating a beer’s positive-rating ratio, where ratings above 3.5 are considered positive.

\highlight{HM}~\footnote{https://www.kaggle.com/competitions/h-and-m-personalized-fashion-recommendations} is a retail-market dataset consisting of customer profiles, item metadata, and fine-grained transaction records. 
It captures purchase behaviors at the item level and enables forecasting future sales dynamics. 
We consider a prediction objective defined over a one-week horizon, where the goal is to estimate the sales volume of each product in the upcoming week. 
The dataset supports two prediction objectives:
    (1) item-sales: estimating the sales volume of each product in the upcoming week, computed as the total transaction value aggregated over that week for each item (regression task), and
    (2) user-churn: predicting whether a customer becomes inactive in future transactions (classification task).

\begin{table}[t]
    \centering
    \caption{Dataset Statistics.}
    \renewcommand{\arraystretch}{1.2}
    \resizebox{0.9\textwidth}{!}{
    \begin{tabular}{c|cccc|ccc|c}
        \toprule[1.5pt]
        Dataset & \#Class & \#Feature & Task & Domain & Train & Valid & Test & Total \\
        \midrule[0.5pt]
        Frappe                     & 2   & 10  & Cls. & AppRec     & 202,027  & 57,722  & 28,860  & 288,609 \\
        Diabetes                  & 2   & 43  & Cls. & Health     & 81,412   & 10,177  & 10,177  & 101,766 \\
        Criteo                    & 2   & 39  & Cls. & CTR        & 33,003,326 & 8,250,124 & 4,587,167 & 45,840,617 \\
        
        Event (user-repeat)       & 2   & 25  & Cls. & RecSys     & 3,842    & 268     & 246     & 4,356 \\
        Event (user-attend)       & --  & 19  & Regr.     & RecSys     & 19,239   & 2,013   & 1,958   & 23,210 \\
        
        Beer (user-active)        & 2   & 47  & Cls. & Review     & 16,656   & 2,794   & 3,558   & 23,008 \\
        Beer (beer-pos)           & --  & 43  & Regr.     & Review     & 45,922   & 12,858  & 7,218   & 66,998 \\
        
        Trial (study-outcome)     & 2   & 50  & Cls. & Health     & 11,994   & 960     & 825     & 13,779 \\
        Trial (site-success)      & --  & 14  & Regr.     & Health     & 100,000  & 19,740  & 22,617  & 142,357 \\
        
        Avito (user-click)        & 2   & 15  & Cls. & Online     & 59,454   & 21,183  & 47,996  & 128,633 \\
        Avito (ad-ctr)            & --  & 18  & Regr.     & Online     & 5,100    & 1,766   & 1,816   & 8,682 \\
        
        HM (user-churn)           & 2   & 13  & Cls. & Retail     & 100,000  & 76,556  & 74,575  & 251,131 \\
        HM (item-sales)           & --  & 32  & Regr.     & Retail     & 100,000  & 100,000 & 100,000 & 300,000 \\
        \bottomrule[1.5pt]
    \end{tabular}
    }
    \label{table:data_statistics}
\end{table}

\subsection{Training Protocol and Hyperparameters}
\label{app:train_detail}

For all baseline models as listed in Table~\ref{tab:rel_bench_acc}, we adopt a unified training protocol. We use the Adam optimizer with a fixed learning rate of $10^{-3}$ (no learning-rate schedule) and a batch size of 256. Training runs for up to 200 epochs with early stopping (patience 10), and we cap the computation per epoch by using at most 20 batches to standardize per-epoch computation across methods.

For the classification task, we optimize BCEWithLogits loss and report ROC-AUC; for regression objectives, we optimize L1 loss and report MAE. Following our implementation, dropout is deactivated for the regression task to improve training stability. The final architecture is selected by validation performance and then evaluated on the held-out test split.

\subsection{Implementation Details}
\label{app:code_data}
All experiments in this paper are implemented in PyTorch and run on a local server equipped with an Intel(R) Xeon(R) Silver 4214R CPU (12 cores), 128\, GB memory, and eight NVIDIA GeForce RTX 3090 GPUs.
To support reproducibility and fair comparisons in NAS research, we release the source code, reproduction scripts, and links to the precomputed benchmark/result artifacts at \url{https://github.com/NLGithubWP/pTNAS}. 
The repository contains the training and evaluation pipelines, example outputs, and step-by-step instructions in \texttt{README.md}; the same \texttt{README.md} provides download links for \nasbench and other large artifacts.

\section{NAS-Bench-Tabular and Search Space}
\label{app:nasbench}

\subsection{Training Hyperparameters for Building \nasbench}
\label{app:nasbench_parameter}

We build \nasbench using three datasets: Frappe, UCI Diabetes, and Criteo.
Training hyperparameters (e.g., batch size, training epochs/iterations, and learning rate) can interact with architecture size, materially affecting measured performance.
To obtain reliable training configurations, for each dataset, we first select three representative architectures from our search space, covering small, medium, and large model sizes, and perform a grid search to tune training hyperparameters for each representative architecture separately.

We require that, under their tuned hyperparameters, these three representative architectures achieve DNN performance consistent with prior results on the same datasets~\cite{arm_net,kadra2021well,klambauer2017self,fernandez2014we}.
The resulting three sets of hyperparameters (small/medium/large) for each dataset are summarized in Table~\ref{table:train_hyperparams}.
For Adam, we use $\beta_1=0.9$, $\beta_2=0.999$, decay $=0$, and $\eps=1\mathrm{e}{-8}$, following~\cite{yang2022tabnas,kingma2014adam}.

Using the above protocol, we fully train every architecture in the search space on each dataset.
Each architecture is trained using the hyperparameter set corresponding to its size category (small, medium, or large), where the category is determined by bucketing architectures by their parameter counts using two fixed thresholds.

For each architecture-dataset pair, we record five metrics: training/validation AUC, training/validation loss, and training time for all trained architectures, where training AUC is used for analyzing overfitting behavior and space statistics, and all NAS evaluations query validation AUC.

\begin{table}[t]
    \centering
    \caption{Training Hyperparameters.}
        \renewcommand{\arraystretch}{0.9}
        \begin{tabular}{c|ccccccc}
        \toprule[1.5pt]
            Dataset & \makecell{batch \\size} & \makecell{learning\\rate} & \makecell{learning\\rate schedule} & optimizer & \makecell{training \\ epoch} & \makecell{iteration \\ per epoch} & \makecell{loss \\function} \\
            \midrule[0.5pt]
            Frappe & 512 & 0.001 & cosine decay & Adam & 20 & 200 & BCELoss\\
            Diabetes &1024 & 0.001 & cosine decay & Adam & 1 & 200 & BCELoss \\
            Criteo &  1024 & 0.001 & cosine decay & Adam & 10 & 2000 & BCELoss\\
        \bottomrule[1.5pt]
        \end{tabular}
    \label{table:train_hyperparams}
\end{table}

\subsection{Search Space Design and Characterization}
\label{app:nasbench_space}

We consider a DNN-based search space that varies the hidden-layer widths while keeping other backbone components fixed.
Specifically, we fix the number of hidden layers $L$ to be four for all three tabular datasets and define an architecture $a$ by its layer-width tuple
$a = (h_1, h_2, h_3, h_4)$, where each $h_\ell$ is selected from a candidate set $\mathcal{H}$.
We fix the activation (ReLU), output layer, and training-time regularization choices (e.g., dropout rate), and only vary hidden widths.
We set $\mathcal{H}$ as follows:
\begin{itemize}
\item \textbf{Frappe and Diabetes:} $\mathcal{H} = [8, 16, 24, 32, 48, 64, 80, 96, 112, 128, 144, 160, 176, 192, 208, 224, 240, 256, 384, 512]$.
\item \textbf{Criteo:} $\mathcal{H} = [8, 16, 32, 48, 64, 112, 144, 176, 240, 384]$.
\end{itemize}

\begin{figure*}[t]
\centering
\begin{subfigure}[b]{1\textwidth}
    \centering
    \hspace{1mm}
    \includegraphics[width=0.35\columnwidth]{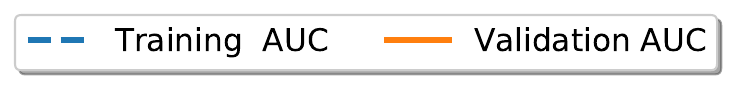}
\end{subfigure}
\begin{subfigure}[b]{0.32\textwidth}
    \centering
    \includegraphics[width=0.98\columnwidth]{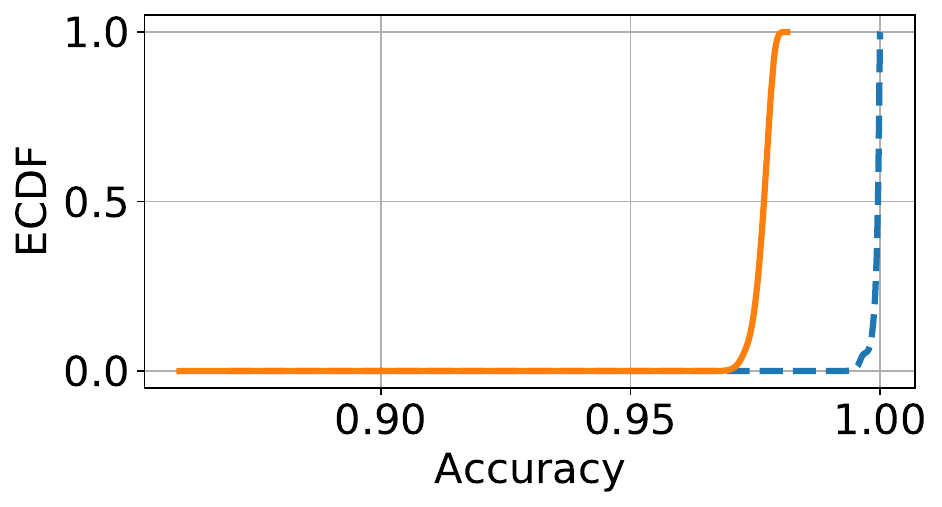}
    \caption{Frappe.}
\end{subfigure}
~
\begin{subfigure}[b]{0.32\textwidth}
    \centering
    \includegraphics[width=0.98\columnwidth]{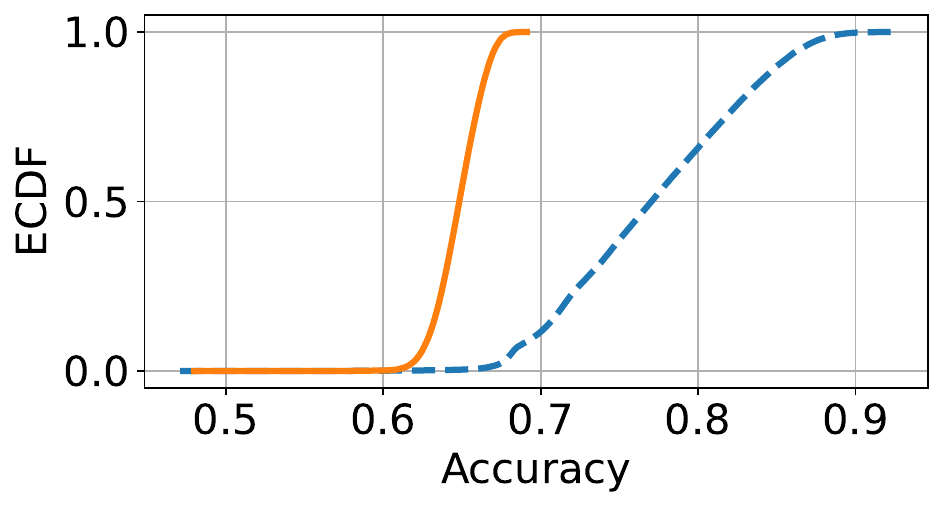}
    \caption{Diabetes.}
\end{subfigure}
~
\begin{subfigure}[b]{0.32\textwidth}
    \centering
    \includegraphics[width=0.98\columnwidth]{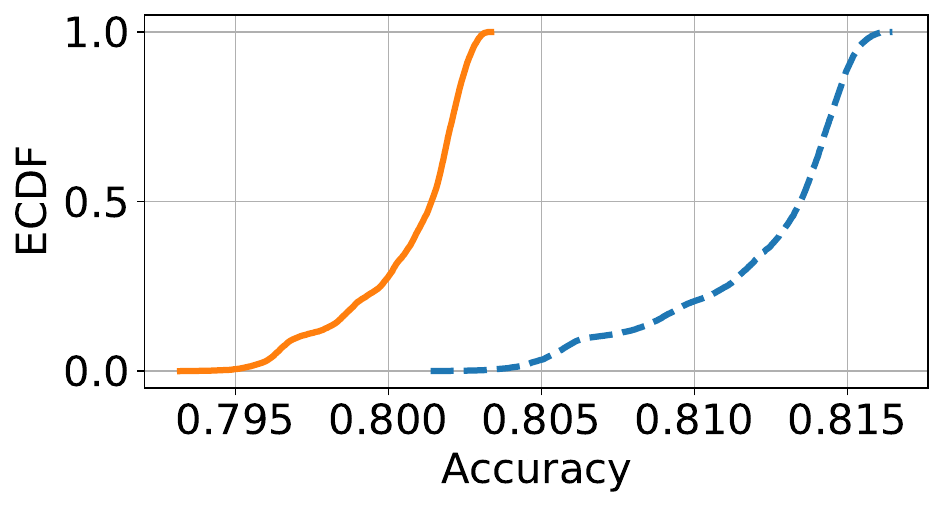}
    \caption{Criteo.}
\end{subfigure}
\caption{The empirical cumulative distribution function (ECDF) of the training and validation AUC recorded across all architectures in \nasbench.}
\label{fig:nasbench_statistics}
\end{figure*}

\begin{figure*}[t]
\centering
\begin{subfigure}[b]{0.32\textwidth}
    \centering
    \includegraphics[width=0.98\columnwidth]{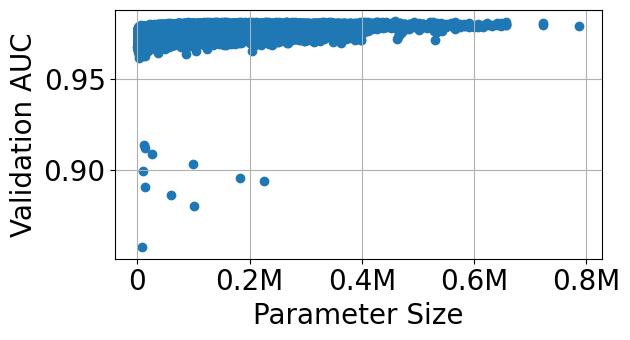}
    \caption{Frappe.}
\end{subfigure}
~
\begin{subfigure}[b]{0.32\textwidth}
    \centering
    \includegraphics[width=0.98\columnwidth]{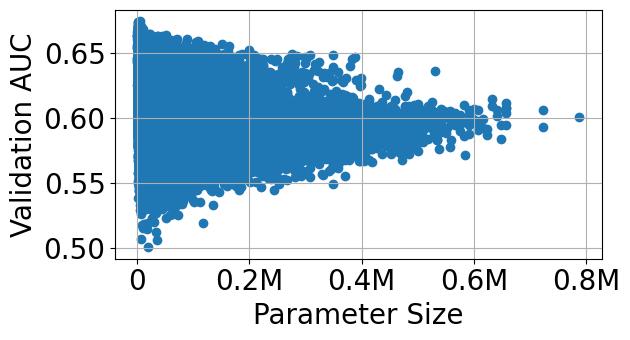}
    \caption{Diabetes.}
\end{subfigure}
~
\begin{subfigure}[b]{0.32\textwidth}
    \centering
    \includegraphics[width=0.98\columnwidth]{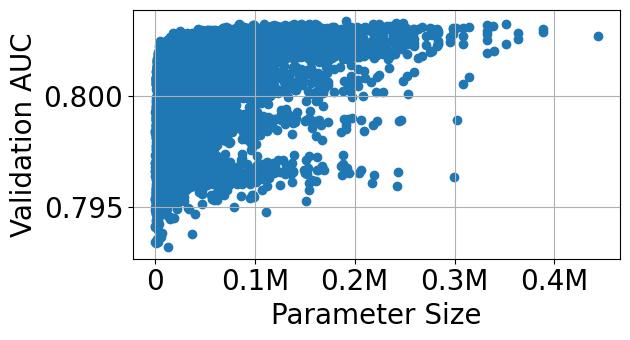}
    \caption{Criteo.}
\end{subfigure}
\caption{Validation AUC vs.\ the number of trainable parameters across all architectures in the search space.}
\label{fig:parameter_auc}
\end{figure*}

\begin{figure*}[t]
\centering
\begin{subfigure}[b]{1\textwidth}
    \centering
    \hspace{1mm}
    \includegraphics[width=0.6\columnwidth]{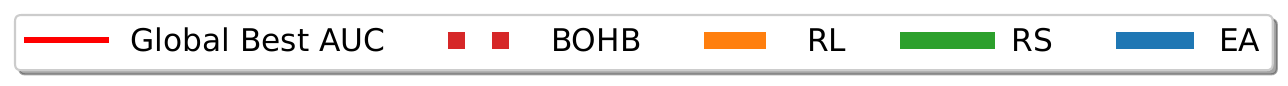}
\end{subfigure}
\begin{subfigure}[b]{0.32\textwidth}
    \centering
    \includegraphics[width=0.98\columnwidth]{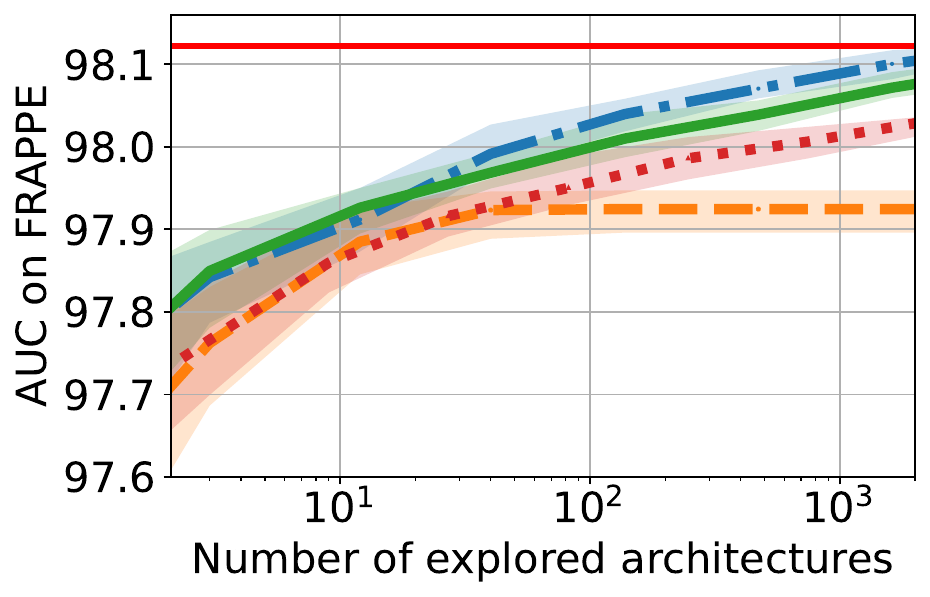}
    \caption{Frappe.}
\end{subfigure}
~
\begin{subfigure}[b]{0.32\textwidth}
    \centering
    \includegraphics[width=0.98\columnwidth]{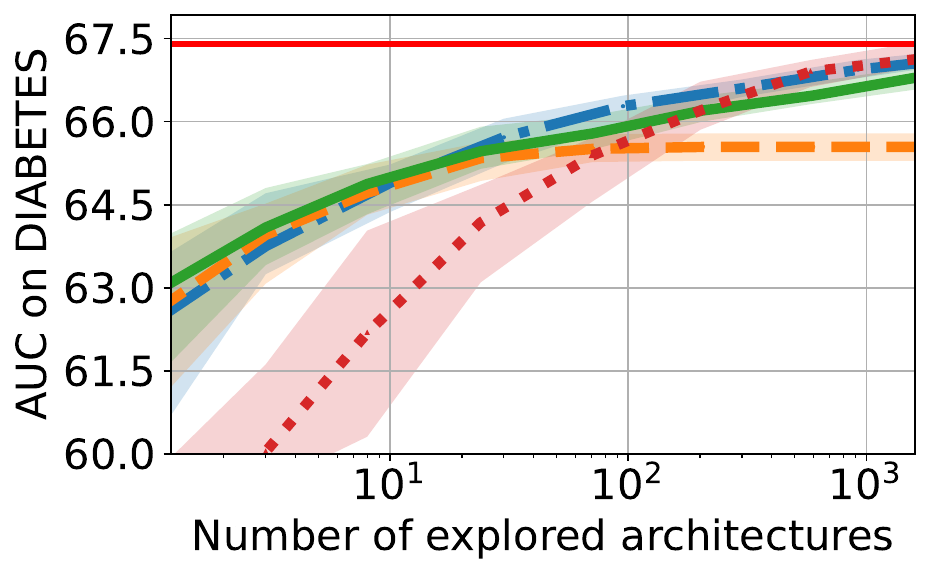}
    \caption{Diabetes.}
\end{subfigure}
~
\begin{subfigure}[b]{0.32\textwidth}
    \centering
    \includegraphics[width=0.98\columnwidth]{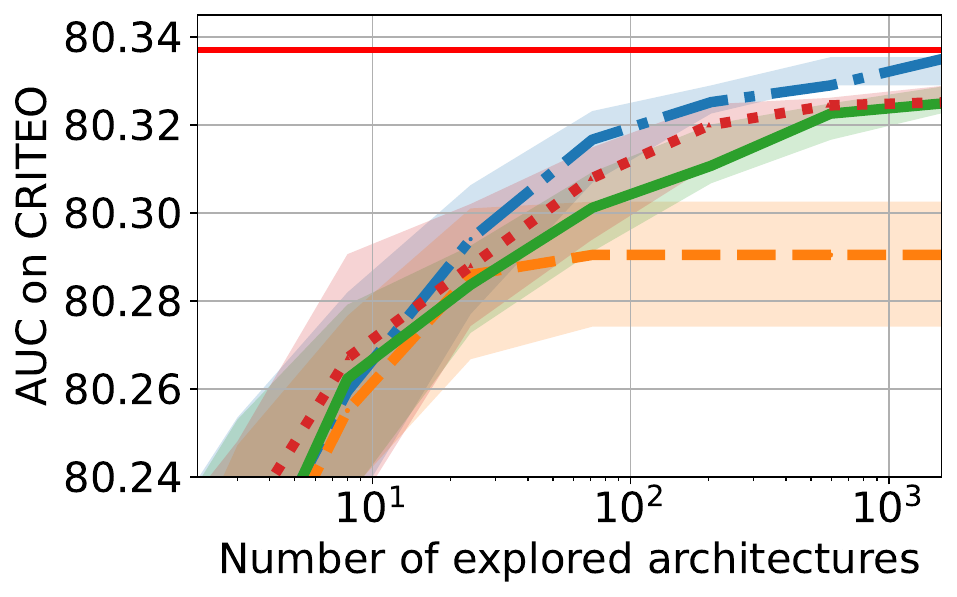}
    \caption{Criteo.}
\end{subfigure}
\caption{Benchmarking four search strategies on \nasbench. The x-axis denotes the number of explored architectures, and the y-axis denotes the median value of the best validation AUC achieved across 100 runs.}
\label{fig:benchmark_ss}
\end{figure*}

Accordingly, the search space size is $|\mathcal{H}|^{L}$.
For Criteo, the space is smaller with only $10^4 = 10{,}000$ architectures, while the best-performing architecture found already achieves performance comparable to prior reported results under similar settings.

We characterize the performance landscape of the \nasbench search space from two complementary perspectives:
(1) the overall distribution of achievable performance across all architectures, and
(2) the relationship between model scale (parameter count) and validation performance.
Figure~\ref{fig:nasbench_statistics} reports the empirical cumulative distribution function (ECDF) of the recorded training and validation AUC across all architectures, while Figure~\ref{fig:parameter_auc} correlates each architecture's parameter count with its validation AUC.

From Figure~\ref{fig:nasbench_statistics}, median validation AUCs are 0.9772, 0.6269, and 0.8014 for Frappe, Diabetes, and Criteo, while optimal architectures achieve 0.9814, 0.6750, and 0.8033, respectively.
These results are consistent with the performance benchmarks reported in prior work~\cite{arm_net,aucref}, validating our training configurations and confirming that \nasbench captures a meaningful spectrum of architecture qualities.
Moreover, Figure~\ref{fig:parameter_auc} shows that the parameter count does not strongly correlate with validation AUC, suggesting that simply scaling the model is insufficient to reliably improve tabular performance.
Together, these observations motivate topology searching (i.e., allocating widths across layers) to identify high-performing architectures and justify the necessity of NAS for tabular DNNs.

\subsection{Benchmarking Search Strategies on \nasbench}
\label{app:nasbench_search_strategy}

Finally, we benchmark four representative search strategies on \nasbench, serving as reference baselines for evaluating future NAS algorithms on our datasets.
The four search strategies include Random Search (RS), Evolutionary Algorithm (EA), Reinforcement Learning (RL), and Bayesian Optimization with HyperBand (BOHB).
All strategies explore the same search space defined in Appendix~\ref{app:nasbench_space} and use training-based evaluation by directly querying the validation AUC from \nasbench.

For EA, we set the population size to 10 and the sample size to 3.
In the RL setup, we employ a categorical distribution for each hidden-layer size and optimize the probabilities using policy gradient methods.
For BOHB, we follow the standard setup that couples Bayesian optimization with HyperBand-style resource allocation.

Figure~\ref{fig:benchmark_ss} reports the anytime performance of these strategies.
The x-axis denotes the number of explored architectures, and the y-axis denotes the median of the best validation AUC achieved across 100 runs.

With \nasbench, NAS algorithms can significantly reduce search times to seconds by querying recorded evaluations.
For example, each of the four strategies can explore over 1k architectures in around 15 seconds in our implementation by querying validation AUCs, highlighting the benefit of recorded evaluations.
As shown in Figure~\ref{fig:benchmark_ss}, EA targets high-performing architectures after exploring around $10^3$ candidates and is consistently sample-efficient across datasets.
We therefore adopt the evolutionary algorithm (EA) as the search strategy in \framework.

\subsection{Generalization to Additional Search Spaces: ResDNN and BlockMixed}
\label{app:new_search_spaces}

\begin{figure}[t]
  \begin{center}
    \includegraphics[width=1\columnwidth]{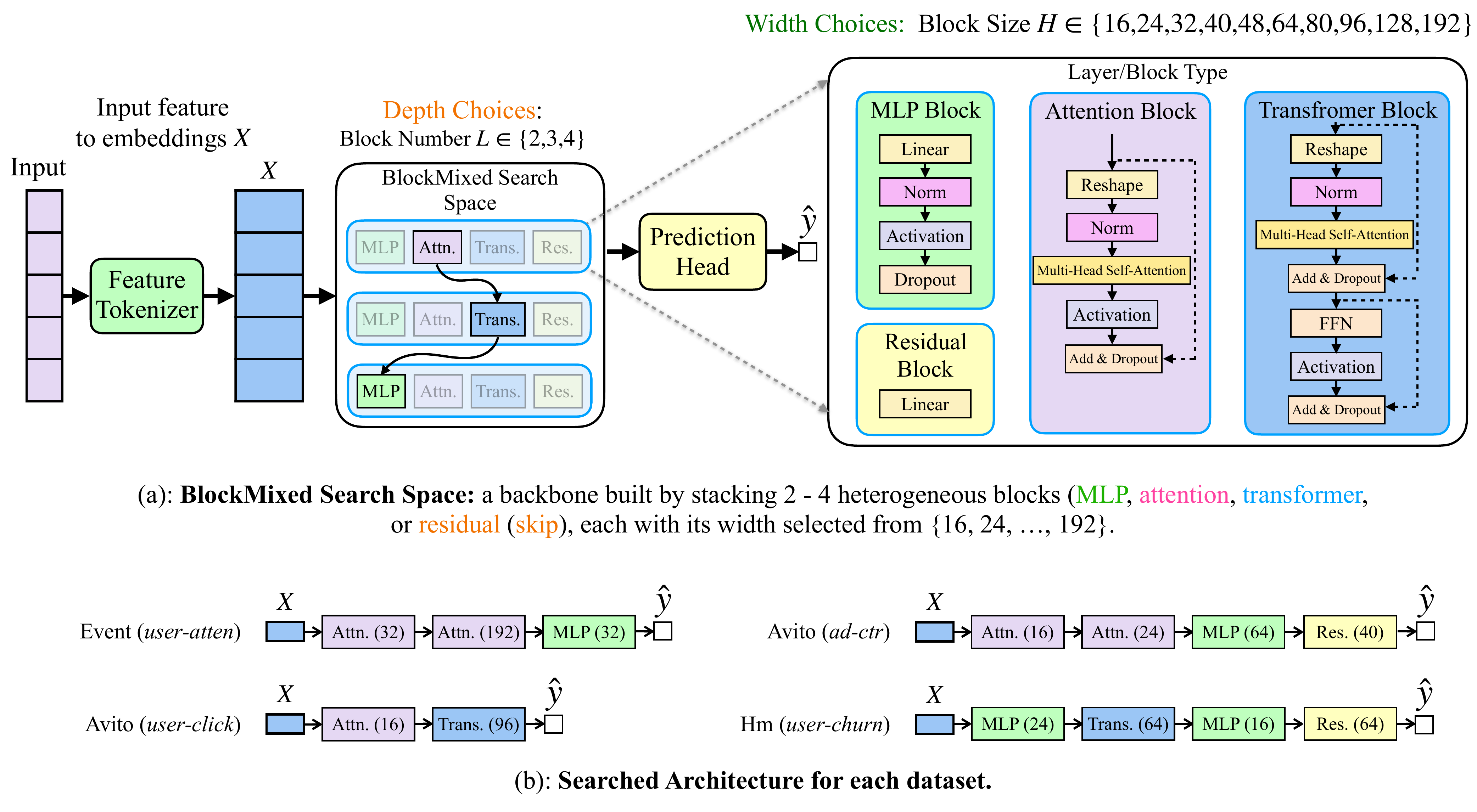}
  \end{center}
  \caption{BlockMixed search space and searched architectures. (a) Each architecture is constructed from heterogeneous block types, including MLP, attention, transformer, and residual blocks, with block-specific widths and training-time design choices. (b) Examples of final architectures selected by \framework after search.}
  \label{fig:blockmixed_space_app}
\end{figure}

To examine whether \newtfmem and \framework generalize beyond the fixed-depth MLP search space used by \nasbench, we introduce two additional search spaces with richer architectural variation.
\begin{itemize}
    \item \textbf{ResDNN} is a DNN-based residual search space with approximately 2.5K architectures.
    It varies network depth, hidden width, and residual skip connections, thereby extending the \nasbench MLP space with variable-depth connectivity patterns.
    \item \textbf{BlockMixed} is a more heterogeneous search space with approximately 1K architectures.
    It is composed of MLP, attention, transformer, and residual blocks, where each block has its own width.
    In addition, BlockMixed varies block-level normalization (LayerNorm or BatchNorm), activation function (ReLU, GELU, or SiLU), and dropout rate (0.0, 0.1, or 0.2).
    Figure~\ref{fig:blockmixed_space_app}(a) illustrates the BlockMixed design.
\end{itemize}

We evaluate these two spaces on four datasets that cover different task types and data regimes.
Specifically, we use Event (\textit{user-attendance}), a relatively small-scale, high-dimensional regression task; Avito (\textit{user-clicks}), a large-scale, low-dimensional classification task; and two additional datasets, Avito (\textit{ad-ctr}) and H\&M (\textit{user-churn}).
This setup allows us to test whether the proxy ranking quality and the downstream search behavior remain robust when both the search space and the dataset distribution change.

Table~\ref{tab:newspace_proxy_avg_rank} compares \newtfmem with the top-5 existing zero-cost proxies from Table~\ref{table:tfmem_srcc}.
For each dataset and search space, we rank the proxies by the absolute SRCC between proxy score and final test performance, and report the average rank across the four datasets.
\newtfmem achieves the best average rank on both ResDNN and BlockMixed, indicating that its neuron-level saliency remains effective under variable-depth residual architectures and heterogeneous block-level architectures.

\begin{table}[t]
\centering
\caption{Average SRCC rank of \newtfmem and the top-5 existing proxies over four datasets for each additional search space. Proxies are ranked by $|\mathrm{SRCC}|$ on each dataset; a lower rank indicates stronger correlation with final architecture performance.}
\label{tab:newspace_proxy_avg_rank}
\resizebox{0.75\columnwidth}{!}{
{\renewcommand{\arraystretch}{1.05}
\begin{tabular}{l|cccccc}
\toprule
\textbf{Search Space} & \textbf{SynFlow} & \textbf{SNIP} & \textbf{NTKCond} & \textbf{NASWOT} & \textbf{NTKTrace} & \textbf{\newtfmem} \\
\midrule
ResDNN     & 2.25 & 4.00 & 5.25 & 2.75 & 5.75 & \best{1.00} \\
BlockMixed & 4.00 & 3.50 & 3.00 & 5.50 & 3.75 & \best{1.25} \\
\bottomrule
\end{tabular}
}}
\end{table}

\begin{figure*}[t]
\centering
\includegraphics[width=0.24\textwidth]{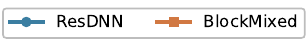}

\vspace{0.2em}

\begin{minipage}[t]{0.24\textwidth}
\centering
\includegraphics[width=\linewidth]{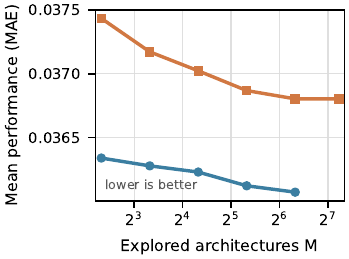}\\[-0.3em]
{\scriptsize (a) Avito \textit{ad-ctr}}
\end{minipage}
\hfill
\begin{minipage}[t]{0.24\textwidth}
\centering
\includegraphics[width=\linewidth]{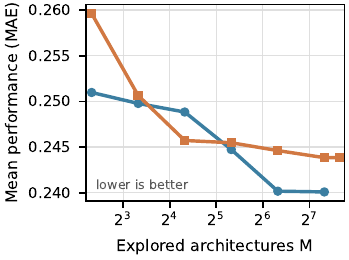}\\[-0.3em]
{\scriptsize (b) Event \textit{user-attend}}
\end{minipage}
\hfill
\begin{minipage}[t]{0.24\textwidth}
\centering
\includegraphics[width=\linewidth]{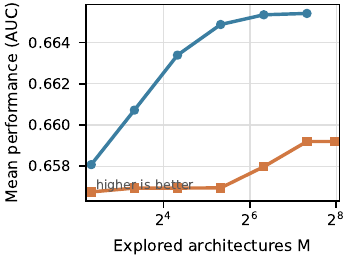}\\[-0.3em]
{\scriptsize (c) Avito \textit{user-clicks}}
\end{minipage}
\hfill
\begin{minipage}[t]{0.24\textwidth}
\centering
\includegraphics[width=\linewidth]{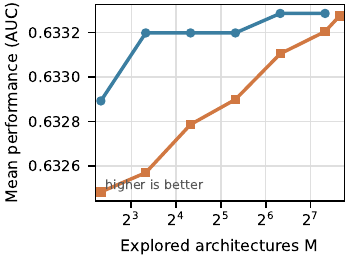}\\[-0.3em]
{\scriptsize (d) H\&M \textit{user-churn}}
\end{minipage}

\caption{\textbf{Search performance of \framework on two additional search spaces.}
We report the mean best-so-far performance over five random seeds as the number of explored architectures $M$ increases.}
\label{fig:newspace_search_performance_app}
\end{figure*}

Figure~\ref{fig:newspace_search_performance_app} visualizes the search behavior of \framework on these additional spaces.
As the number of explored architectures $M$ increases, \framework generally improves the best-so-far architecture quality across both ResDNN and BlockMixed, showing that the search process remains effective beyond the original \nasbench space.
For BlockMixed, examples of the architectures selected at the end of the search are shown in Figure~\ref{fig:blockmixed_space_app}(b).

Overall, the results across \nasbench, ResDNN, and BlockMixed show that \framework is not tied to a single fixed-depth MLP search space.
Instead, \newtfmem continues to provide useful architecture rankings in more diverse spaces.

\section{Theoretical Analysis of \newtfmem}
\label{app:theoretical_analysis}

As defined in Section~\ref{sec:ptproxy}, the neuron saliency of the $n$-th neuron is
\begin{equation}
\nu_n = \left|\frac{\partial \mathcal{L}}{\partial z_n}\right| \odot z_n,
\end{equation}
where $z_n=\sigma(h_n)$ is the post-activation value and
\begin{equation}
h_n=\sum_u \mathbf{w}^{in}_{nu} z_u + b
\end{equation}
is the pre-activation. Here $\mathbf{w}^{in}_{nu}$ denotes the incoming weight from neuron $u$ in the previous layer to neuron $n$.

We further denote $\mathbf{w}^{out}_{vn}$ as the outgoing weight from neuron $n$ to neuron $v$ in the next layer, whose pre-activation is
\begin{equation}
h_v=\sum_k \mathbf{w}^{out}_{vk} z_k + b.
\end{equation}
By the chain rule,
\begin{equation}
\frac{\partial \mathcal{L}}{\partial z_n}
= \sum_{v}\frac{\partial \mathcal{L}}{\partial h_v}\frac{\partial h_v}{\partial z_n}.
\end{equation}
Since $\frac{\partial h_v}{\partial z_n}=\mathbf{w}^{out}_{vn}$, we obtain
\begin{equation}
\frac{\partial \mathcal{L}}{\partial z_n}
= \sum_{v}\frac{\partial \mathcal{L}}{\partial h_v}\mathbf{w}^{out}_{vn}.
\end{equation}
Therefore, the neuron saliency can be written as
\begin{equation}
\nu_n
= \left|\sum_{v}\frac{\partial \mathcal{L}}{\partial h_v}\mathbf{w}^{out}_{vn}\right|\odot z_n.
\end{equation}

For ReLU networks, $z_n=\sigma(h_n)\ge 0$.
Moreover, since $z_n$ is a scalar, the Hadamard product $\odot$ reduces to scalar multiplication, yielding
\begin{equation}\label{eq:10}
\nu_n
= \left|\sum_{v}\frac{\partial \mathcal{L}}{\partial h_v}\mathbf{w}^{out}_{vn} z_n\right|
= \left|\sum_{v}\left(\frac{\partial \mathcal{L}}{\partial h_v} z_n\right)\mathbf{w}^{out}_{vn}\right|.
\end{equation}

Next, note that
\begin{equation}\label{eq:eq_zn}
\frac{\partial h_v}{\partial \mathbf{w}^{out}_{vn}} = z_n,
\end{equation}
which implies $\frac{\partial \mathcal{L}}{\partial \mathbf{w}^{out}_{vn}}
= \frac{\partial \mathcal{L}}{\partial h_v}\frac{\partial h_v}{\partial \mathbf{w}^{out}_{vn}}
= \frac{\partial \mathcal{L}}{\partial h_v} z_n$.
Substituting into Eq.~\ref{eq:10}, we obtain
\begin{equation}\label{eq:11}
\nu_n
= \left|\sum_{v}\left(\frac{\partial \mathcal{L}}{\partial h_v}
\frac{\partial h_v}{\partial \mathbf{w}^{out}_{vn}}\right)\mathbf{w}^{out}_{vn}\right|
= \left|\sum_{v}\frac{\partial \mathcal{L}}{\partial \mathbf{w}^{out}_{vn}} \mathbf{w}^{out}_{vn}\right|.
\end{equation}

Evidently, the term $\frac{\partial \mathcal{L}}{\partial \mathbf{w}^{out}_{vn}} \mathbf{w}^{out}_{vn}$ matches the notion of synaptic saliency used in SynFlow~\cite{synflow}, which was originally introduced to quantify the importance of individual parameters.
Eq.~\ref{eq:11} shows that the neuron saliency of neuron $n$ is the absolute value of the aggregated SynFlow-style saliency over all its outgoing parameters.
Compared with SynFlow, \newtfmem performs aggregation at the neuron level, and the absolute value prevents sign cancellation across outgoing connections, which leads to a more consistent proxy for identifying promising DNN architectures.

Moreover, for homogeneous activations we have $\sigma(\alpha h)=\alpha\sigma(h)$ for $\alpha>0$, and thus one can write $z_n=\sigma(h_n)=\sigma'(h_n)\,h_n$ almost everywhere (e.g., for ReLU away from the kink at $h_n=0$).
For a fixed neuron $n$, this yields
\begin{equation}\label{eq:12}
\begin{aligned}
\nu_n
&= \left|\frac{\partial \mathcal{L}}{\partial z_n}\right|\odot z_n
= \left|\frac{\partial \mathcal{L}}{\partial h_n}\right|\odot h_n
= \left|\frac{\partial \mathcal{L}}{\partial h_n} h_n\right| \\
&= \left|\frac{\partial \mathcal{L}}{\partial h_n}\left(\sum_u \mathbf{w}^{in}_{nu} z_u + b\right)\right| \\
&= \left|\sum_u \left(\frac{\partial \mathcal{L}}{\partial h_n}\mathbf{w}^{in}_{nu} z_u\right) + \frac{\partial \mathcal{L}}{\partial h_n} b\right| \\
&= \left|\sum_u \left(\frac{\partial \mathcal{L}}{\partial h_n}
\frac{\partial h_n}{\partial \mathbf{w}^{in}_{nu}}\,\mathbf{w}^{in}_{nu}\right)
+ \frac{\partial \mathcal{L}}{\partial h_n} b\right| \\
&= \left|\sum_u \left(\frac{\partial \mathcal{L}}{\partial \mathbf{w}^{in}_{nu}}\mathbf{w}^{in}_{nu}\right)
+ \frac{\partial \mathcal{L}}{\partial h_n} b\right|.
\end{aligned}
\end{equation}

The term $\frac{\partial \mathcal{L}}{\partial \mathbf{w}^{in}_{nu}}\mathbf{w}^{in}_{nu}$ is again a synaptic-saliency form for the incoming parameters of neuron $n$.
Therefore, $\nu_n$ aggregates synaptic saliency over both outgoing (Eq.~\ref{eq:11}) and incoming (Eq.~\ref{eq:12}) connections at a neuron-wise granularity, which is well suited for capturing complex feature interactions in tabular models.

Finally, we perform the weighted aggregation of neuron saliency across neurons and samples to obtain the \newtfmem score for an architecture $a$:
\begin{equation}
s_a
= \sum_{i=1}^{B}\sum_{n=1}^{N} \frac{\mathcal{K}_{l(n)}}{d_{l(n)}} \nu_{in}
= \sum_{i=1}^{B}\sum_{n=1}^{N}\frac{\mathcal{K}_{l(n)}}{d_{l(n)}}\left|\frac{\partial \mathcal{L}}{\partial z_{in}}\right|\odot z_{in},
\end{equation}
where $l(n)$ denotes the layer containing neuron $n$.
This provides a training-free proxy that captures both trainability and expressivity signals at initialization.

\section{Extended Experiments}
\label{app:extended_exp}

This section provides additional empirical evidence to complement the main results.
We organize the experiments into four parts: 
(i) ablations of \newtfmem, 
(ii) ablations of the two-phase design and the budget-aware coordinator, 
(iii) comparisons with additional baselines, and 
(iv) a broader comparison and analysis of different \tfmem[s].

\subsection{Ablation Studies of \newtfmem}
\label{app:abs_tfmem}

In this section, we present ablation studies that justify key design choices in computing \newtfmem at initialization.
We focus on how implementation details (e.g., parameter positivity, initialization, batch size, and recalibration weights) affect the correlation between the \newtfmem score and the true architecture performance.

\highlight{Impacts of Parameter Positivity.}
As neuron saliency is computed at architecture initialization (Section~\ref{sec:filter_phase}), we examine the impact of the parameter sign on the effectiveness of \newtfmem.
Specifically, we score each architecture using either the default parameters or their absolute values, and compare the resulting correlations with architecture AUC.
For consistency, we fix the batch size to $B=32$ and use Xavier initialization.
Table~\ref{table:tfmem_abs_para_positive} shows that enforcing parameter positivity significantly improves correlation on all datasets, with the largest gain on Criteo.
Therefore, we set parameters to their absolute values before computing \newtfmem.

\begin{table}[H]
    \centering
    \caption{Impacts of Parameter Positivity.}
        \begin{tabular}{c|cccc}
        \toprule[1.5pt]
            Dataset & Frappe & Diabetes & Criteo \\
            \midrule[0.5pt]
            \newtfmem with positive $\mathbf{w}$ & $0.8364$ & $0.7124$ & $0.8978$  \\
            \newtfmem with default $\mathbf{w}$ & $0.5175$ & $0.5901$ & $0.5020$\\
        \bottomrule[1.5pt]
        \end{tabular}
    \label{table:tfmem_abs_para_positive}
\end{table}

\highlight{Impacts of Initialization Method.}
Based on the advantage of enforcing parameter positivity, we further examine how initialization affects the \newtfmem correlation.
With $B=32$, we compare LeCun~\cite{lecun2002efficient}, Xavier~\cite{glorot2010understanding}, and He~\cite{he2015delving} initializations.
Table~\ref{table:tfmem_abs_init} shows that He initialization consistently performs best on Criteo and Diabetes, which is expected since He is designed for ReLU networks and our search space backbone uses ReLU (Section~\ref{sec:space}).
We therefore adopt the He initialization as the default.

\begin{table}[h]
    \centering
    \caption{Impacts of Initialization method.}
        \begin{tabular}{c|cccc}
        \toprule[1.5pt]
            Dataset & LeCun~\cite{lecun2002efficient} & Xavier~\cite{glorot2010understanding} & He~\cite{he2015delving} \\
            \midrule[0.5pt]
            Frappe & 0.8175 & 0.8364 & 0.8150 \\
            Diabetes & 0.7335 & 0.7124 & 0.7336 \\
            Criteo &  0.8823 & 0.8978 & 0.9005 \\
        \bottomrule[1.5pt]
        \end{tabular}
    \label{table:tfmem_abs_init}
\end{table}

\highlight{Impacts of Batch Size $B$.}
We evaluate the sensitivity of \newtfmem to different batch sizes, varying $B$ from 4 to 128.
For each setting, we fix the He initialization and repeat the experiment five times, reporting the median correlation.
Table~\ref{table:tfmem_abs_batch} shows that batch size has a limited influence on the correlation, suggesting that \newtfmem can be computed efficiently using small batches.

\begin{table}[H]
    \centering
    \caption{Impacts of Batch Size.}
        \begin{tabular}{c|ccccccc}
        \toprule[1.5pt]
            Dataset & $B$ = 4 & $B$ = 8 & $B$ = 16 & $B$ = 32 & $B$ = 64 & $B$ = 128\\
            \midrule[0.5pt]
            
            Frappe & $0.8154$ & $0.8152$ & $0.8150$ & $0.8150$ & $0.8150$ & $0.8149$ \\
            
            Diabetes & $0.7335$  & $0.7336$ & $0.7335$ & $0.7336$ & $0.7336$ & $0.7336$ \\
            
            Criteo & $0.8990$ & $0.8998$ & $0.9008$ & $0.9005$ & $0.9009$ & $0.9009$ \\
        \bottomrule[1.5pt]
        \end{tabular}
    \label{table:tfmem_abs_batch}
\end{table}

\highlight{Impact of Recalibration Weight of Neuron Saliency.}
\newtfmem aggregates neuron saliency across layers with a recalibration weight $\frac{\mathcal{K}_l}{d_l}$ (Section~\ref{sec:filter_phase}).
We compare \newtfmem with two variants:
(1) recalibrated only by width $\mathcal{K}$, and
(2) recalibrated only by depth $\frac{1}{d_l}$.
Table~\ref{table:tfmem_abs_weight} shows that considering both width and depth consistently yields the highest correlations across datasets.

\begin{table}[H]
    \centering
    \caption{Impacts of Recalibration Weights of Neuron Saliency.}
        \begin{tabular}{c|cccc}
        \toprule[1.5pt]
            Dataset 
            & recalibrated by $\mathcal{K}$ 
            & recalibrated by $\frac{1}{d_l}$   
            & recalibrated by $\frac{\mathcal{K}_l}{d_l}$ \\
            \midrule[0.5pt]
            Frappe & $0.7007$ & $0.7296$ & $0.8155$ \\
            Diabetes & $0.6772$ & $0.6900$ & $0.7336$  \\
            Criteo & $0.6402$ & $0.6414$ & $0.8990$ \\
        \bottomrule[1.5pt]
        \end{tabular}
    \label{table:tfmem_abs_weight}
\end{table}

\subsection{Component Analysis of Two-Phase Design}
\label{app:abs_two_phase}

\begin{figure*}[t]
\centering
\begin{subfigure}[b]{1\textwidth}
    \centering
    \hspace{1mm}
    \includegraphics[width=0.5\columnwidth]{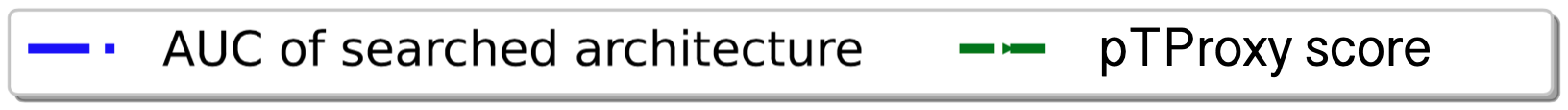}
\end{subfigure}
\begin{subfigure}[b]{0.315\textwidth}
    \centering
    \includegraphics[width=0.98\columnwidth]
    {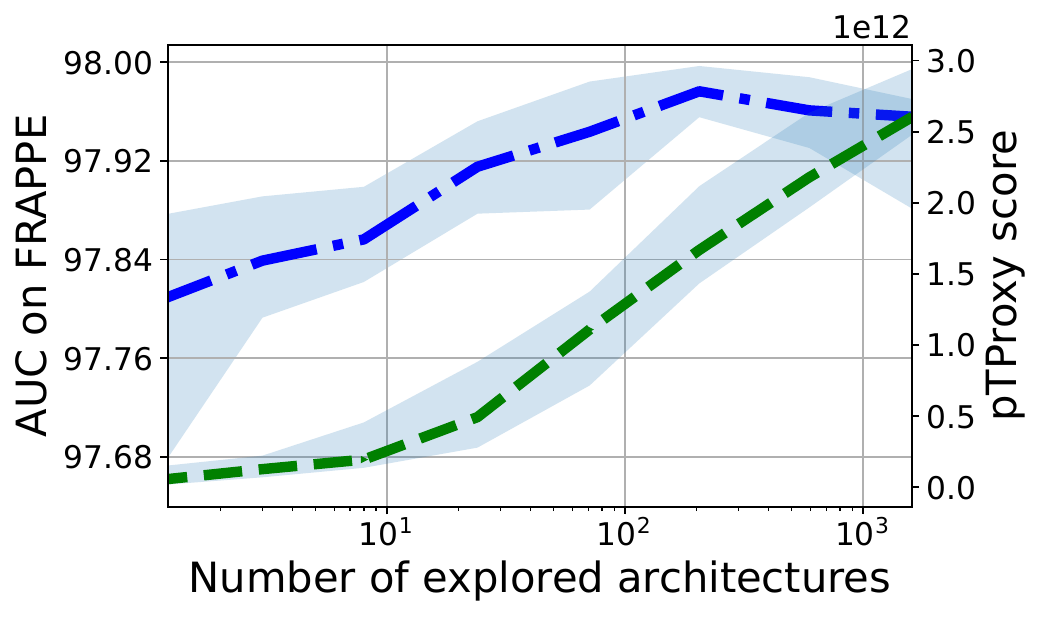}
    \caption{Frappe Dataset.}
\end{subfigure}
~
\begin{subfigure}[b]{0.315\textwidth}
    \centering
    \includegraphics[width=0.98\columnwidth]
    {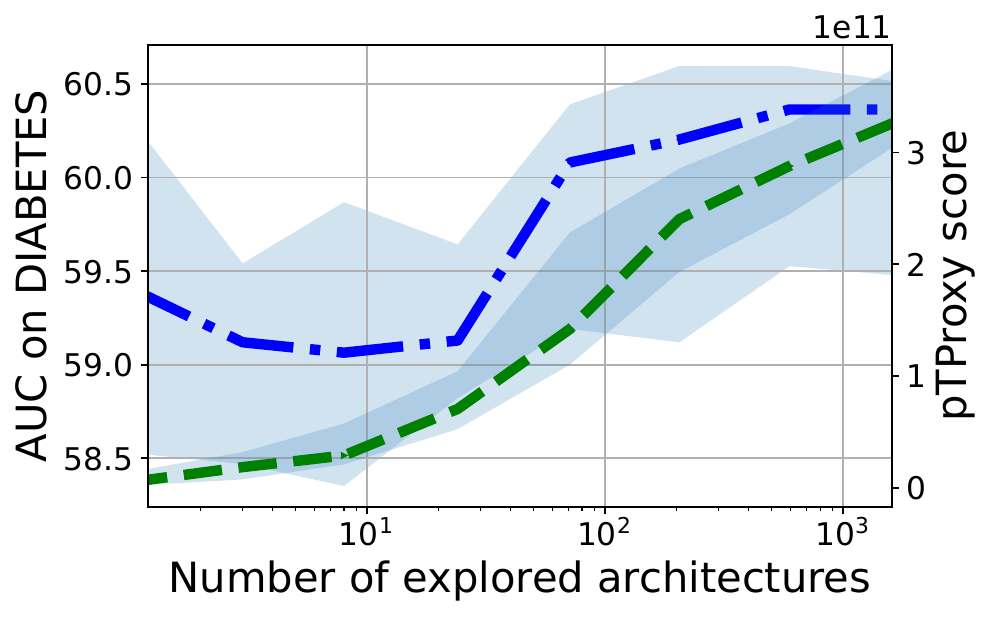}
    \caption{Diabetes Dataset.}
\end{subfigure}
~
\begin{subfigure}[b]{0.325\textwidth}
    \centering
    \includegraphics[width=0.99\columnwidth]
    {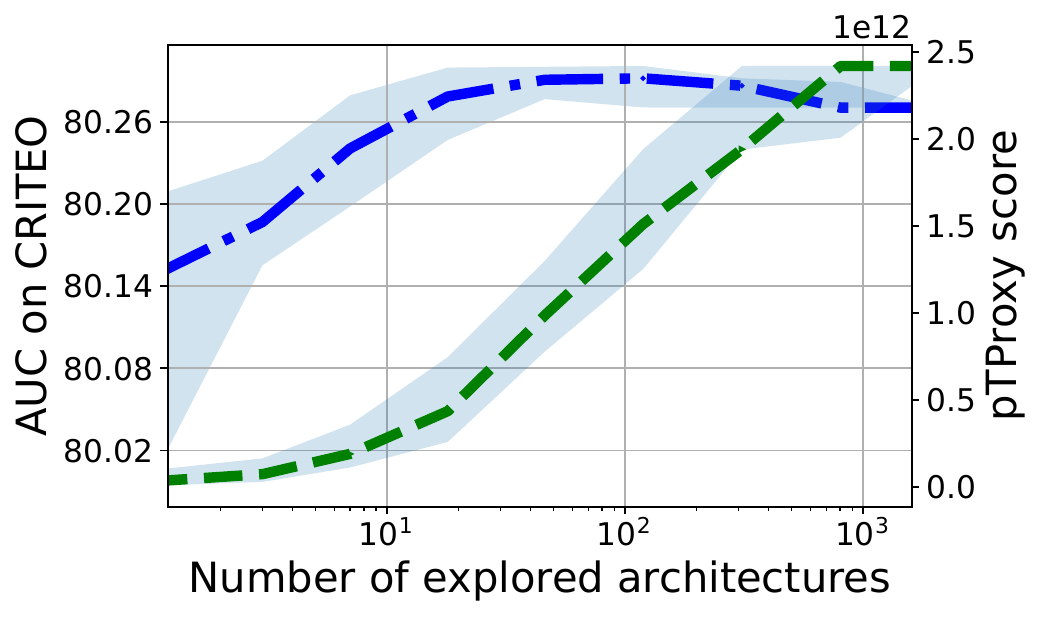}
    \caption{Criteo Dataset.}
\end{subfigure}
\caption{Relationship between the \newtfmem score and validation AUC under \filter-only search as the explored budget increases.}
\label{fig:two_phase_abs_p1}
\end{figure*}

\highlight{Necessity of the \refine Phase}.
To evaluate whether the \filter phase alone can achieve progressive NAS, we run \filter-only search and track the architectures it selects as the budget increases.
Figure~\ref{fig:two_phase_abs_p1} shows that although the search identifies architectures with higher \newtfmem scores when increasing the budget, these higher scores do not consistently translate into higher validation AUC.
In particular, increasing the budget may even lead to selecting architectures with higher \newtfmem scores but inferior AUC.
This confirms that \filter-only search does not reliably satisfy the progressive requirement, and motivates the need for the \refine phase that performs training-based evaluation (Section~\ref{sec:refine_phase}).

\highlight{Noise Impact on \refine Phase}.
Given that \newtfmem scores are imperfect estimates, we investigate whether injecting random noise when selecting the top-$K$ candidates could improve robustness.
We fix the total search time to 20 minutes and vary the \textit{Random Noise Degree} (RND).
For each RND, we keep the top $(1-\text{RND})K$ architectures among the top-$K$ ranked by \newtfmem and randomly replace $\text{RND}\cdot K$ architectures with samples from the search space, then evaluate these $K$ candidates in \refine.
Table~\ref{table:refine_abs_noisy} shows that introducing noise does not improve AUC, and smaller noise yields better performance.
This further confirms the effectiveness of the coordinator in selecting top-$K$ candidates for \refine.

\begin{center}
\begin{minipage}{\textwidth}
    \centering
    \captionof{table}{Impact of noise degrees on the searched AUC. Search for 20 mins.}
    \label{table:refine_abs_noisy}
    \renewcommand{\arraystretch}{0.6}
    \setlength\tabcolsep{3mm}
    \begin{tabular}{c|c|c|c|c|c}
    \toprule[1.5pt]
        & RND = 100\% & RND = 70\% & RND = 50\% & RND = 30\% & RND = 0\% \\
        \midrule[0.5pt]
        AUC & 0.9792 & 0.9794 & 0.9796 & 0.9799 & \textbf{0.9802} \\
    \bottomrule[1.5pt]
    \end{tabular}
\end{minipage}
\end{center}

\subsection{Hyperparameter Analysis of the Coordinator}
\label{app:abs_coord}

\highlight{Sensitivity of $M/K$ and $U$}.
As explained in Section~\ref{sec:coord}, \framework employs a \scheme with joint optimization under \budget.
The primary challenge is choosing $M$ and $K$ given \budget, where $M$ is the number of candidates explored in the \filter phase using \newtfmem and $K$ is the number of promising architectures exploited in the \refine phase by training.
Exploring many architectures while neglecting \refine (e.g., $K=1$) is efficient but can select sub-optimal architectures due to proxy noise, whereas training too many architectures (e.g., $K=M$) violates the progressive requirement.

A second challenge is the trade-off between $K$ and $U$ under a fixed training budget for \refine, where $U$ denotes the computation unit (epochs) used to evaluate each architecture in \refine.
Training each architecture longer improves evaluation fidelity but reduces the number of explored candidates; training shorter enables exploring more candidates but with noisier estimates.

\begin{figure*}[h]
\centering
\begin{subfigure}[b]{0.4\textwidth}
    \centering
    \includegraphics[width=0.99\columnwidth]{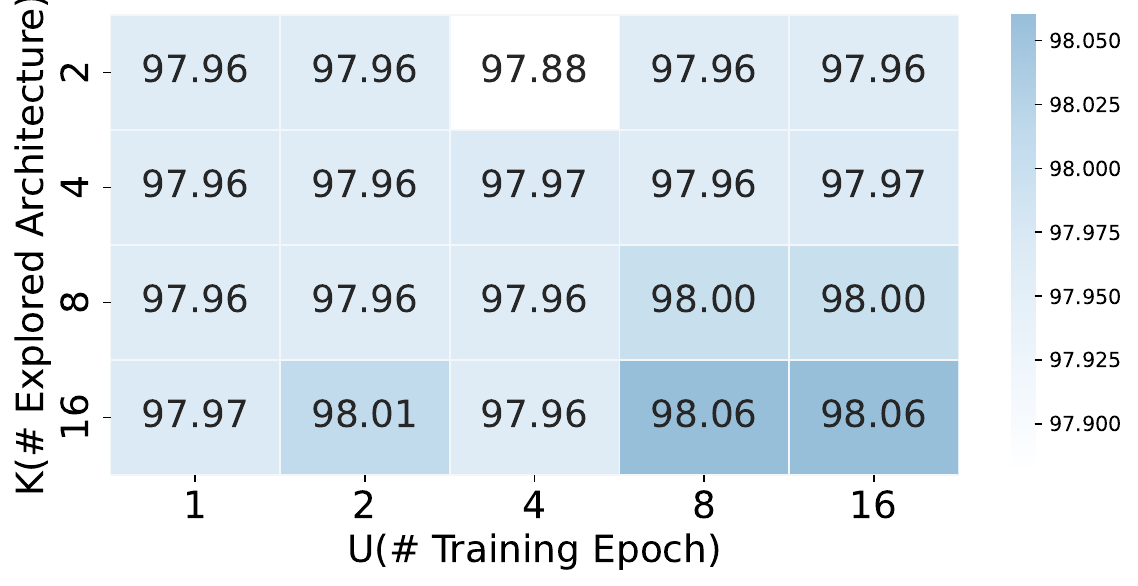}
    \caption{Reached AUC on Frappe Dataset.}
\end{subfigure}
~
\begin{subfigure}[b]{0.4\textwidth}
    \centering
    \includegraphics[width=0.99\columnwidth]{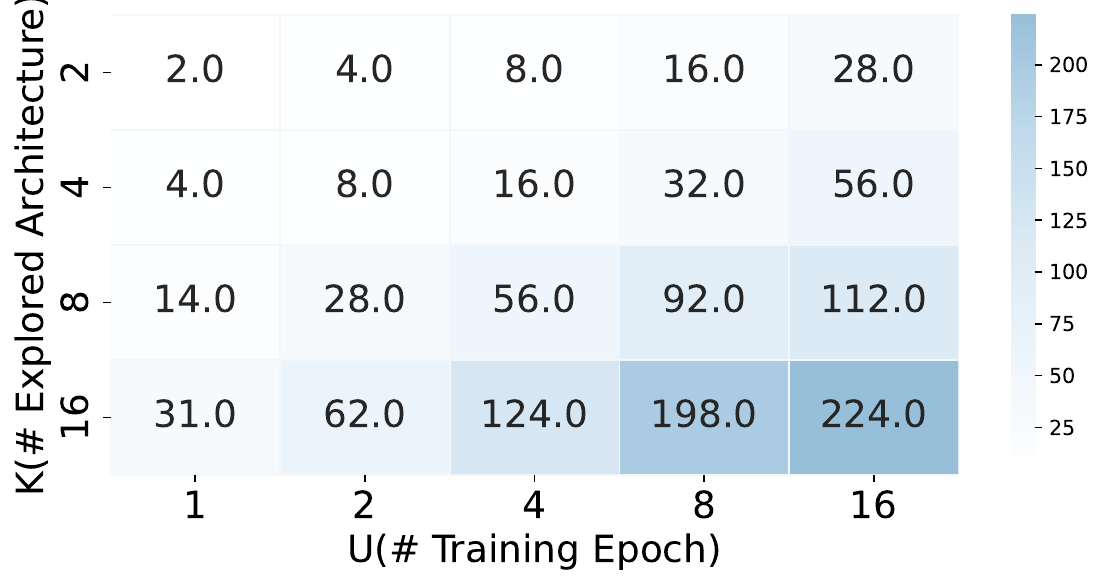}
    \caption{Total Training Epochs on Frappe Dataset.}
\end{subfigure}

\begin{subfigure}[b]{0.4\textwidth}
    \centering
    \includegraphics[width=0.99\columnwidth]{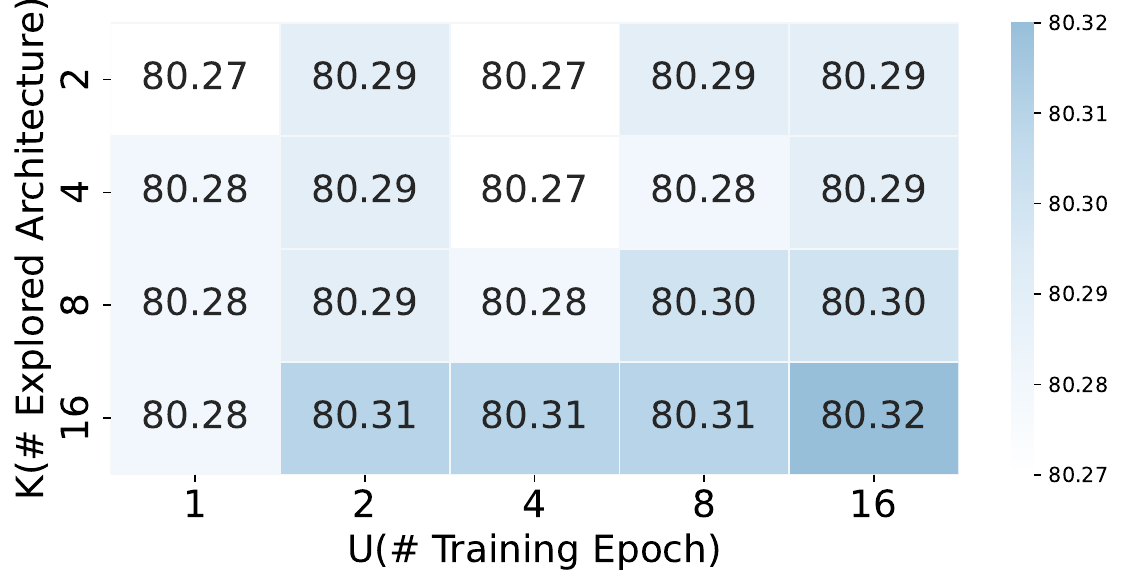}
    \caption{Reached AUC on Criteo Dataset.}
\end{subfigure}
~
\begin{subfigure}[b]{0.4\textwidth}
    \centering
    \includegraphics[width=0.99\columnwidth]{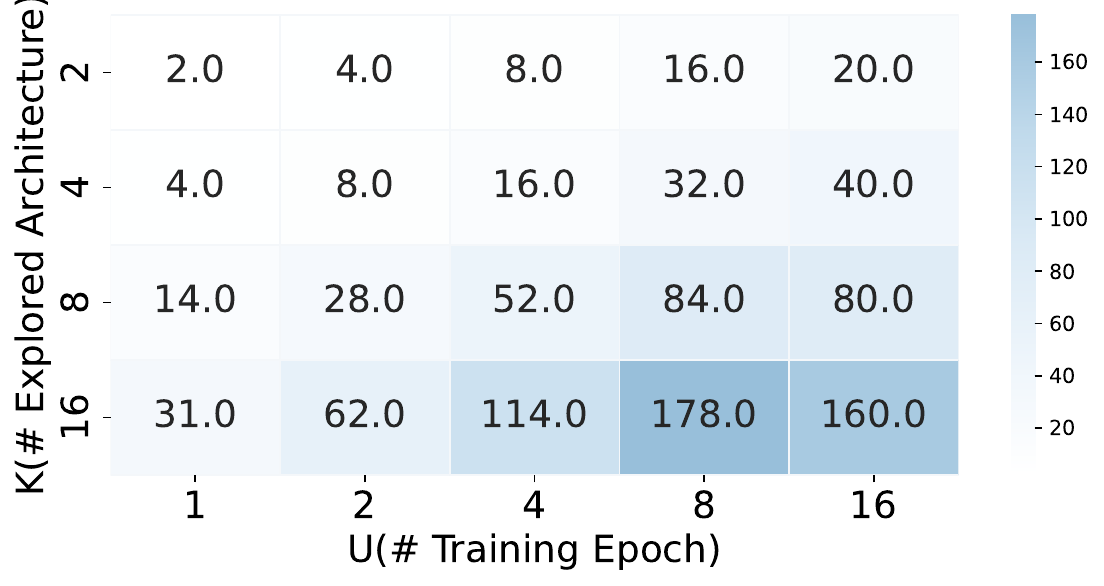}
    \caption{Total Training Epochs on Criteo Dataset.}
\end{subfigure}

\caption{Trade-offs between $K$ and $U$ in the \refine phase.}
\label{fig:coord_abs_ku}
\end{figure*}

Therefore, we first examine the trade-off between $K$ and $U$.
We explore a set of architectures in \filter and vary $(K,U)$ to measure the achieved AUC and the total training epochs in \refine.
Figure~\ref{fig:coord_abs_ku} shows that using a small $U$ (e.g., $U=2$) can improve the final AUC while reducing total training epochs by enabling more candidates to be evaluated.
We therefore set $U=2$ in \framework.

\begin{figure*}[h]
\centering
\begin{subfigure}[b]{0.3\textwidth}
    \centering
    \includegraphics[width=0.98\columnwidth]{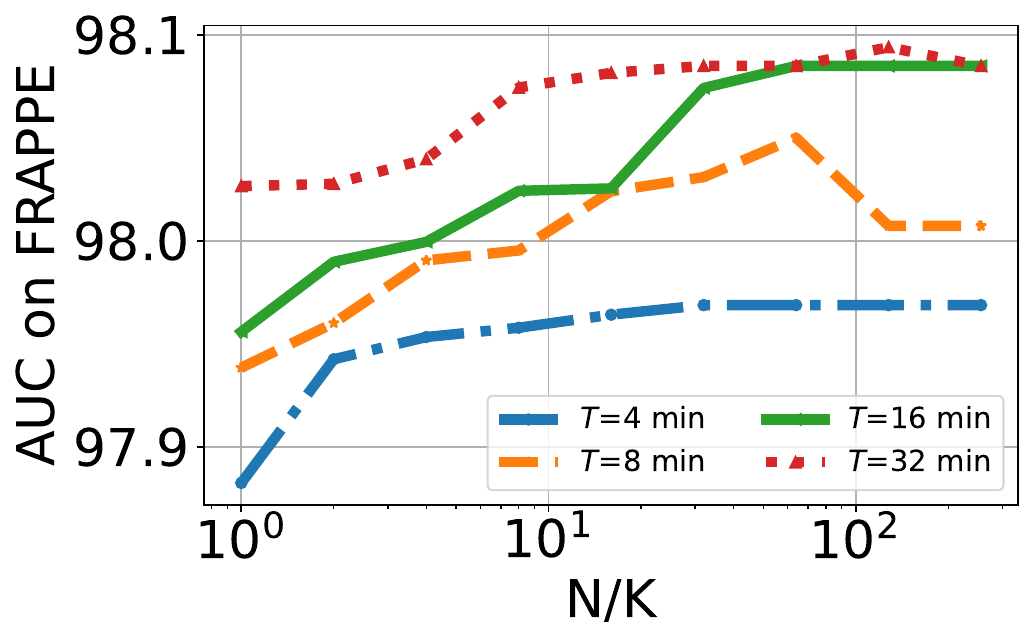}
    \caption{Frappe.}
\end{subfigure}
~
\begin{subfigure}[b]{0.3\textwidth}
    \centering
    \includegraphics[width=0.98\columnwidth]{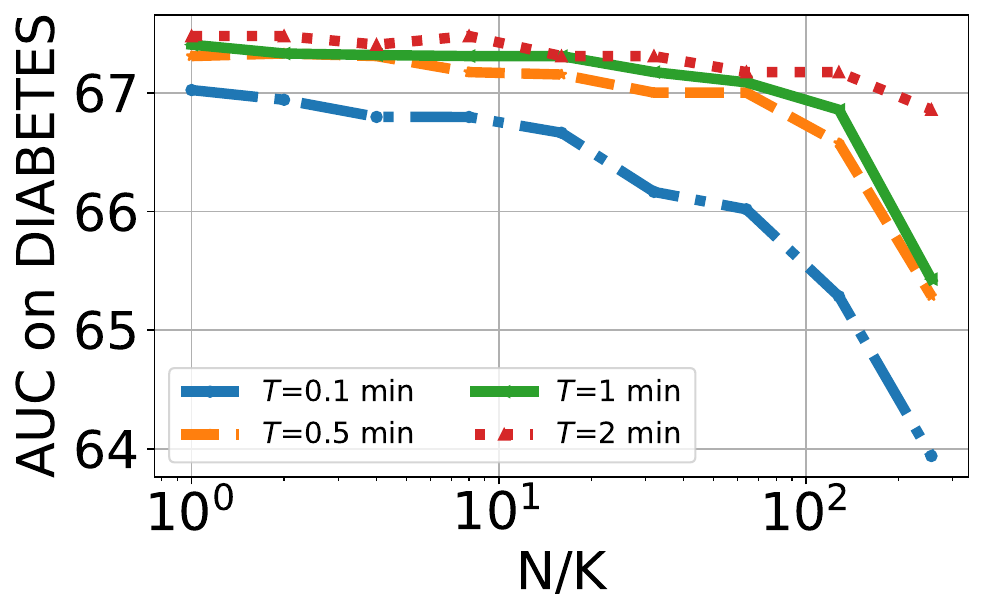}
    \caption{Diabetes.}
\end{subfigure}
~
\begin{subfigure}[b]{0.31\textwidth}
    \centering
    \includegraphics[width=0.99\columnwidth]{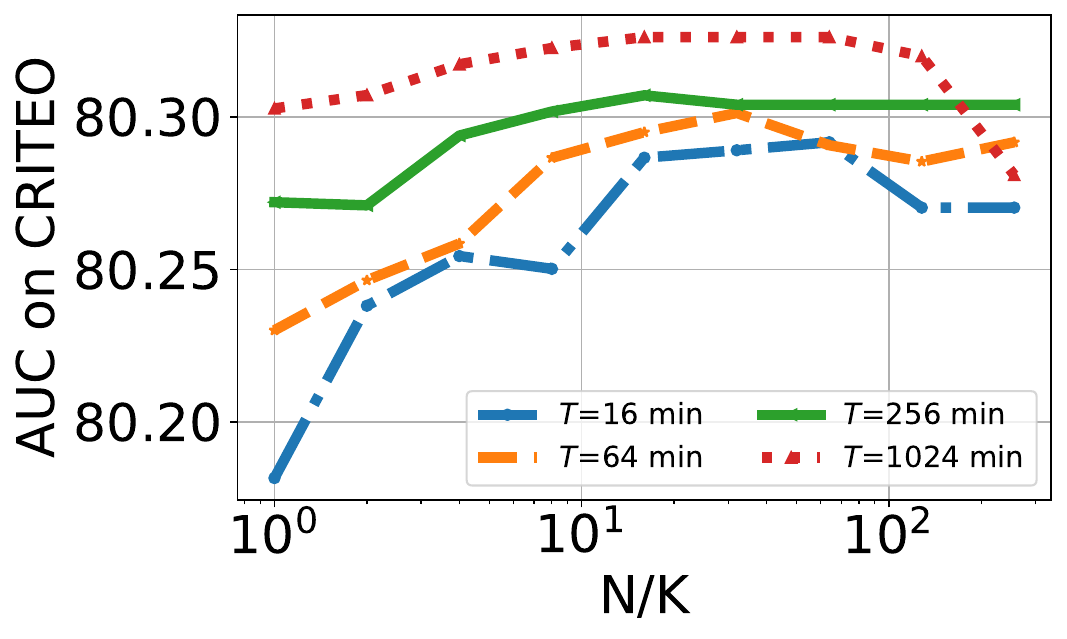}
    \caption{Criteo.}
\end{subfigure}
\caption{Trade-offs between $M$ and $K$ under different \budget.}
\label{fig:coord_abs_nk}
\end{figure*}

We then examine the trade-off between $M$ and $K$ with varying \budget.
Figure~\ref{fig:coord_abs_nk} shows that a ratio $M/K \approx 30$ yields strong performance across different \budget, which is adopted by our coordinator.

\highlight{Necessity of Coordinator}.
To further illustrate the necessity and effectiveness of the coordinator, we fix the search time budget to 100 minutes and compare architectures obtained using different manually set $K$ values with those determined by the coordinator.
Table~\ref{table:coord_abs_k} shows that the choice of $K$ heavily influences the search performance, and the coordinator-selected $K$ achieves the highest AUC.
This confirms that the budget-aware coordinator is effective in holistically optimizing both \filter and \refine phases and is necessary for achieving progressive NAS.

\begin{table}[h]
    \centering
    \captionof{table}{Impact of $K$ on the searched AUC. Search for 100 mins.}
    \renewcommand{\arraystretch}{0.8} 
    \begin{tabular}{c|cccc|c}
        \toprule[1.5pt]
        & K=1 & K=10 & K=100 & K=500 & Coordinator (K=309) \\
        \midrule
        AUC & 0.9789 & 0.9801 & 0.9802 & 0.9791 & \textbf{0.9805} \\
        \bottomrule[1.5pt]
    \end{tabular}
    \label{table:coord_abs_k}
\end{table}


\subsection{Comparison and Analysis of \tfmem[s]}
\label{app:tfmem_compare_table}

\begin{center}
\begin{minipage}{\textwidth}
    \centering
    \footnotesize
    \captionof{table}{Efficiency of \tfmem[s] measured on Frappe dataset (seconds). FC/BC: Forward/Backward computation, N/R: the target AUC is not reached. Search costs are evaluated on Frappe.}
    \label{table:tfmem_ana_time}
    \renewcommand{\arraystretch}{0.5}
    \setlength\tabcolsep{0.8mm}
    \resizebox{0.8\textwidth}{!}{
        \begin{tabular}{c|ccccccccc|c}
        \toprule[1.5pt]
            & \makecell{Grad\\Norm} & \makecell{NAS\\WOT} & \makecell{NTK\\Cond} & \makecell{NTK\\Trace} & \makecell{NTK\\TrAppx} & Fisher & GraSP & SNIP & SynFlow & \noindent\textbf{\newtfmem}\\
    
            \midrule[0.5pt]
            \makecell{Time Complexity} & \makecell{1FC\\1BC} & 1FC & \makecell{1FC\\1BC} & \makecell{1FC\\1BC} & \makecell{1FC\\1BC} & \makecell{1FC\\1BC} & \makecell{1FC\\1BC} & \makecell{1FC\\1BC} & \makecell{1FC\\1BC} & \makecell{2FC\\1BC} \\
            
            \midrule[0.5pt]
            \makecell{Computational Cost} & \makecell{$2.69 \times$ \\ $10^{-3}$} & \makecell{$1.76 \times$ \\$10^{-3}$} & \makecell{$5.75 \times$ \\ $10^{-3}$} & \makecell{$5.66 \times$ \\ $10^{-3}$} & \makecell{$2.85 \times$ \\ $10^{-3}$} & \makecell{$9.59 \times$ \\ $10^{-3}$}  & \makecell{$1.14 \times$ \\ $10^{-2}$} & \makecell{$3.02 \times$ \\ $10^{-3}$} & \makecell{$2.27 \times$ \\ $10^{-3}$} & \makecell{$2.70 \times$ \\ $10^{-3}$} \\
    
            \midrule[0.5pt]
            \makecell{Search Cost \\ for AUC 0.9793 \\ \filter \\ Only } & \makecell{19.9} & \makecell{N/R} & \makecell{N/R} & \makecell{N/R} & \makecell{N/R} & \makecell{N/R}  & \makecell{N/R} & \makecell{18.8} & \makecell{22.1} & \makecell{\textbf{2.8}} \\
    
            \midrule[0.5pt]
            \makecell{Search Cost \\ for AUC 0.9798 \\ \filter+ \\ \refine }  & \makecell{695} & \makecell{1188} & \makecell{N/R} & \makecell{7568} & \makecell{4178} & \makecell{1233}  & \makecell{1141} & \makecell{726} & \makecell{1184} & \makecell{\textbf{62}} \\
        \bottomrule[1.5pt]
        \end{tabular}
    }
\end{minipage}
\end{center}

We report both theoretical complexity and empirical scoring time per architecture on Frappe, as well as search cost to reach a target AUC when using EA guided by different \tfmem[s].
Table~\ref{table:tfmem_ana_time} shows that \newtfmem is computationally efficient and achieves the lowest search cost among the compared proxies.

\subsection{Comparison of \framework with Additional Baselines}
\label{app:more_baseline}

\begin{center}
\begin{minipage}{\textwidth}
    \centering
    \captionof{table}{Target AUC = 0.9798. N/R: the target AUC is not reached. }
    \label{table:more_baseline_search_cost}
    \renewcommand{\arraystretch}{1}
    \setlength\tabcolsep{1.1mm}
    \begin{tabular}{c|c}
    \toprule[1.5pt]
        NAS Approaches & \makecell{Search Cost for AUC 0.9798 (Sec)} \\
        \midrule[0.5pt]
        Training-based (RS)     & 21560  \\
        Training-based (RL)      & N/R \\
        Training-based (EA)      & 8462 \\
        \midrule[0.5pt]
        Training-free (SNIP)      & N/R  \\
        Training-free (NASWOT)       & N/R \\
        Training-free (SynFlow)     & N/R  \\
        Training-free (\newtfmem)     & N/R  \\
        \midrule[0.5pt]
        Warmup (NASWOT)      & 234 \\
        Warmup (SNIP)     & 464  \\
        Warmup (SynFlow)      & 289 \\
        Warmup (\newtfmem)     & \textbf{227}  \\
        \midrule[0.5pt]
        Move-Proposal (NASWOT)     & 9503  \\
        Move-Proposal (SNIP)     & 8659  \\
        Move-Proposal (SynFlow)      & 9106 \\
        Move-Proposal (\newtfmem)     & 6940 \\
        \midrule[0.5pt]
        \filter + \refine (\newtfmem+Full Training)     & 329   \\
        \midrule[0.5pt]
        \framework & \best{62}  \\
    \bottomrule[1.5pt]
    \end{tabular}
\end{minipage}
\end{center}

\subsubsection{Comparison with Different Combinations of Training-Free and Training-Based Methods}
\label{app:more_baseline_combs}

In this section, we compare \framework with representative NAS variants that combine training-free and training-based evaluations.
We include two strategies from prior work~\cite{zero-cost}: the fully decoupled \textit{warmup} strategy and the coupled \textit{move proposal} strategy.
In contrast, \framework employs a decoupled two-phase design: the \filter phase is guided by EA using \tfmem[s], and the \refine phase uses training-based evaluation scheduled by successive halving; the two phases are holistically optimized by a budget-aware coordinator (Section~\ref{sec:coord}) to support progressive NAS.

To empirically compare search efficiency, we evaluate the time required to reach a target AUC under different combinations.
Specifically, we compare: training-based only (RS/RL/EA), training-free only (best-performing \tfmem[s]), and Warmup/Move-Proposal (both using EA as in \framework with best-performing \tfmem[s]).

The results in Table~\ref{table:more_baseline_search_cost} show that:
(1) training-based only and Move-Proposal are the most time-consuming approaches due to repeated costly training;
(2) training-free only fails to reach the target AUC;
(3) \framework reaches the target AUC with substantially reduced time, outperforming Warmup and Move-Proposal;
(4) the two-phase scheme with the coordinator achieves better search performance than other combinations, confirming the necessity of combining training-free and training-based evaluation in a budget-aware manner.

\subsubsection{Comparison with One-Shot NAS Methods}
\label{app:more_baseline_oneshot}

One-shot NAS methods reduce search cost by training a single supernet and evaluating subnets via weight sharing~\cite{DARTS,ENAS}.
Their effectiveness depends on the correlation between the inherited-weight performance estimates and true performance after full training.

Given our DNN-based search space, we train a supernet with four layers, each with a maximum width of 512.
We evaluate subnet performance using inherited weights and compare it with full-training performance on Frappe.
The resulting SRCC is only 0.12, which is substantially lower than \newtfmem (SRCC 0.82), leading to inferior search performance under weight sharing.

\subsection{Visualization of Correlation for \tfmem[s]}
\label{app:more_exp_visual}

\begin{figure*}[!htbp]
\centering
\begin{subfigure}[b]{0.15\textwidth}
    \centering
    \includegraphics[width=\linewidth]{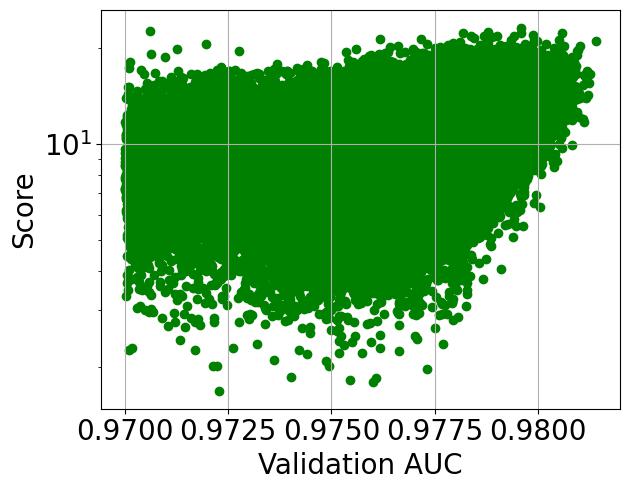}
    \caption{Frappe}
\end{subfigure}\hspace{0.6mm}
\begin{subfigure}[b]{0.15\textwidth}
    \centering
    \includegraphics[width=\linewidth]{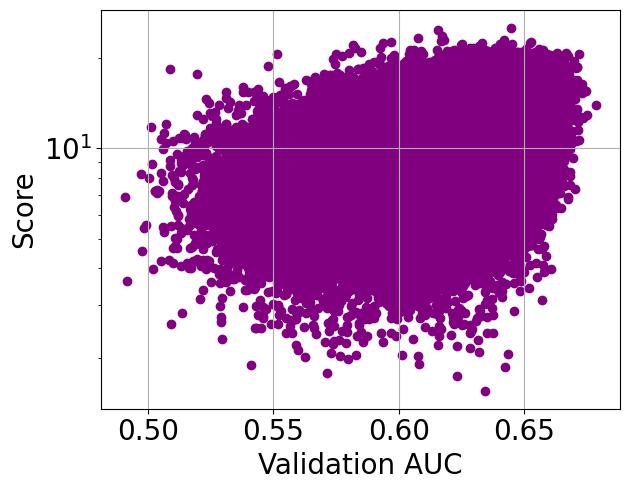}
    \caption{Diabetes}
\end{subfigure}\hspace{0.6mm}
\begin{subfigure}[b]{0.15\textwidth}
    \centering
    \includegraphics[width=\linewidth]{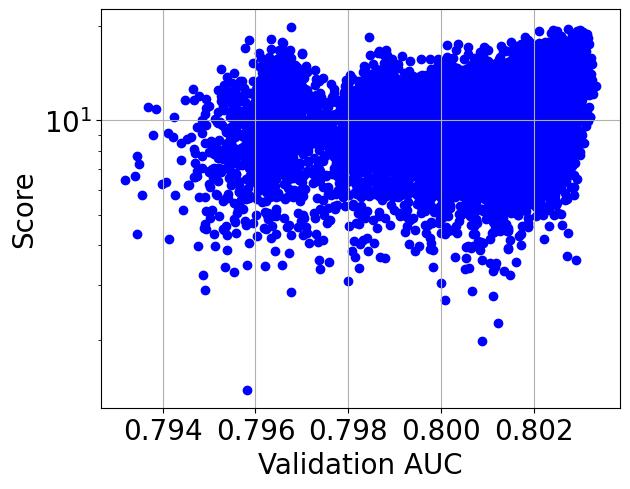}
    \caption{Criteo}
\end{subfigure}\hspace{2.0mm}
\begin{subfigure}[b]{0.15\textwidth}
    \centering
    \includegraphics[width=\linewidth]{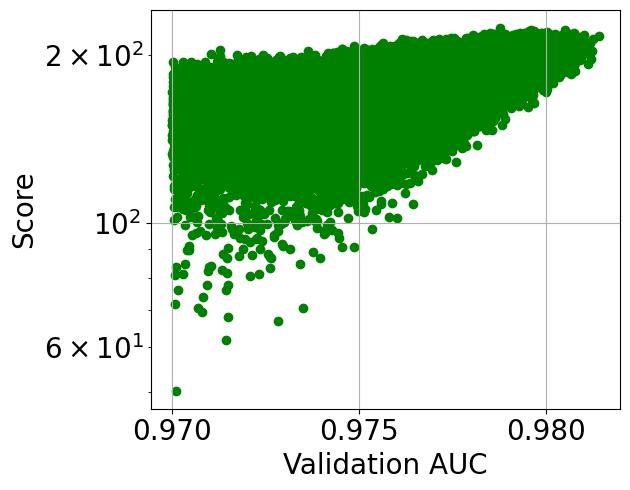}
    \caption{Frappe}
\end{subfigure}\hspace{0.6mm}
\begin{subfigure}[b]{0.15\textwidth}
    \centering
    \includegraphics[width=\linewidth]{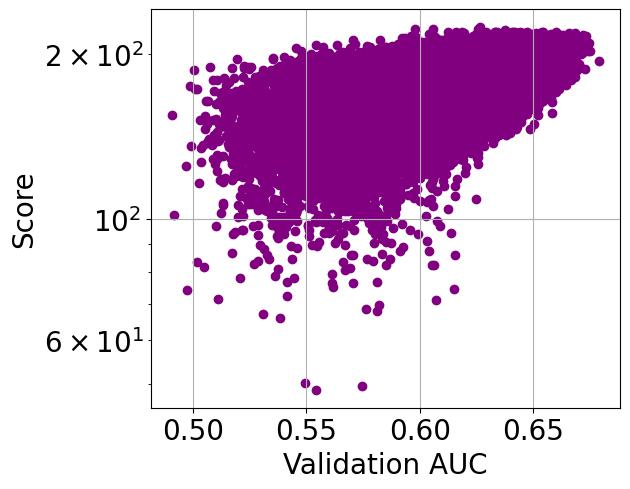}
    \caption{Diabetes}
\end{subfigure}\hspace{0.6mm}
\begin{subfigure}[b]{0.15\textwidth}
    \centering
    \includegraphics[width=\linewidth]{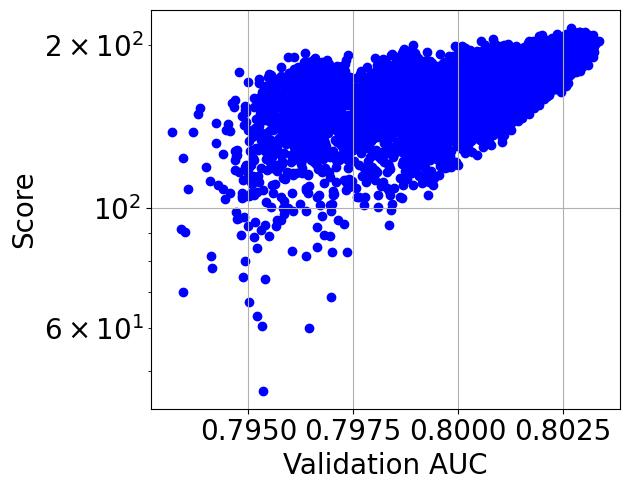}
    \caption{Criteo}
\end{subfigure}
\caption{\textbf{GradNorm} (left three) and \textbf{NASWOT} (right three): proxy score vs.\ validation AUC.}
\label{fig:gradnorm_naswot}
\end{figure*}

\begin{figure*}[!htbp]
\centering
\begin{subfigure}[b]{0.15\textwidth}
    \centering
    \includegraphics[width=\linewidth]{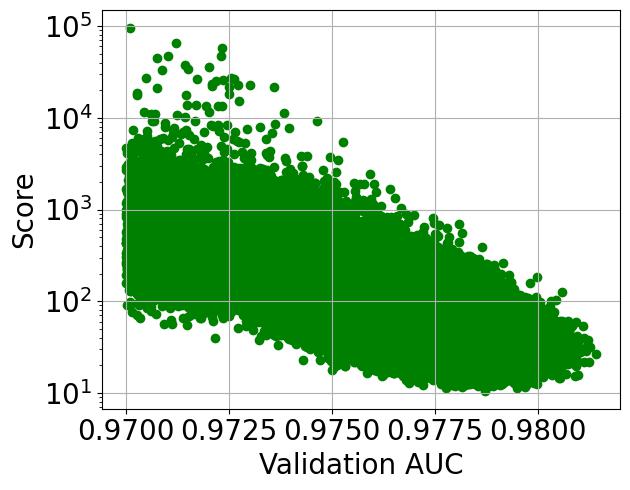}
    \caption{Frappe}
\end{subfigure}\hspace{0.6mm}
\begin{subfigure}[b]{0.15\textwidth}
    \centering
    \includegraphics[width=\linewidth]{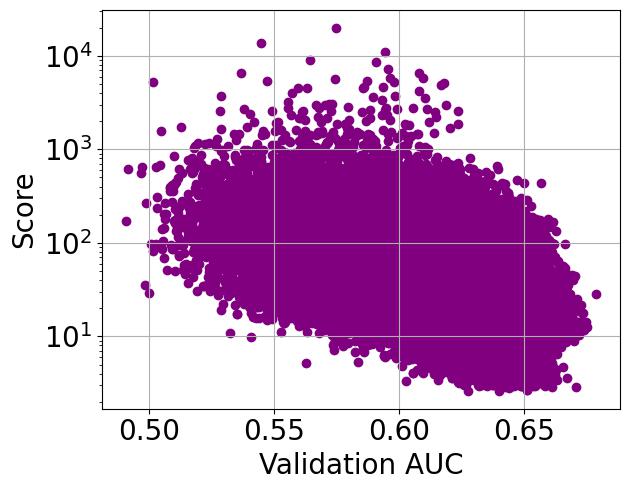}
    \caption{Diabetes}
\end{subfigure}\hspace{0.6mm}
\begin{subfigure}[b]{0.15\textwidth}
    \centering
    \includegraphics[width=\linewidth]{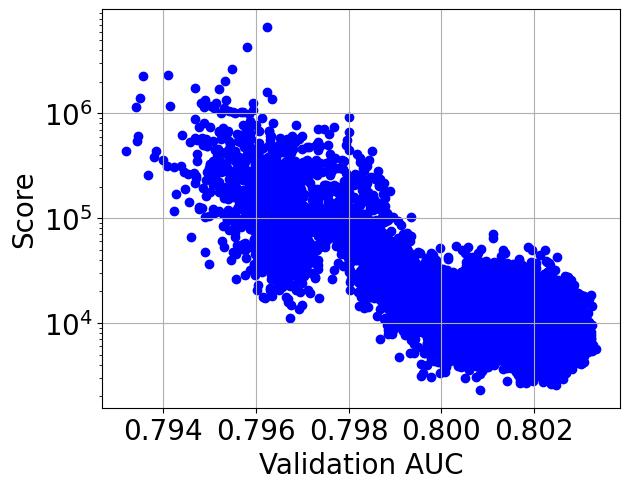}
    \caption{Criteo}
\end{subfigure}\hspace{2.0mm}
\begin{subfigure}[b]{0.15\textwidth}
    \centering
    \includegraphics[width=\linewidth]{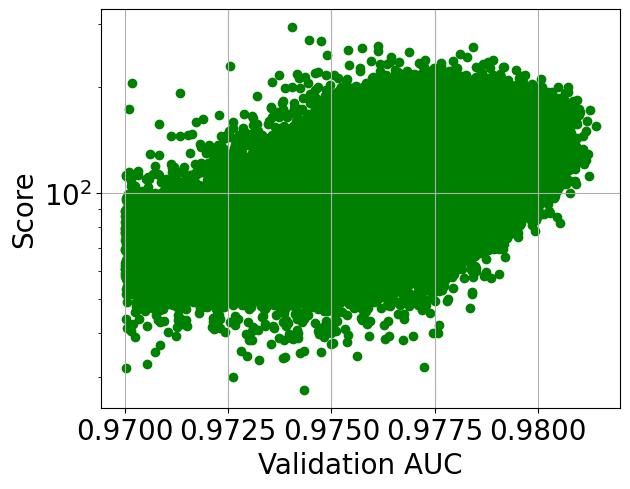}
    \caption{Frappe}
\end{subfigure}\hspace{0.6mm}
\begin{subfigure}[b]{0.15\textwidth}
    \centering
    \includegraphics[width=\linewidth]{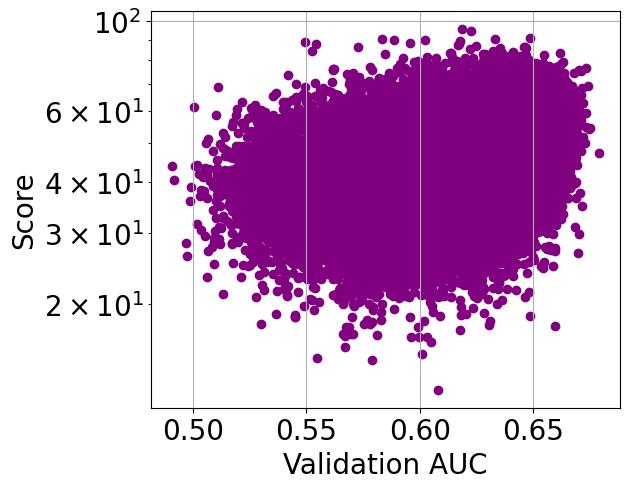}
    \caption{Diabetes}
\end{subfigure}\hspace{0.6mm}
\begin{subfigure}[b]{0.15\textwidth}
    \centering
    \includegraphics[width=\linewidth]{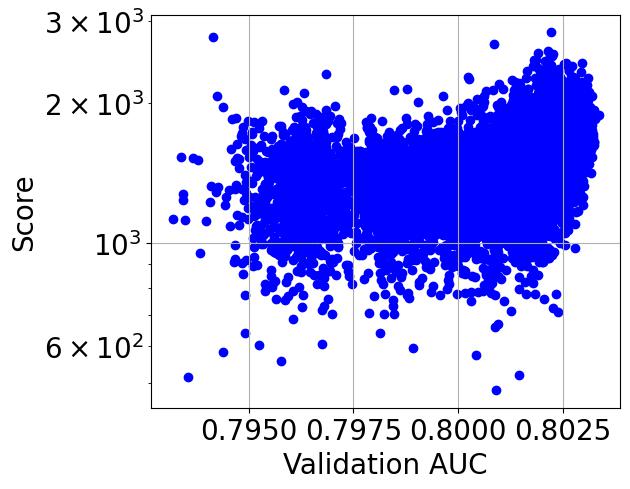}
    \caption{Criteo}
\end{subfigure}
\caption{\textbf{NTKCond} (left three) and \textbf{NTKTrace} (right three): proxy score vs.\ validation AUC.}
\label{fig:ntkcond_ntktrace}
\end{figure*}

\begin{figure*}[!htbp]
\centering
\begin{subfigure}[b]{0.15\textwidth}
    \centering
    \includegraphics[width=\linewidth]{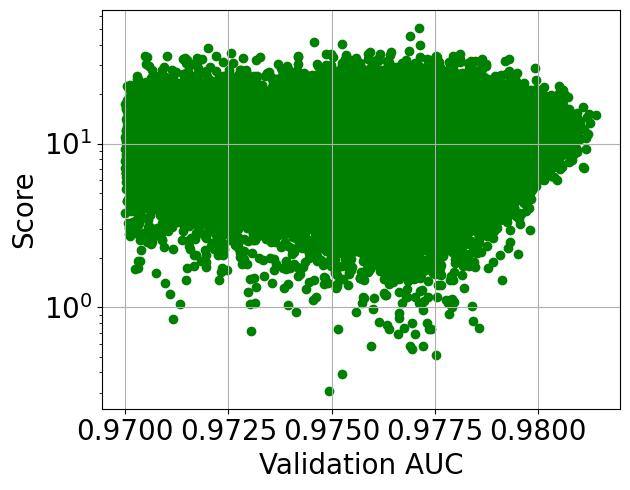}
    \caption{Frappe}
\end{subfigure}\hspace{0.6mm}
\begin{subfigure}[b]{0.15\textwidth}
    \centering
    \includegraphics[width=\linewidth]{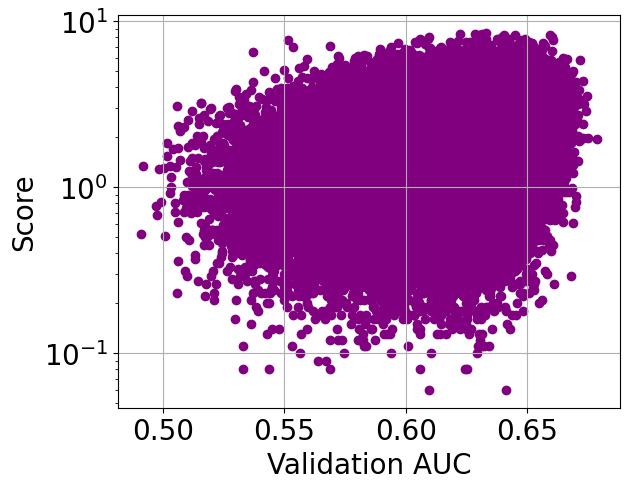}
    \caption{Diabetes}
\end{subfigure}\hspace{0.6mm}
\begin{subfigure}[b]{0.15\textwidth}
    \centering
    \includegraphics[width=\linewidth]{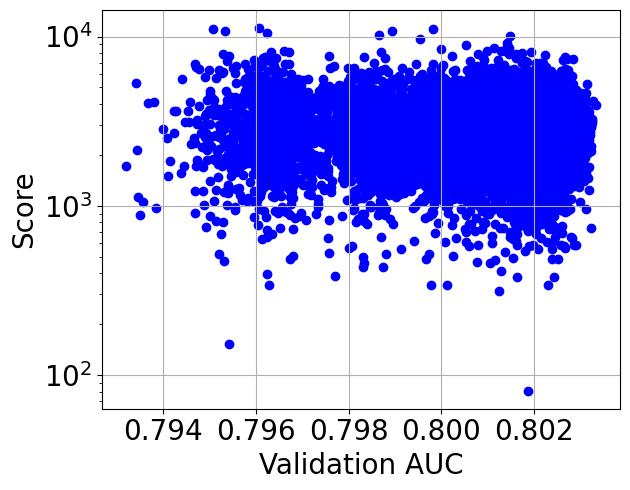}
    \caption{Criteo}
\end{subfigure}\hspace{2.0mm}
\begin{subfigure}[b]{0.15\textwidth}
    \centering
    \includegraphics[width=\linewidth]{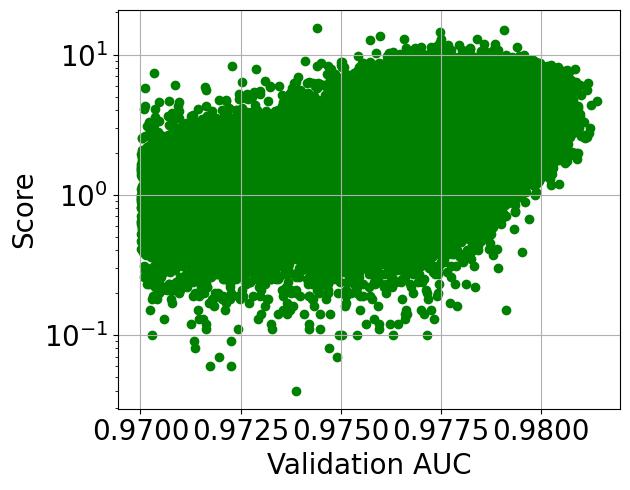}
    \caption{Frappe}
\end{subfigure}\hspace{0.6mm}
\begin{subfigure}[b]{0.15\textwidth}
    \centering
    \includegraphics[width=\linewidth]{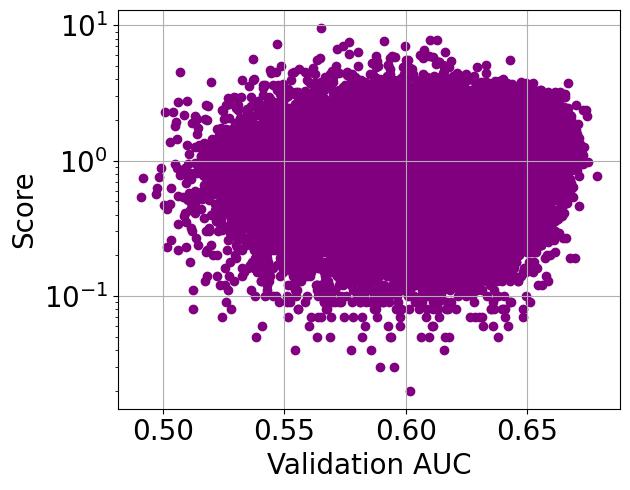}
    \caption{Diabetes}
\end{subfigure}\hspace{0.6mm}
\begin{subfigure}[b]{0.15\textwidth}
    \centering
    \includegraphics[width=\linewidth]{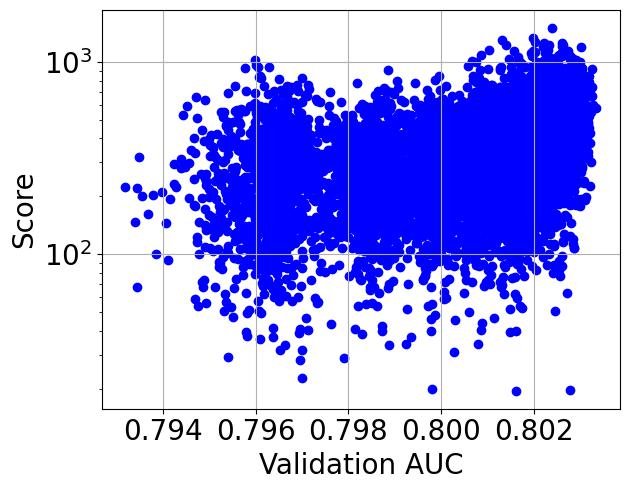}
    \caption{Criteo}
\end{subfigure}
\caption{\textbf{NTKTraceAppx} (left three) and \textbf{Fisher} (right three): proxy score vs.\ validation AUC.}
\label{fig:ntktraceappx_fisher}
\end{figure*}

\begin{figure*}[!htbp]
\centering
\begin{subfigure}[b]{0.15\textwidth}
    \centering
    \includegraphics[width=\linewidth]{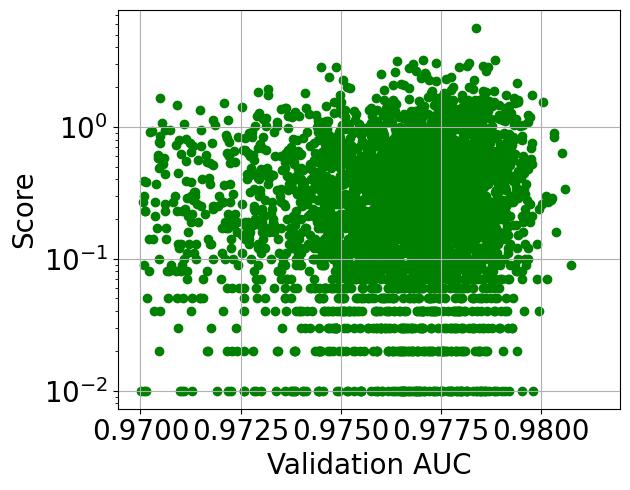}
    \caption{Frappe}
\end{subfigure}\hspace{0.6mm}
\begin{subfigure}[b]{0.15\textwidth}
    \centering
    \includegraphics[width=\linewidth]{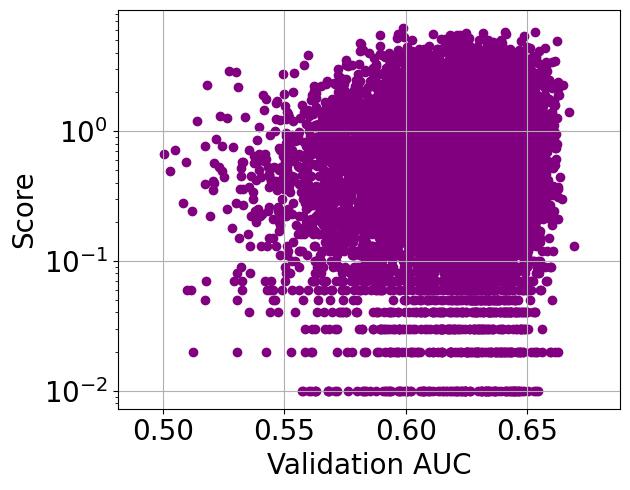}
    \caption{Diabetes}
\end{subfigure}\hspace{0.6mm}
\begin{subfigure}[b]{0.15\textwidth}
    \centering
    \includegraphics[width=\linewidth]{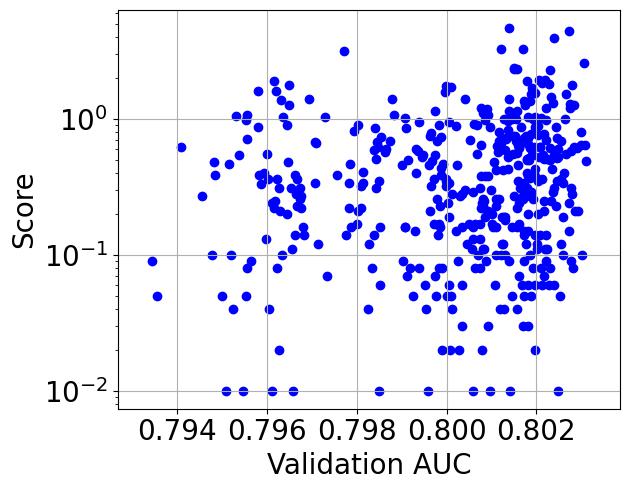}
    \caption{Criteo}
\end{subfigure}\hspace{2.0mm}
\begin{subfigure}[b]{0.15\textwidth}
    \centering
    \includegraphics[width=\linewidth]{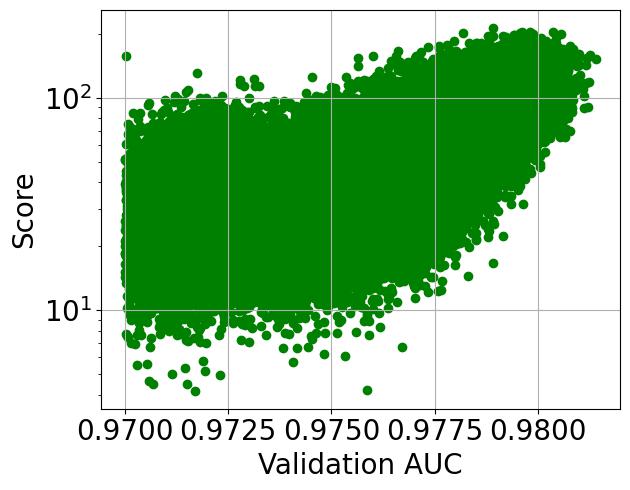}
    \caption{Frappe}
\end{subfigure}\hspace{0.6mm}
\begin{subfigure}[b]{0.15\textwidth}
    \centering
    \includegraphics[width=\linewidth]{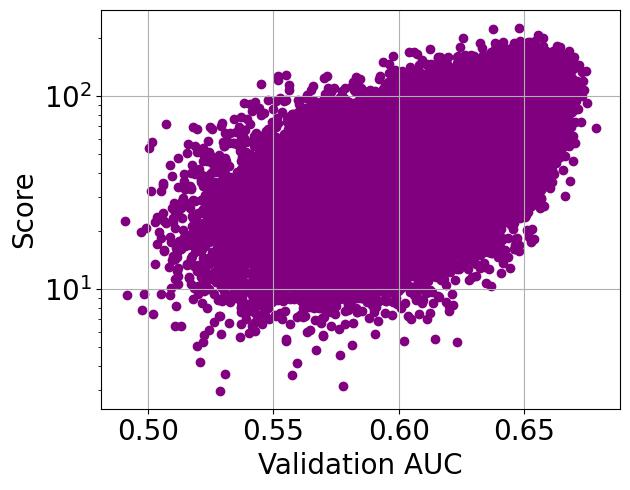}
    \caption{Diabetes}
\end{subfigure}\hspace{0.6mm}
\begin{subfigure}[b]{0.15\textwidth}
    \centering
    \includegraphics[width=\linewidth]{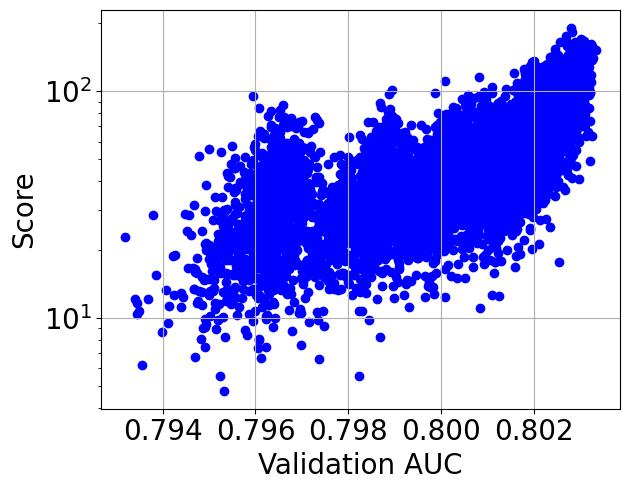}
    \caption{Criteo}
\end{subfigure}
\caption{\textbf{GraSP} (left three) and \textbf{SNIP} (right three): proxy score vs.\ validation AUC.}
\label{fig:grasp_snip}
\end{figure*}

\begin{figure*}[!htbp]
\centering
\begin{subfigure}[b]{0.15\textwidth}
    \centering
    \includegraphics[width=\linewidth]{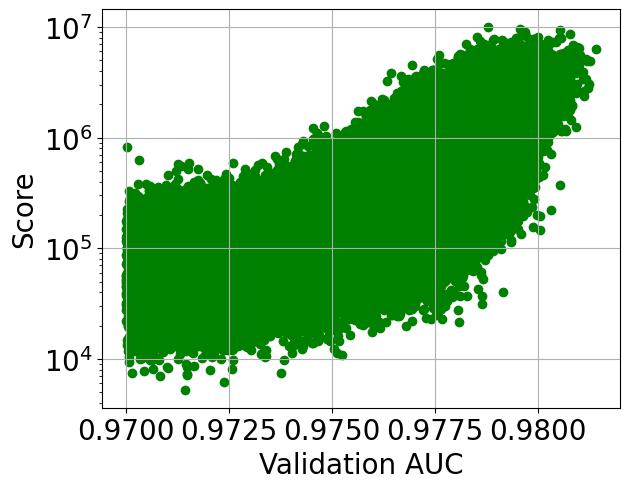}
    \caption{Frappe}
\end{subfigure}\hspace{0.6mm}
\begin{subfigure}[b]{0.15\textwidth}
    \centering
    \includegraphics[width=\linewidth]{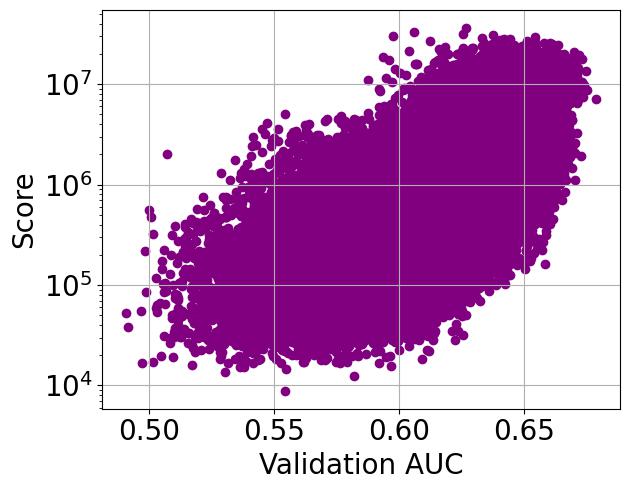}
    \caption{Diabetes}
\end{subfigure}\hspace{0.6mm}
\begin{subfigure}[b]{0.15\textwidth}
    \centering
    \includegraphics[width=\linewidth]{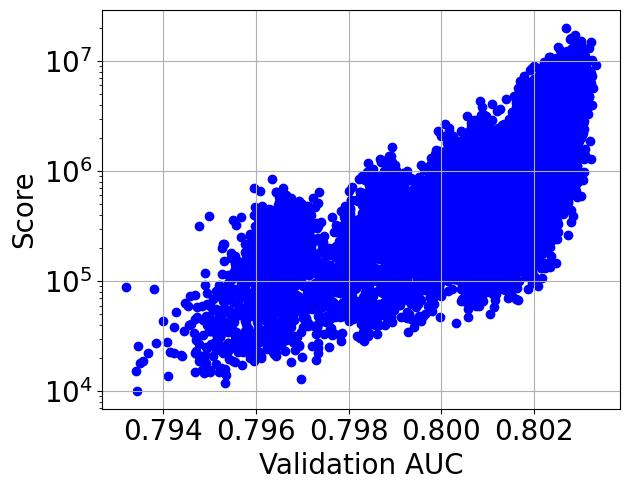}
    \caption{Criteo}
\end{subfigure}\hspace{2.0mm}
\begin{subfigure}[b]{0.15\textwidth}
    \centering
    \includegraphics[width=\linewidth]{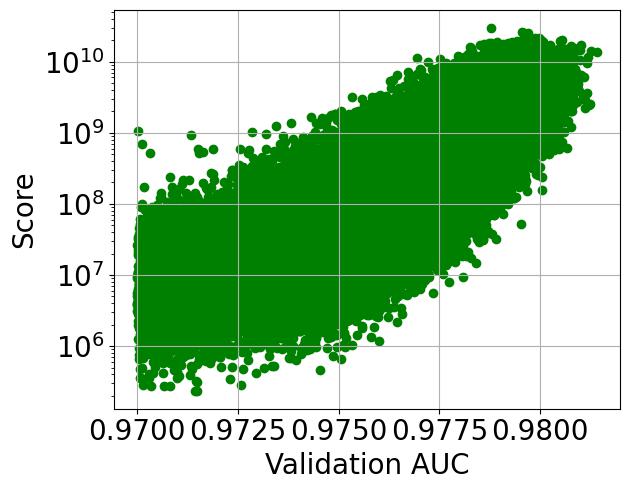}
    \caption{Frappe}
\end{subfigure}\hspace{0.6mm}
\begin{subfigure}[b]{0.15\textwidth}
    \centering
    \includegraphics[width=\linewidth]{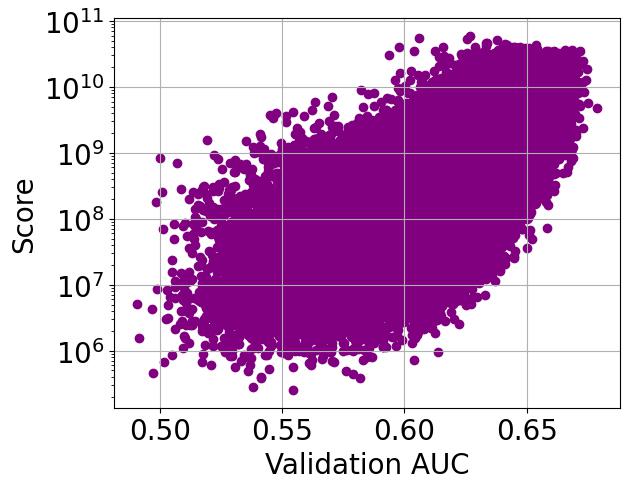}
    \caption{Diabetes}
\end{subfigure}\hspace{0.6mm}
\begin{subfigure}[b]{0.15\textwidth}
    \centering
    \includegraphics[width=\linewidth]{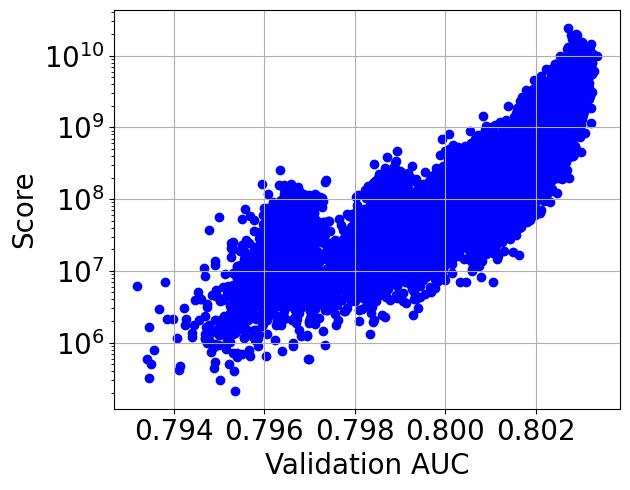}
    \caption{Criteo}
\end{subfigure}
\caption{\textbf{SynFlow} (left three) and \textbf{\newtfmem} (right three): proxy score vs.\ validation AUC.}
\label{fig:synflow_newtfmem}
\end{figure*}

For each \tfmem, we randomly sample 4000 architectures from \nasbench and compute both their validation AUC after training and the \tfmem score at initialization.
We visualize the score--AUC relationship in Figures~\ref{fig:gradnorm_naswot}, \ref{fig:ntkcond_ntktrace}, \ref{fig:ntktraceappx_fisher}, \ref{fig:grasp_snip}, and \ref{fig:synflow_newtfmem}.

Overall, these \tfmem scores exhibit a positive association with validation AUC across datasets.
In particular, \newtfmem (right three plots in Figure~\ref{fig:synflow_newtfmem}) shows a more consistent monotonic trend with AUC, supporting its effectiveness as a training-free proxy for tabular DNN architectures.

\section{Complete Results Across All Datasets}
\label{app:relbench_full}

Table~\ref{tab:result_effect_detail} reports the complete per-dataset results for the eight realistic multi-table datasets used in our paper.
Table~\ref{tab:result_time_detail} reports the efficiency results in tabular form, corresponding to Figure~\ref{fig:rel_bench_time}.

\begin{table*}[]
  \centering
  \caption{Effectiveness evaluation of \framework against existing tabular models.}
  \label{tab:result_effect_detail}
  \resizebox{0.96\textwidth}{!}{%
  \begin{threeparttable}
  \begin{tabular}{@{}c|l|ccccc|ccccc|c@{}}
  \toprule[1.5pt]
  \multicolumn{2}{c}{Task Type (Metric)} &
    \multicolumn{5}{c|}{Regression (MAE; $\downarrow$)} &
    \multicolumn{5}{c|}{Classification (AUC; $\uparrow$)} &
    \multicolumn{1}{c}{Rank ($\downarrow$)} \\ \midrule
  \multicolumn{2}{c|}{Dataset/Task} &
    \begin{tabular}[c]{@{}c@{}}Event\\ \textit{user-attend}\end{tabular} &
    \begin{tabular}[c]{@{}c@{}}Beer\\ \textit{beer-pos}\end{tabular} &
    \begin{tabular}[c]{@{}c@{}}Trial\\ \textit{site-success}\end{tabular} &
    \begin{tabular}[c]{@{}c@{}}HM\\ \textit{item-sales}\end{tabular} &
    \begin{tabular}[c]{@{}c@{}}Rank\\\textit{avg}\end{tabular} &
    \begin{tabular}[c]{@{}c@{}}Event\\ \textit{user-repeat}\end{tabular} &
    \begin{tabular}[c]{@{}c@{}}Beer\\ \textit{user-active}\end{tabular} &
    \begin{tabular}[c]{@{}c@{}}Trial\\ \textit{study-out}\end{tabular} &
    \begin{tabular}[c]{@{}c@{}}Avito\\ \textit{user-click}\end{tabular} &
    \begin{tabular}[c]{@{}c@{}}Rank\\\textit{avg}\end{tabular} &
    \begin{tabular}[c]{@{}c@{}}Global\\Rank\end{tabular}
    \\ \cmidrule(r){1-2}
  Type & \multicolumn{1}{c|}{Method} & & & & & & & & & & \\ \midrule
  
  \multicolumn{1}{c|}{\multirow{4}{*}{CM}} &
    \multicolumn{1}{l|}{LR} &
    0.3912 & 0.2046 & 0.4594 & 0.0659 & 12.25 &
    0.7376 & 0.8812 & 0.6881 & 0.6407 & 11.25 & 11.75 \\
    
  \multicolumn{1}{c|}{} &
    \multicolumn{1}{l|}{RF} &
    0.3745 & 0.1997 & 0.4565 & 0.0584 & 10.63 &
    0.7270 & 0.8737 & 0.6770 & 0.6450 & 12.75 & 11.69 \\
  \multicolumn{1}{c|}{} &
    \multicolumn{1}{l|}{CatBoost} &
    0.2607 & 0.1975 & \textbf{0.4429} & 0.0557 & 5.75 &
    0.7429 & 0.9060 & 0.6945 & 0.6504 & 7.00 & 6.25 \\
  \multicolumn{1}{c|}{} &
    \multicolumn{1}{l|}{LightGBM} &
    0.2547 & 0.1903 & 0.4595 & \best{0.0495} & 5.63 &
    0.7340 & 0.9061 & 0.6999 & 0.6527 & 7.00 & 6.31 \\ \midrule
  
  \multicolumn{1}{c|}{\multirow{2}{*}{TFM}} &
    \multicolumn{1}{l|}{TabPFN} &
    0.3406 & 0.1903 & 0.4639 & 0.0583 & 9.88 &
    \textbf{0.7754} & 0.9164 & \textbf{0.7051} & 0.6437 & 5.00 & 7.44 \\
    
  \multicolumn{1}{c|}{} &
    \multicolumn{1}{l|}{TabICL\textsuperscript{$\ast$}} &
    - & - & - & - & - &
    0.7692 & 0.9077 & 0.7018 & 0.6503 & \textbf{4.50} & \textbf{4.50} \\ \midrule
  
  \multicolumn{1}{c|}{\multirow{4}{*}{DTM}} &
    \multicolumn{1}{l|}{DNN} &
    0.2523 & 0.1947 & 0.4500 & 0.0551 & 5.25 &
    0.7294 & 0.9064 & 0.6830 & 0.6546 & 8.25 & 6.75 \\
    
  \multicolumn{1}{c|}{} &
    \multicolumn{1}{l|}{DeepFM} &
    \textbf{0.2491} & 0.2091 & 0.4581 & 0.0541 & 7.50 &
    0.7130 & 0.9047 & 0.7013 & 0.6283 & 10.50 & 9.00 \\
    
  \multicolumn{1}{c|}{} &
    \multicolumn{1}{l|}{FTTrans} &
    0.2539 & \textbf{0.1825} & \best{0.4290} & 0.0584 & \textbf{4.13} &
    0.7346 & 0.9131 & 0.6836 & 0.6502 & 8.00 & 6.06 \\
    
  \multicolumn{1}{c|}{} &
    \multicolumn{1}{l|}{ARM-Net} &
    0.2642 & 0.1912 & 0.4468 & 0.0515 & 5.50 &
    0.7402 & 0.9016 & 0.6965 & \textbf{0.6604} & 6.75 & 6.13 \\ \midrule
  
  \multicolumn{1}{c|}{\multirow{3}{*}{LLM}} &
    \multicolumn{1}{l|}{TP-BERTa} &
    0.2768 & 0.3155 & 0.4612 & 0.3514 & 13.75 &
    0.5457 & 0.5170 & - & 0.5122 & 16.00 & 14.71 \\
    
  \multicolumn{1}{c|}{} &
    \multicolumn{1}{l|}{Nomic} &
    0.2677 & 0.3439 & 0.4545 & 0.2063 & 12.25 &
    0.6896 & 0.8896 & 0.6533 & 0.5771 & 13.75 & 13.00 \\
    
  \multicolumn{1}{c|}{} &
    \multicolumn{1}{l|}{BGE} &
    0.2645 & 0.2829 & 0.4511 & 0.0772 & 10.50 &
    0.6787 & 0.8868 & 0.6503 & 0.6462 & 13.25 & 11.88 \\ \midrule
  
  \multicolumn{1}{c|}{\multirow{3}{*}{\shortstack{NAS\\($T_{\text{max}}=10$s)}}} &
    \multicolumn{1}{l|}{TabNAS} &
    0.2635 & 0.1994 & 0.4520 & 0.0507 & 6.50 &
    0.7512 & \textbf{0.9357} & 0.6902 & 0.6480 & 6.00 & 6.25 \\
  \multicolumn{1}{c|}{} &
    \multicolumn{1}{l|}{EA-NAS} &
    0.2639 & 0.2002 & 0.4485 & 0.0797 & 9.00 &
    0.7518 & 0.9227 & 0.7017 & 0.6473 & 5.00 & 7.00 \\
  
  \multicolumn{1}{c|}{} &
    \multicolumn{1}{l|}{\cellcolor{blue!4}\textbf{\framework} } &
    \cellcolor{blue!4}\best{0.2432} & \cellcolor{blue!4}\best{0.1794} & \cellcolor{blue!4}0.4466 & \cellcolor{blue!4}\textbf{0.0497} & \cellcolor{blue!4}\best{1.75} &
    \cellcolor{blue!4}\best{0.7769} & \cellcolor{blue!4}\best{0.9370} & \cellcolor{blue!4}\best{0.7068} & \cellcolor{blue!4}\best{0.6680} & \cellcolor{blue!4}\best{1.00} & \cellcolor{blue!4}\best{1.38} \\
  
  \bottomrule[1.5pt]
  \end{tabular}
  \begin{tablenotes}
    \item $\ast$ TabICL is inherently limited to classification objectives and is not applicable to regression-based tasks.
  \end{tablenotes}
  \end{threeparttable}
  }
  \end{table*}

\begin{table*}[t]
\centering
\caption{\textbf{Average fitting, inference, and total time (seconds) per model, aggregated over all datasets in each task type}. Fitting includes NAS time if applicable: $\text{fit}=\text{NAS}+\text{train}$; Total $=\text{fit}+\text{inference}$. \\
Efficiency-effectiveness trade-off: while \framework incurs additional fitting overhead due to NAS, it delivers stronger predictive performance (lower MAE on regression and higher AUC on classification), achieving the best average ranks on both task types and the best global rank in Tab.~\ref{tab:result_effect_detail} (Regression rank avg: 1.75; Classification rank avg: 1.00; Global rank: 1.38). Notably, although \framework is not always the fastest on classification, its inference latency remains small, and the accuracy gains are substantial.
}
\label{tab:result_time_detail}
\resizebox{0.90\textwidth}{!}{%
\begin{tabular}{@{}cl|ccc|ccc@{}}
\toprule[1.5pt]
\multicolumn{2}{c|}{Type / Method} &
\multicolumn{3}{c|}{Regression $\downarrow$} &
\multicolumn{3}{c}{Classification $\downarrow$} \\
\cmidrule(lr){3-5}\cmidrule(lr){6-8}
& & Fitting (NAS+Train) & Inference & Total &
    Fitting (NAS+Train) & Inference & Total \\
\midrule

\multicolumn{1}{c|}{\multirow{2}{*}{CM}} &
\multicolumn{1}{l|}{CatBoost} &
804.49 & \best{0.22} & 804.71 &
47.24 & \best{0.04} & 47.28 \\
\multicolumn{1}{c|}{} &
\multicolumn{1}{l|}{LightGBM} &
\textbf{101.84} & \textbf{0.27} & 102.11 &
\textbf{10.78} & \best{0.04} & \textbf{10.82} \\
\midrule

\multicolumn{1}{c|}{\multirow{1}{*}{TFM}} &
\multicolumn{1}{l|}{TabPFN} &
105.33 & 54.91 & 160.24 &
20.74 & 17.99 & 38.73 \\
\midrule

\multicolumn{1}{c|}{\multirow{3}{*}{DTM}} &
\multicolumn{1}{l|}{DNN} &
254.16 & 3.32 & 257.48 &
56.34 & 0.24 & 56.58 \\
\multicolumn{1}{c|}{} &
\multicolumn{1}{l|}{DeepFM} &
149.47 & 0.35 & 149.82 &
\best{9.66} & \textbf{0.12} & \best{9.78} \\
\multicolumn{1}{c|}{} &
\multicolumn{1}{l|}{FTTrans} &
407.03 & 3.63 & 410.66 &
37.81 & 0.33 & 38.14 \\
\midrule

\multicolumn{1}{c|}{NAS} &
\multicolumn{1}{l|}{\textbf{pTNAS}} &
\best{33.12} & 0.41 & \best{33.53} &
18.81 & \textbf{0.12} & 18.93 \\
\bottomrule[1.5pt]
\end{tabular}%
}
\end{table*}

\end{appendix}

\clearpage


\end{document}